\documentclass[10pt,journal,compsoc]{IEEEtran}

\usepackage{hyperref,graphicx,amsmath,booktabs,multirow,ragged2e,amssymb,array}
\usepackage[caption=false,font=normalsize,labelfont=sf,textfont=sf]{subfig}
\usepackage[ruled,linesnumbered]{algorithm2e}
\usepackage[nocompress]{cite}

\DeclareMathOperator*{\mmd}{{MMD}^2}
\DeclareMathOperator*{\ed}{{ED}^2}
\DeclareMathOperator*{\swd}{SWD_1}
\DeclareMathOperator*{\m}{\mathcal{M}}
\DeclareMathOperator*{\exc}{\mathbb{E}}
\newcolumntype{P}[1]{>{\centering\arraybackslash}p{#1}}
\newcommand\impath{.}
\newcommand\hl{\underline}
\newcommand\colwid{1.1cm}

\begin{document}
	
\title{Enhancing Backdoor Attacks with Multi-Level MMD Regularization}

\author{
Pengfei~Xia, Hongjing~Niu, Ziqiang~Li, and~Bin~Li,~\IEEEmembership{Member,~IEEE}
\IEEEcompsocitemizethanks{
\IEEEcompsocthanksitem P. Xia, Z. Li and B. Li are with the Department of Electronic Engineering and Information Science, University of Science and Technology of China, Hefei, China. \\ E-mail: \{xpengfei, iceli\}@mail.ustc.edu.cn, binli@ustc.edu.cn.
\IEEEcompsocthanksitem H. Niu is with the Department of Automation, University of Science and Technology of China, Hefei, China. \\ E-mail: sasori@mail.ustc.edu.cn.
}
}

\IEEEtitleabstractindextext{
\begin{abstract}
While Deep Neural Networks (DNNs) excel in many tasks, the huge training resources they require become an obstacle for practitioners to develop their own models. It has become common to collect data from the Internet or hire a third party to train models. Unfortunately, recent studies have shown that these operations provide a viable pathway for maliciously injecting hidden backdoors into DNNs. Several defense methods have been developed to detect malicious samples, with the common assumption that the latent representations of benign and malicious samples extracted by the infected model exhibit different distributions. However, a comprehensive study on the distributional differences is missing. In this paper, we investigate such differences thoroughly via answering three questions: 1) What are the characteristics of the distributional differences? 2) How can they be effectively reduced? 3) What impact does this reduction have on difference-based defense methods? First, the distributional differences of multi-level representations on the regularly trained backdoored models are verified to be significant by introducing Maximum Mean Discrepancy (MMD), Energy Distance (ED), and Sliced Wasserstein Distance (SWD) as the metrics. Then, ML-MMDR, a difference reduction method that adds multi-level MMD regularization into the loss, is proposed, and its effectiveness is testified on three typical difference-based defense methods. Across all the experimental settings, the F1 scores of these methods drop from 90\%-100\% on the regularly trained backdoored models to 60\%-70\% on the models trained with ML-MMDR. These results indicate that the proposed MMD regularization can enhance the stealthiness of existing backdoor attack methods. The prototype code of our method is now available at \url{https://github.com/xpf/Multi-Level-MMD-Regularization}.
\end{abstract}

}

\maketitle

\IEEEraisesectionheading{\section{Introduction} \label{sec:int}}
\IEEEPARstart{I}{n} the past years, Deep Neural Networks (DNNs) have shown impressive achievements in computer vision \cite{girshick2015fast,long2015fully,redmon2018yolov3,yin2019online,xia2020boosting}, natural language processing \cite{sutskever2014sequence,chen2016enhanced,devlin2018bert}, and some other fields \cite{chan2016listen,silver2016mastering,jumper2021highly}. It is widely believed that the success of DNNs is closely related to their large-scale models, consumption of a huge amount of training data, and computational power. For example, GPT-3 \cite{brown2020language}, a deep model demonstrated to be effective in various tasks, comprises 175 billion parameters and is pretrained on 45 TB of text data. A training cycle for this model on a Tesla V100 GPU would require \$4.6 million and 355 years\footnote{\url{https://lambdalabs.com/blog/demystifying-gpt-3/}}. The superior performance of DNNs usually comes at the cost of massive time and economic resources.

\begin{figure}[!ht]
\centering
\includegraphics[width=7.6cm]{\impath/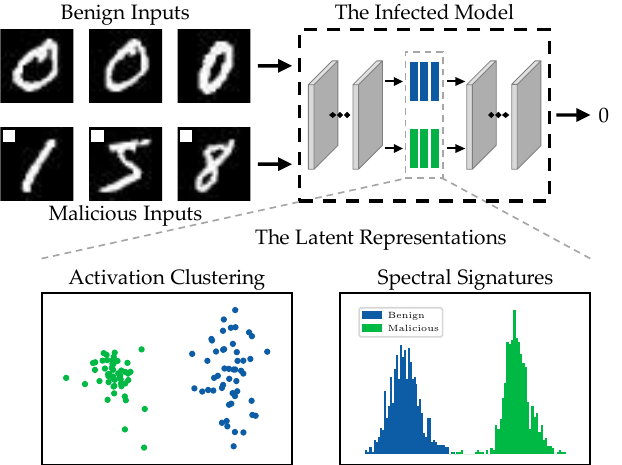}
\caption{An illustrative example of difference-based defense methods. The infected model is trained to classify all inputs with a trigger, i.e., the white square in the upper left corner of the images, as the number 0. Some defense methods, such as activation clustering \cite{chen2019detecting} and spectral signatures \cite{tran2018spectral}, use the latent representations extracted by the infected model to distinguish malicious samples from benign ones.}
\label{fig:ill_exa}
\end{figure}

To save on training costs, it has become common for users and companies to collect data from the Internet, use pretrained parameters, or hire a third party to train models. Unfortunately, these operations pose security risks due to the possible presence of malicious sources and parties. One of the major threats is dubbed as \textit{backdoor attacks} or \textit{Trojan attacks} \cite{chen2017targeted,gu2017badnets,liu2017trojaning,xue2020one}, where a hidden backdoor is injected into the victim model during the training phase and can be activated by a predefined trigger. The infected model would exhibit duel characteristics. When the backdoor is not activated, it behaves as normal as a benign model. But once triggered, the predictions are forced to an attacker-specific target. Backdoor attacks have threatened the deployment of DNNs in security-sensitive scenarios, such as face recognition systems \cite{chen2017targeted} and autonomous vehicles \cite{gu2017badnets}.

Several defense methods \cite{chen2019detecting,tran2018spectral,jin2020unified,hayase2021spectre} have been presented by distinguishing malicious samples from benign ones according to their latent representations extracted by the infected model. For example, Chen et al. \cite{chen2019detecting} proposed activation clustering, which is based on the observation that the projected activations of the last hidden layer are divided into two distinct clusters. Tran et al. \cite{tran2018spectral} proposed calculating an outlier score for each sample by performing singular value decomposition on the representations. The inputs with the top scores would be removed as malicious samples. An illustrative example is shown in Fig. \ref{fig:ill_exa}. 

The previously presented defense methods all rely on a common assumption: the latent representations of benign and malicious samples exhibit different distributions. A naturally arisen question is whether this assumption would always be true. There is a shortage of a comprehensive investigation on this issue. Some reduction methods \cite{tan2020bypassing,doan2021backdoor,ren2021simtrojan} have been proposed to reduce the distributional difference in the latent space. However, in Section \ref{sec:red_dif} and Section \ref{sec:exp_def}, we will show that these methods are not sufficient to provide a solid solution.

In this paper, we focus on this commonly adopted assumption and investigate the following questions:
\begin{itemize}
\item What are the characteristics of the distributional differences between benign and malicious samples in the latent space extracted by the infected model?
\item How can they be effectively reduced when conducting backdoor attacks?
\item What impact does this reduction have on difference-based defense methods?
\end{itemize}

The main contributions of this paper include:
\begin{itemize}
\item Three typical metrics, i.e., Maximum Mean Discrepancy (MMD) \cite{gretton2012kernel}, Energy Distance (ED) \cite{szekely2013energy}, and Sliced Wasserstein Distance (SWD) \cite{kolouri2019generalized}, are introduced to explicitly quantify the distributional differences in the latent spaces. The differences between benign and malicious samples in multi-level representations are verified to be large. This motivates us to take the multiple levels of features into consideration when designing the reduction method, other than those of the last hidden layer.
\item A new method, ML-MMDR, is proposed by introducing \textbf{M}ulti-\textbf{L}evel \textbf{MMD} \textbf{R}egularization into the loss when training a backdoored model. The experimental results indicate that the proposed method can fully reduce the differences without compromising the attack strength.
\item The effectiveness of ML-MMDR is testified on three typical difference-based defense methods. The experimental results show that the performance of these methods is substantially degraded by the proposed regularization. This illustrates that ML-MMDR can enhance existing attack methods to escape detection.
\end{itemize}

The rest of this paper is organized as follows. In Section \ref{sec:pre}, the preliminaries are briefly introduced. Section \ref{sec:set} provides the setup used in the following three sections. Section \ref{sec:exp_cha}, Section \ref{sec:red_dif}, and Section \ref{sec:exp_def} provide our detailed works addressing the three mentioned questions. Some open issues are discussed in Section \ref{sec:dis}, and the related works are reviewed in Section \ref{sec:rel_wor}. Section \ref{sec:con} concludes this paper. 

\section{Preliminaries} \label{sec:pre}
\subsection{Deep Neural Networks}
A DNN is composed of multiple layers and can be defined as $f_{\theta} = f_1 \circ f_2 \circ \cdots \circ f_m$, where $\theta$ denotes the parameters of the model, $f_1$, $f_2$, $\cdots$, $f_{m - 1}$ denote the $m - 1$ hidden layers, and $f_m$ denotes the output layer. For convenience, let $z_i = f_1 \circ f_2 \circ \cdots \circ f_i$ denote the structure from the input to the $i$-th hidden layer and $z_i(x)$ denote the extracted representation of the input $x$. Given a training set $D = \{(x, y)\}$, the procedure of training a DNN can be formulated as:
\begin{equation}
\min_{\theta} \frac{1}{|D|} \sum_{(x, y) \in D} L(f_{\theta}(x), y) \text{,}
\end{equation}
where $(x, y)$ denote the input and its ground-truth label, and $L$ denotes the loss function. The trained model is expected to perform well on a test set $T$, and $D \cap T = \varnothing$ is required.

\subsection{Backdoor Attacks}
In this paper, the scenario of injecting a backdoor into a DNN by dataset poisoning \cite{chen2017targeted,gu2017badnets,liu2017trojaning,barni2019new} is considered. Let $U$ and $V$ denote the malicious training and test sets. The generation of malicious samples is associated with a target class $t$, a trigger $b$, and a fusion function $G$. For a benign pair $(x, y)$, the corresponding malicious pair is $(x', t)$, where $x' = G(x, b)$. To train a DNN with a backdoor, one can optimize the equation:
\begin{equation}
\min_{\theta} {\frac{1}{|D|} \sum_{(x, y) \in D} L(f_{\theta}(x), y) + \frac{1}{|U|} \sum_{(x', t) \in U} L(f_{\theta}(x'), t)} \text{,} \label{equ:rbt}
\end{equation}
and we suppose the trained model can generalize to the benign and malicious test sets $T$ and $V$. The ratio of malicious sample volume to benign sample volume in training data, $r = |U|/|D|$, is an important hyperparameter.

\subsection{Statistical Distance}
Let $X \sim p$ and $Y \sim q$ denote two sample sets drawn from the distributions $p$ and $q$. A statistical distance takes the two sets as its input and returns a real number as the distance between $p$ and $q$. In this paper, since the latent representations to be measured are high-dimensional data and their distribution functions are difficult to solve, MMD, ED, SWD are chosen as the metrics.

\noindent\textbf{Maximum Mean Discrepancy.} Gretton et al. \cite{gretton2012kernel} introduced a kernel-based metric, MMD, to quantify the distance between two distributions. It can be calculated as:
\begin{equation}
\begin{split}
\mmd(X, Y) =& \frac{1}{m^2} \sum_{i,j=1}^{m} k(x_{i}, x_{j}) + \frac{1}{n^2} \sum_{i,j=1}^{n} k(y_{i}, y_{j}) \\
& - \frac{2}{mn} \sum_{i,j=1}^{m,n}   k(x_{i}, y_{j})
\end{split} \text{,}
\end{equation}
where $m$ and $n$ denote the sizes of $X$ and $Y$, respectively. $k$ is a kernel function and $k(x, y) = \langle\phi(x), \phi(y)\rangle$, where $\phi$ denotes a feature mapping. 

\noindent\textbf{Energy Distance.} Szekely and Rizzo \cite{szekely2013energy} first introduced this metric. It is a specific case of MMD, where no kernel is applied. The equation is:

\begin{equation}
\begin{split}
\ed(X, Y) =& \frac{1}{m^2} \sum_{i,j=1}^{m} ||x_{i} - x_{j}|| + \frac{1}{n^2} \sum_{i,j=1}^{n} ||y_{i} - y_{j}|| \\
& - \frac{2}{mn} \sum_{i,j=1}^{m,n}   ||x_{i} - y_{j}||
\end{split} \text{.}
\end{equation}

\noindent\textbf{Sliced Wasserstein Distance.} SWD is a potential alternative to Wasserstein distance \cite{villani2009optimal}, which can be  be approximated more easily. The underlying idea is to decompose high-dimensional data into one-dimensional data via random linear projection. SWD with $l_1$ cost can be computed as:
\begin{equation}
\swd(X, Y) = \exc_{R} \bigg[\min_{\sigma \in S_m} \sum_{i=1}^{m} |R(x_{\sigma_i}) - R(y_i)| \bigg] \text{,}
\end{equation}
where $R$ denotes a linear projection, and $\sigma \in S_m$ denotes a possible sample sequence \cite{kolouri2019generalized}. 

\section{Setup} \label{sec:set}
\subsection{Threat Model}
Our threat model considers that the user needs to use a DNN trained in an untrusted environment. This model is prevalent among companies, such as car manufacturers and autonomous driving solution providers. We assume that the attacker has complete control over the training data and the training procedure. The defender's goal is to distinguish malicious samples from benign ones. The defender does not have any prior knowledge about the attack but has access to the parameters of the DNN and keeps a few (about 200) benign validation data.

\subsection{Attack Methods}
\begin{figure}[!ht]
\centering
\includegraphics[width=8.8cm]{\impath/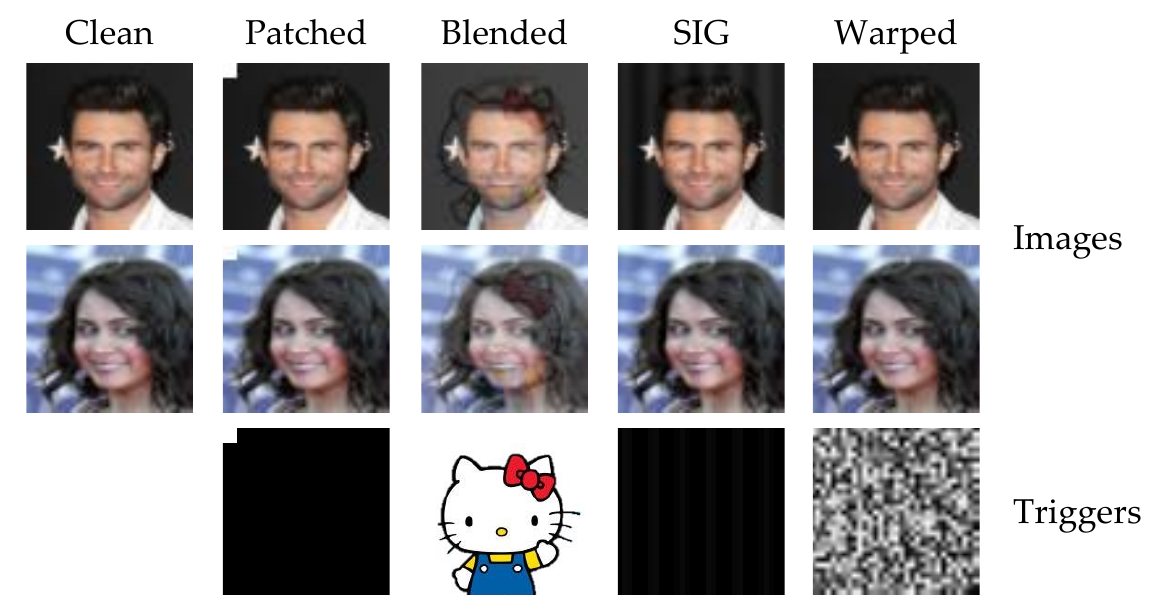}
\caption{Malicious examples from CelebA \cite{liu2015deep} generated by four backdoor attack methods.}
\label{fig:attacks}
\end{figure}

Four backdoor attack methods are considered, including patch-based attack (Patched) \cite{gu2017badnets}, blending-based attack (Blended) \cite{chen2017targeted}, SIG \cite{barni2019new}, and warping-based attack (Warped) \cite{nguyen2021wanet}. The main differences between these methods are the trigger $b$ and the fusion function $G$. The forms of generating malicious samples are shown in TABLE \ref{tab:form}, and some examples are shown in Fig. \ref{fig:attacks}. The other settings are kept the same for these methods. The target $t$ is set to the 0th category, and the poisoning data ratio, $r$, is set to 0.1.

\begin{table}[h] 
\centering 
\caption{Forms of generating malicious samples. $m$: a 2D mask. $\alpha$: a hyperparameter between 0 to 1. $\mathcal{W}$: the warp function.} 
\label{tab:form}
\begin{tabular}{l|l} 
\toprule
Attack  & Fusion Function $x' = G(x, b)$               \\ \midrule
Patched & $x' = (1 - m) \odot x + m \odot b$           \\
Blended & $x' = (1 - \alpha) \cdot x + \alpha \cdot b$ \\
SIG     & $x' = x + b$                                 \\
Warped  & $x' = \mathcal{W}(x, b)$                     \\
\bottomrule
\end{tabular}
\end{table}

\subsection{Defense Methods}
Four defense methods are adopted to test the effect of attack methods, three of which are difference-based detection methods. Note that some of these methods were proposed for filtering malicious samples in the training phase. We believe that it is also reasonable to apply them in the test phase, and the experimental results show that they perform well. Besides, the original detection methods mainly utilize the outputs of the last hidden layer. We extend them to multi-level representations for a more thorough analysis.

\noindent\textbf{Activation Clustering.} Chen et al. \cite{chen2019detecting} proposed activation clustering to detect malicious samples, which does not require verified data. The author first observed that the features of benign and malicious inputs, which are classified into the same category by the backdoored model, visually show two distinct clusters. Then, they proposed filtering malicious samples with three steps: 1) Perform independent component analysis on the latent representations to reduce the dimensionality. 2) Use k-means to cluster the data represented by the reduced features into two clusters. 3) Analyze whether each cluster is benign or malicious by exclusionary reclassification, relative size comparison, or silhouette score.

\noindent\textbf{Spectral Signatures.} Tran et al. \cite{tran2018spectral} identified a characteristic of backdoor attacks called spectral signatures. A spectral signature is a detectable trace in the spectrum of the representations' covariance that these attacks tend to leave behind. They then proposed an algorithm to show that one could use this trace to remove malicious samples with three steps: 1) Perform singular vector decomposition on the centered features. 2) Calculate the outlier scores with the top right singular vector. 3) Remove data with the top-k scores.

\noindent\textbf{Subspace Reconstruction.} The core idea of subspace reconstruction for backdoor detection is to construct a subspace learned from a small number of benign representations and assume that projecting malicious representations into this space would cause considerable information to be lost. As a result, the reconstruction loss of these features is higher than that of benign features. Javaheripi et al. \cite{javaheripi2020cleann} first introduced this idea into their Trojan detection framework.

\noindent\textbf{Neural Cleanse.} Wang et al. \cite{wang2019neural} proposed neural cleanse, a detection and mitigation system for backdoor attacks. In the detection step, the method first uses reverse engineering to obtain potential triggers towards every category and then determines the final synthetic trigger according to its $l_1$ norm. The authors introduced three techniques, i.e., input filtering, neuron pruning, and unlearning, in the mitigation step to remove the backdoor with the synthetic trigger. Since input filtering is similar to the three detection methods described above, unlearning needs to retrain the model, we apply neuron pruning in this paper.

\subsection{Experimental Settings}
\noindent\textbf{Datasets.} Two image datasets, CIFAR-10 \cite{krizhevsky2009learning} and CelebA \cite{liu2015deep}, are selected, which are often used for backdoor learning tasks. For CelebA, following the configuration in \cite{salem2020dynamic,nguyen2021wanet}, we use the three most balanced attributes, i.e., ``Heavy Makeup'', ``Mouth Slightly Open'', and ``Smiling'', to create eight categories for image classification.

\noindent\textbf{DNN Architectures.} Four DNN architectures, VGG-11 (V-11) \cite{simonyan2014very}, VGG-16 (V-16) \cite{simonyan2014very}, ResNet-18 (R-18) \cite{he2016deep}, and PreActResNet-18 (P-18) \cite{he2016identity}, are used in our experiments. Since the activations of the hidden layers need to be extracted as the latent representations, we define $s_1$, $s_2$, and $s_3$ to represent three locations in the network structure. $z_{s_1}(x)$, $z_{s_2}(x)$, and $z_{s_3}(x)$ denote the extracted features of $x$ from the corresponding levels. Specifically, we set $s_1$, $s_2$, and $s_3$ to the 14th, 21st, and 28th layers for V-11, the 23rd, 33rd, and 43rd layers for V-16, and the 27th, 39th, and 51st layers for R-18 and P-18. More details of the DNN architectures can be seen in Appendix \ref{app:dnn_arc}.

\noindent\textbf{Evaluation Metrics.} Benign Accuracy (BA) and Attack Success Rate (ASR) are adopted as the metrics for measuring the performance of an infected model on the test sets $T$ and $V$. For a difference-based defense method, we use F1 score to evaluate its performance.

\noindent\textbf{Implementation Details.} We adopt stochastic gradient descent with a momentum of 0.9 and a weight decay of 5e-4 as the optimizer for all experiments. The batch size is set to 256, and the total training duration is 100 epochs for CIFAR-10 and 40 epochs for CelebA. The initial learning rate is set to 0.01 and it is dropped by 10 after 50 and 70 epochs for CIFAR-10 and 20 and 30 epochs for CelebA. All images are resized to 32$\times$32 and normalized between 0 and 1. Our code\footnote{\url{https://github.com/xpf/Multi-Level-MMD-Regularization}} is implemented with PyTorch \cite{paszke2017automatic}.

\section{Distributional Differences between Benign and Malicious Samples} \label{sec:exp_cha}
What are the characteristics of the distributional differences between benign and malicious samples in the latent spaces learned by an infected model? In this section, we attempt to answer this question. We first need to quantify the differences. Given an infected DNN $f_{\theta}$, let $Z_{ij}^T = \{z_i(x) | (x, y) \in T \land f_{\theta}(x) == j\}$ and $Z_{ij}^V = \{z_i(x') | (x', t) \in V \land f_{\theta}(x') == j\}$ denote the extracted representations of benign and malicious samples, which are classified into the same category $j$ by $f_{\theta}$. A distributional difference refers to the dissimilarity between the distributions of $Z_{ij}^T$ and $Z_{ij}^V$. In particular, since a well-trained backdoored model classifies almost all malicious samples to the target $t$, we mainly focus on the difference between $Z_{it}^T$ and $Z_{it}^V$. Given that the latent representations are high-dimensional, and their distributions are difficult to estimate explicitly, in this paper, MMD, ED, and SWD are adopted to measure the distributional differences between benign and malicious samples.

To comprehensively analyze the differences, we calculate the three metrics on the test data at multiple levels, i.e., $\m(Z_{it}^T, Z_{it}^V)$, where $\m$ is one of the three measures and $i = \{s_1, s_2, s_3\}$. For MMD, we use a Gaussian mixture kernel. To give intuitive comparisons, for each type of attack, we provide the intra-class distance and the minimal inter-class distance among clean samples as two baselines, that is, $\m(Z^T_{it}, Z^T_{it})$ and $\min_{j \ne t} \m(Z^T_{it}, Z^T_{ij})$. To avoid being influenced by the numerical scale, we use the relative ordering (ascending) of $\m(Z_{it}^T, Z_{it}^V)$ in $\m(Z_{it}^T, Z_{it}^V)$, $\m(Z^T_{it}, Z^T_{it})$, and $\min_{j \ne t} \m(Z^T_{it}, Z^T_{ij})$ as the indicator. TABLE \ref{tab:aro_rbt} presents Average Relative Ordering (ARO) over the three metrics, and the specific distance values are shown in Appendix \ref{app:qua_dif}.

\begin{table}[!ht] 
\centering 
\caption{ARO over MMD, ED and SWD at three levels.}
\begin{tabular}{c|c||c|c|c||c|c|c} 
\toprule
\multirow{2}{*}{Method}  & \multirow{2}{*}{Model} & \multicolumn{3}{c||}{CIFAR-10} & \multicolumn{3}{c}{CelebA} \\ 
                         &      & $s_1$ & $s_2$ & $s_3$ & $s_1$ & $s_2$ & $s_3$ \\ \midrule
\multirow{4}{*}{Patched} & V-11 & 2.67 & 3.00 & 2.67 & 3.00 & 3.00 & 3.00 \\
	                     & V-16 & 2.67 & 3.00 & 2.33 & 3.00 & 3.00 & 3.00 \\ 
	                     & R-18 & 2.67 & 3.00 & 2.67 & 3.00 & 3.00 & 3.00 \\ 
	                     & P-18 & 2.67 & 2.67 & 2.00 & 3.00 & 3.00 & 3.00 \\ \midrule
\multirow{4}{*}{Blended} & V-11 & 2.67 & 2.67 & 2.67 & 3.00 & 3.00 & 3.00 \\
	                     & V-16 & 2.67 & 2.67 & 2.33 & 3.00 & 3.00 & 3.00 \\ 
	                     & R-18 & 2.67 & 2.67 & 2.00 & 3.00 & 3.00 & 3.00 \\ 
	                     & P-18 & 3.00 & 2.67 & 3.00 & 3.00 & 3.00 & 3.00 \\ \midrule
\multirow{4}{*}{SIG}     & V-11 & 2.67 & 3.00 & 3.00 & 3.00 & 3.00 & 3.00 \\
	                     & V-16 & 3.00 & 3.00 & 2.33 & 3.00 & 3.00 & 3.00 \\ 
	                     & R-18 & 2.67 & 2.67 & 3.00 & 3.00 & 3.00 & 3.00 \\ 
	                     & P-18 & 3.00 & 3.00 & 2.00 & 3.00 & 3.00 & 3.00 \\ \midrule 
\multirow{4}{*}{Warped}  & V-11 & 3.00 & 3.00 & 3.00 & 3.00 & 3.00 & 3.00 \\
	                     & V-16 & 3.00 & 3.00 & 2.67 & 3.00 & 3.00 & 3.00 \\ 
	                     & R-18 & 2.67 & 2.67 & 2.33 & 3.00 & 3.00 & 3.00 \\ 
	                     & P-18 & 3.00 & 3.00 & 2.00 & 3.00 & 3.00 & 3.00 \\ \midrule            
\bottomrule	                 
\end{tabular}
\label{tab:aro_rbt}
\end{table}

Some characteristics can be identified:
\begin{itemize}
\item The distributional differences of multi-level representations are all large. We observe that for all types of backdoor attacks, all DNN architectures, and all the two datasets, there are substantial increases in the MMD, ED, and SWD values compared to the corresponding intra-class values at all three levels (see in Appendix \ref{app:qua_dif}). The results of ARO are more intuitive, where 69.8\% of the values are 3.0 and 95.8\% of the values are greater than 2.0. This means that in most cases, $\m(Z_{it}^T, Z_{it}^V)$ is greater than $\min_{j \ne t} \m(Z^T_{it}, Z^T_{ij})$. Both of the two observations provide strong evidence of this characteristic.
\item The distributional differences have some relationship to the dataset. In general, the infected models trained on CelebA have larger ARO values than the models trained on CIFAR-10.
\end{itemize}

The first characteristic is more inspiring. Since the distributional differences are all significant at multi-level representations, does it mean that the previous detection methods can be extended to multiple layers? We conduct experiments on this and find that this is indeed the case. The results are shown in Fig. \ref{fig:ac_p18}, Fig. \ref{fig:ss_p18}, and Fig. \ref{fig:sr_p18}, where the F1 scores of the three difference-based detection methods at three levels on the regularly trained backdoored models are all high. This provides guidance for us to design a proper reduction method.

\section{Method for Reducing the Distributional Differences} \label{sec:red_dif}
How to effectively reduce the distributional differences the infected models exhibit that defense methods could utilize? According to the analysis in Section \ref{sec:exp_cha}, we know that the differences in multi-level representations are all large. Therefore, we propose a method named ML-MMDR to reduce the differences by adding multi-level MMD regularization to the loss during the training of a backdoored model. The proposed method can be formulated as:
\begin{equation}
\begin{split}
\min_{\theta} & \frac{1}{|D|} \sum_{(x, y) \in D} L(f_{\theta}(x), y) + \frac{1}{|U|} \sum_{(x', t) \in U} L(f_{\theta}(x'), t) \\
&  + \lambda \cdot \frac{1}{|I|} \sum_{i \in I} \mmd(Z_{it}^D, Z_{it}^U) 
\end{split} \text{,} 
\label{equ:mlmmdr}
\end{equation}
where $I$ is a set of levels, and $\lambda$ is the hyperparameter that controls the constraint strength. The optimization goal consists of three parts, the first two of which are included in the regular backdoor training, as shown in \ref{equ:rbt}. We add the third item to reduce the distributional differences. In practice, we adopt the mini-batch ML-MMDR, and the procedure is presented in Algorithm \ref{alg:mlmmdr}.

\begin{algorithm}[t]
\caption{Mini-batch ML-MMDR} 
\label{alg:mlmmdr}
\SetAlgoLined
\KwIn{Benign training set $D$; malicious training set $U$; Learning rate $\eta$; Number of training epochs $N$; Constraint strength $\lambda$; Representation level set $I$; Target label $t$}
\KwOut{Model parameters $\theta$}
\BlankLine
Initialize parameters $\theta$\;
Create the concatenated training set $C = D \cup U$\;
\For{$n \leftarrow 1$ \KwTo $N$}{
	Shuffle the training set $C$\;
	\For{each mini-batch $(X, Y) \subset C$}{
		$(X_1, Y_1) = \{(x, y) \mid (x, y) \in (X, Y) \land (x, y) \in D\}$\;
		$(X_2, Y_2) = \{(x, y) \mid (x, y) \in (X, Y) \land (x, y) \in U\}$\;
		$(X_3, Y_3) = \{(x, y) \mid (x, y) \in (X, Y) \land (x, y) \in D \land y == t\}$\;
		$L_1 \leftarrow \frac{1}{|X_1|} \cdot L(f_{\theta}(X_1), Y_1)$\;
		$L_2 \leftarrow \frac{1}{|X_2|} \cdot L(f_{\theta}(X_2), Y_2)$\;
		$L_3 \leftarrow \frac{1}{|I|} \cdot \sum_{i \in I} \mmd(Z_i^{X_2}, Z_i^{X_3})$\;
		$L_t \leftarrow L_1 + L_2 + \lambda \cdot L_3$\;
		$\theta \leftarrow \theta - \eta \cdot \nabla_{\theta} L_t$\;
	}
}
\end{algorithm}	

Two settings of $I$ are considered, including $I = \{s_3\}$ and $I = \{s_1, s_2, s_3\}$, which correspond to constraining the features of the last hidden layer (SL-MMDR) and all three levels (ML-MMDR), respectively. We set $\lambda = \{0.0, 0.1, 0.2, 0.3\}$, where $\lambda = 0.0$ stands for Regular Backdoor Training (RBT). For convenience, we use the \{A,B,C\} models to denote the infected models trained on dataset A with architecture B and attack method C. The experimental results on the \{P-18\} models are shown in Fig. \ref{fig:mlmmdr_p18}. The results on other architectures are similar and are shown in Appendix \ref{app:res_mlmmdr}.

\begin{figure*}[!ht]
\centering
\subfloat[CIFAR-10, Patched]{\includegraphics[width=8.8cm]{\impath/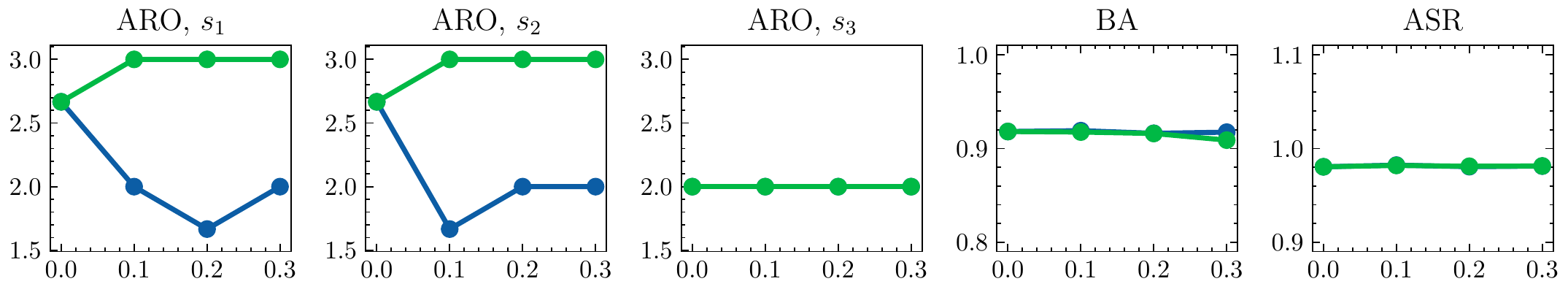}} \hfill
\subfloat[CIFAR-10, Blended]{\includegraphics[width=8.8cm]{\impath/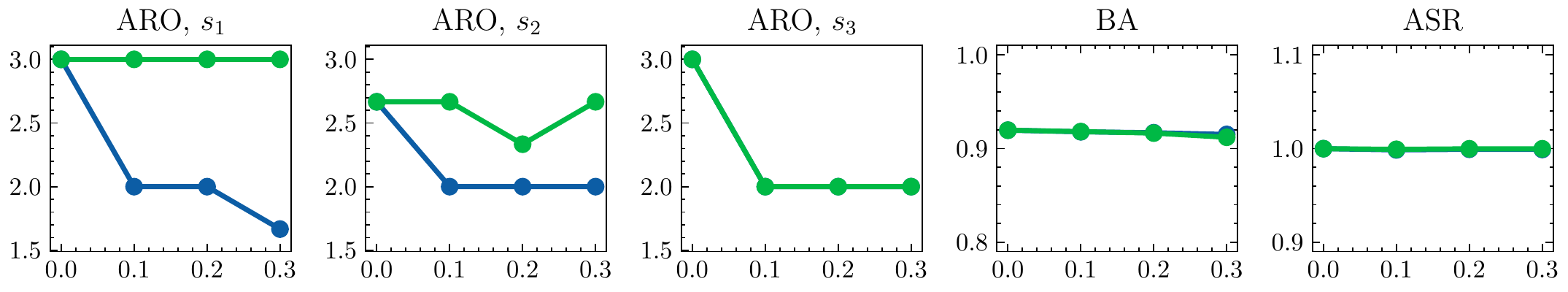}} \\
\subfloat[CIFAR-10, SIG]{\includegraphics[width=8.8cm]{\impath/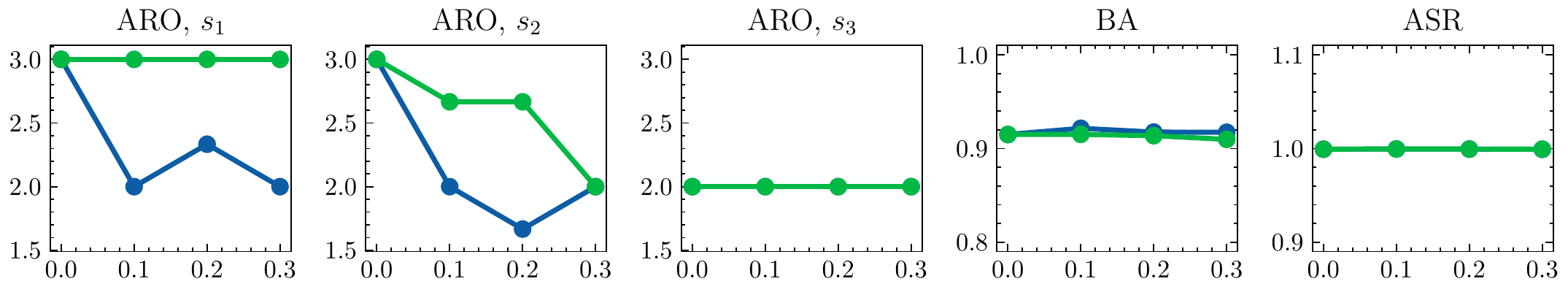}} \hfill 
\subfloat[CIFAR-10, Warped]{\includegraphics[width=8.8cm]{\impath/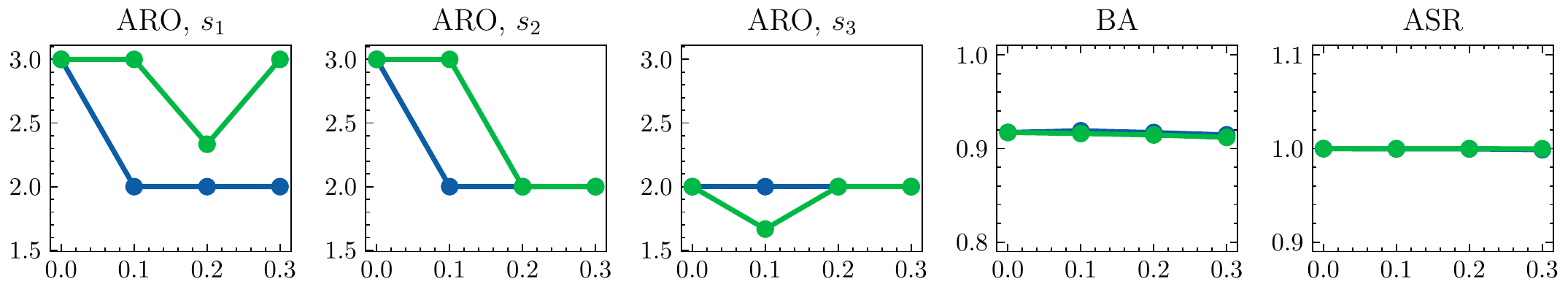}} \\
\subfloat[CelebA, Patched]{\includegraphics[width=8.8cm]{\impath/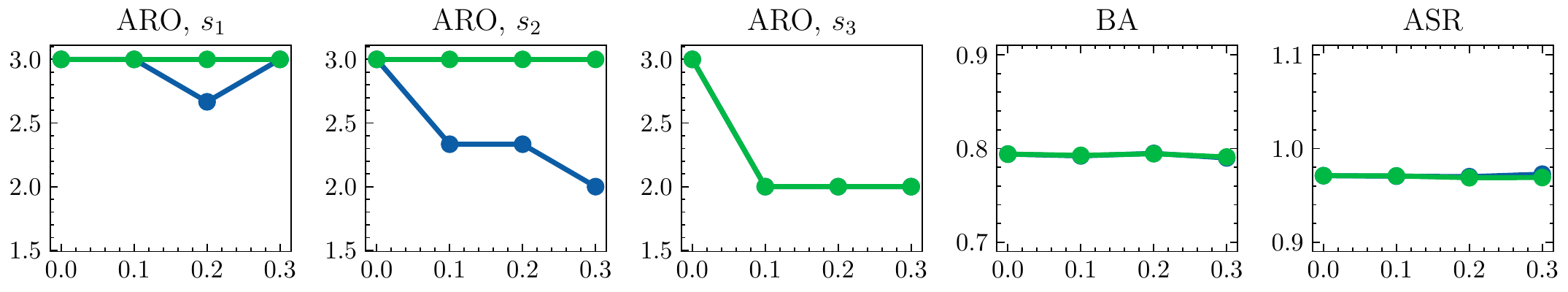}} \hfill
\subfloat[CelebA, Blended]{\includegraphics[width=8.8cm]{\impath/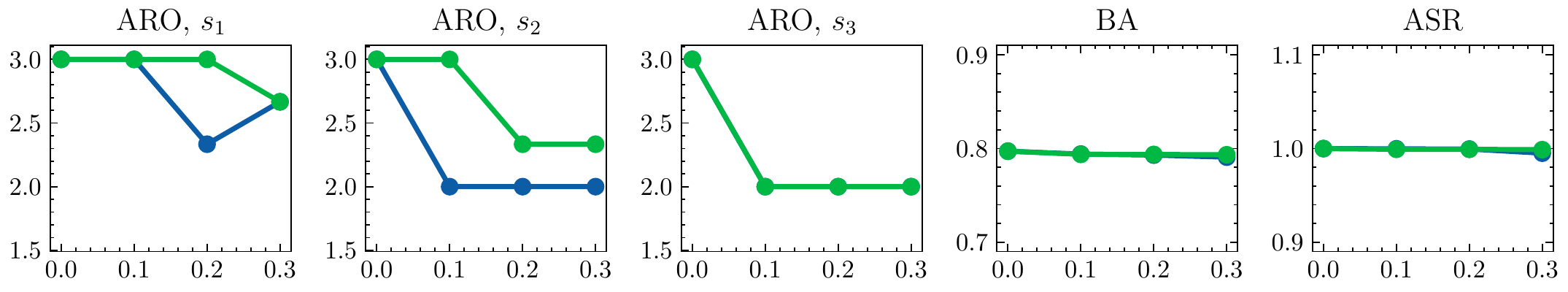}} \\
\subfloat[CelebA, SIG]{\includegraphics[width=8.8cm]{\impath/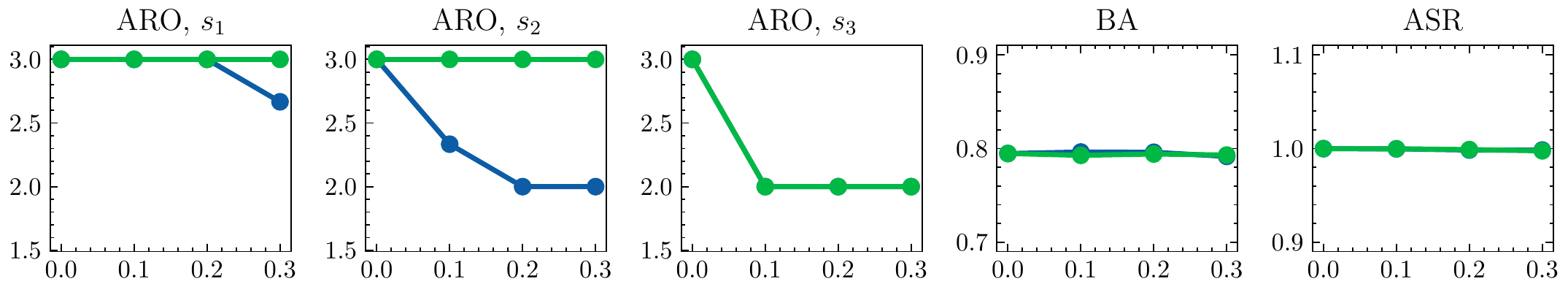}} \hfill 
\subfloat[CelebA, Warped]{\includegraphics[width=8.8cm]{\impath/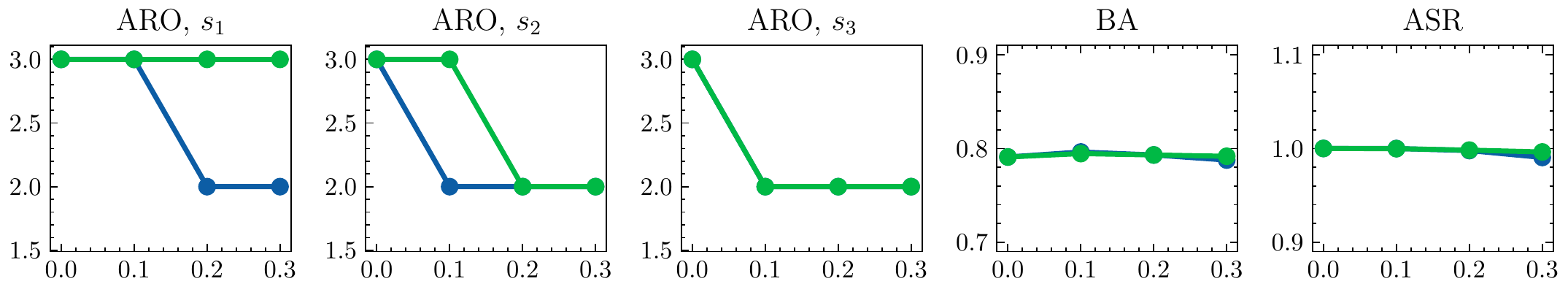}} \\
\subfloat{\includegraphics[height=0.5cm]{\impath/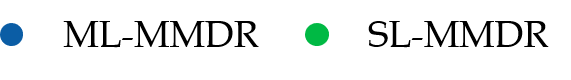}}
\caption{Results of SL-MMDR and ML-MMDR on the \{P-18\} models with different $\lambda$. X-axis: the value of $\lambda$. Y-axis: the value of each indicator.}
\label{fig:mlmmdr_p18} 
\end{figure*}

Some observations are summarized as follows:
\begin{itemize}
\item The proposed ML-MMDR can significantly reduce the distributional differences at multi-level representations without harming the attack power. For example, when $\lambda = 0.3$, the average BA and ASR over the models trained with ML-MMDR decrease from 0.907 and 0.993 to 0.902 and 0.992, respectively, and the average ARO decreases from 2.792, 2.854, and 2.5 to 1.979, 1.979, and 2.0 at $s_1$, $s_2$, and $s_3$ for CIFAR-10, respectively, compared to the models trained with RBT ($\lambda = 0.0$). 2.0 is basically the minimum value that ARO can achieve.
\item Constraining only the features of the last hidden layer, i.e., SL-MMDR, causes the features of other intermediate layers to have large differences still. For example, for $\lambda = 0.3$, the average ARO over the models trained with SL-MMDR is 2.833, 2.25, and 2.0 at $s_1$, $s_2$, and $s_3$ for CIFAR-10. It can be seen that the difference at the first level is even increased by SL-MMDR, which preserves the possibility of using these representations for defense. These results demonstrate that the methods of constraining only the last hidden layer \cite{tan2020bypassing,doan2021backdoor,ren2021simtrojan} are not sufficient.
\end{itemize}

\begin{figure*}[!ht]
\centering
\subfloat[RBT]{\includegraphics[width=8.8cm]{\impath/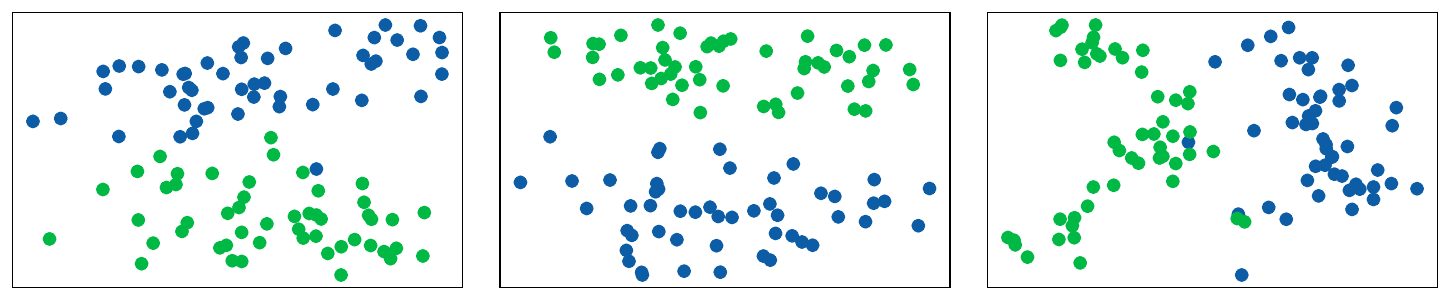}} \hfill
\subfloat[ML-MMDR, $\lambda=0.3$]{\includegraphics[width=8.8cm]{\impath/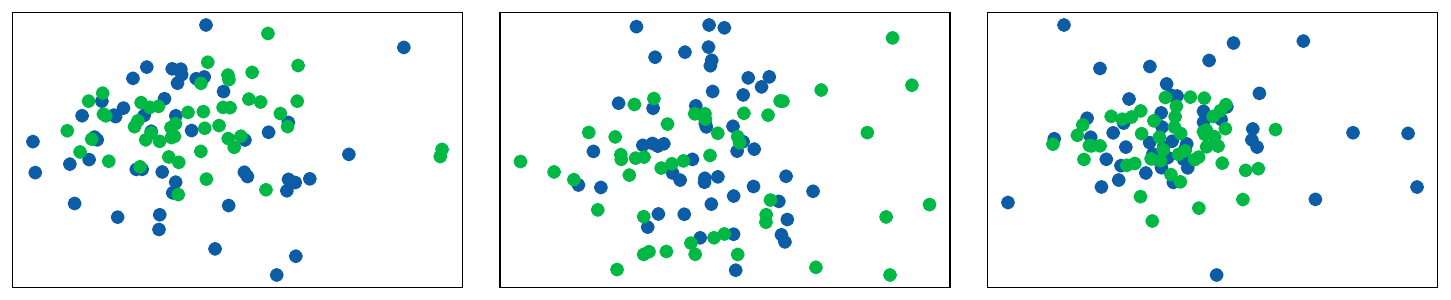}} \\
\subfloat[SL-MMDR, $\lambda=0.3$]{\includegraphics[width=8.8cm]{\impath/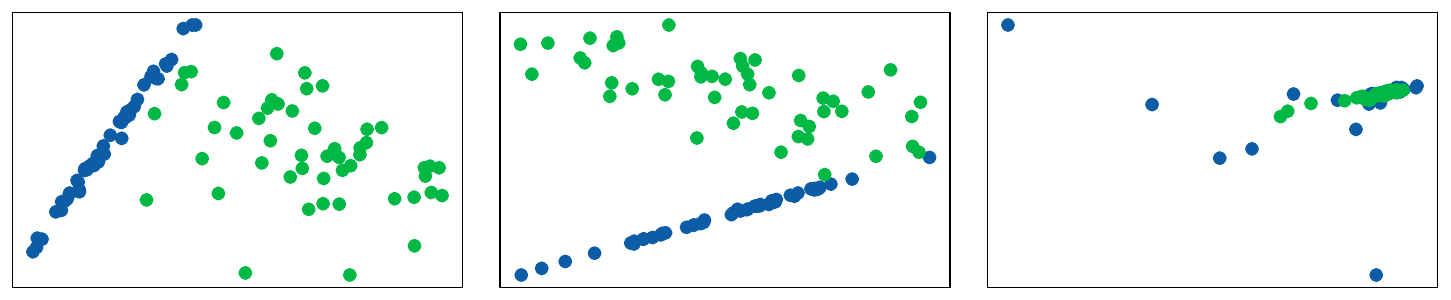}} \\
\subfloat{\includegraphics[height=0.5cm]{\impath/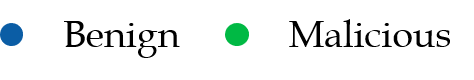}}
\caption{Visualization examples of the features extracted from the \{CIFAR-10, P-18, Patched\} models.}
\label{fig:sdrm_vis_c10_p18_p} 
\end{figure*}

To show the differences more intuitively, the representations are visualized using the dimensionality reduction technique, as shown in Fig. \ref{fig:sdrm_vis_c10_p18_p}. In the model trained with RBT, the representations of benign and malicious inputs can obviously be divided into two groups. For the model trained with ML-MMDR, these two types of representations almost overlap at all three levels. The first two levels of representations are still distinguishable for the model trained with SL-MMDR. The visualization results support our observations. 

\section{Empirical Study on the Effectiveness of the Method} \label{sec:exp_def}
Since the proposed ML-MMDR is feasible in reducing the distributional differences effectively without harming the attack intensity, in this section, we investigate what impact this reduction has on difference-based defense methods. Four typical methods, including Activation Clustering (AC) \cite{chen2019detecting}, Spectral Signatures (SS) \cite{tran2018spectral}, Subspace Reconstruction (SR) \cite{javaheripi2020cleann}, and Neural Cleanse (NC) \cite{wang2019neural}, are selected to testify our regularization. The first three methods are difference-based detection methods, and the last method is not. The backdoored models trained with RBT, ML-MMDR, and SL-MMDR are tested, where $\lambda = 0.3$.

\subsection{Results of AC, SS, and SR}
AC, SS, and SR are all difference-based detection methods that distinguish malicious samples from benign ones by using the latent representations extracted by the infected model. In our experiments, these methods are performed on the representations of three levels, i.e., $s_1$, $s_2$, and $s_3$. Because the effects of the three methods are affected by the number of samples $N$ and the ratio of malicious inputs to benign inputs $r'$, we set up four combinations for each defense method. The detection performance of AC, SS, and SR is measured by F1 score, and the results on the \{P-18\} models are shown in this section. The results on other architectures are similar and are presented in Appendix \ref{app:per_def}. Below, we first show the results of each defense method separately before making a unified summary.

\begin{figure*}[!ht]
\centering
\subfloat[CIFAR-10, Patched]{\includegraphics[width=8.8cm]{\impath/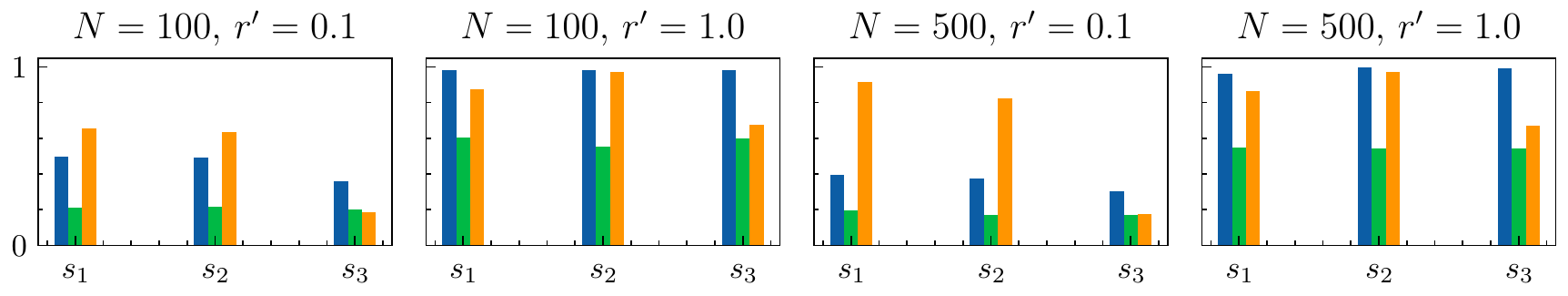}} \hfill
\subfloat[CIFAR-10, Blended]{\includegraphics[width=8.8cm]{\impath/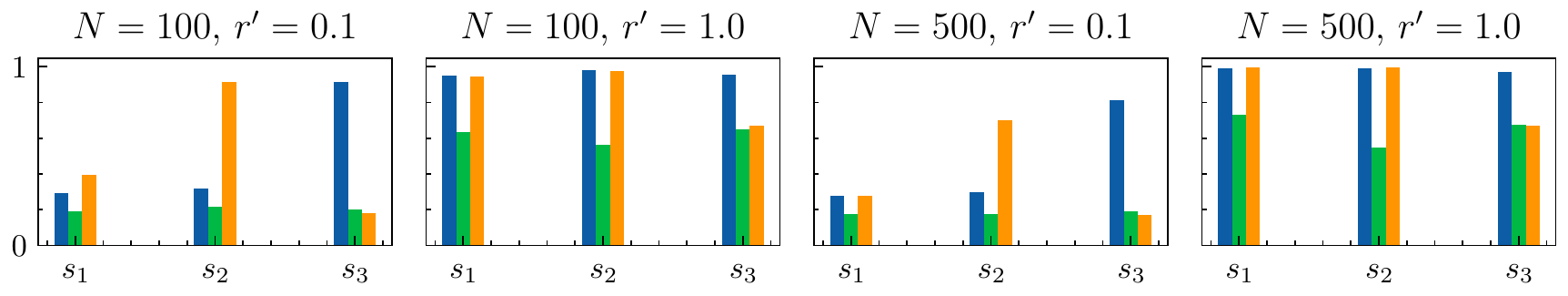}} \\
\subfloat[CIFAR-10, SIG]{\includegraphics[width=8.8cm]{\impath/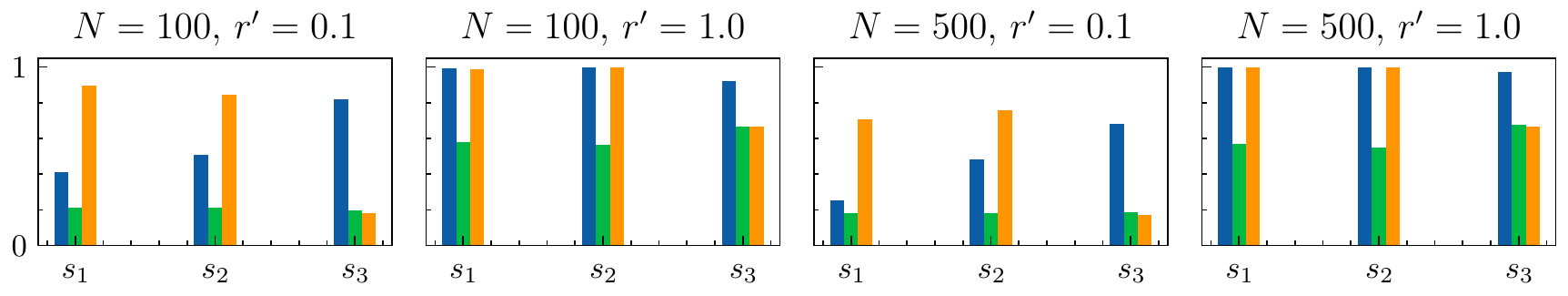}} \hfill 
\subfloat[CIFAR-10, Warped]{\includegraphics[width=8.8cm]{\impath/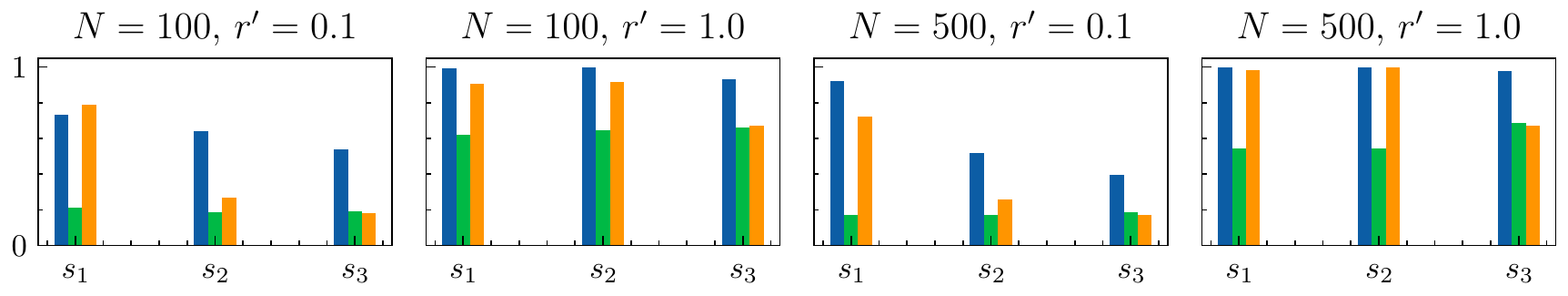}} \\
\subfloat[CelebA, Patched]{\includegraphics[width=8.8cm]{\impath/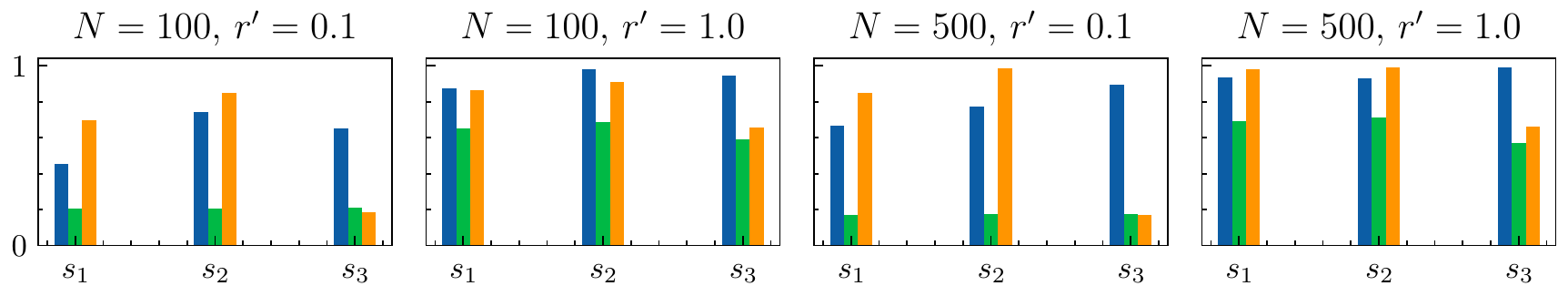}} \hfill
\subfloat[CelebA, Blended]{\includegraphics[width=8.8cm]{\impath/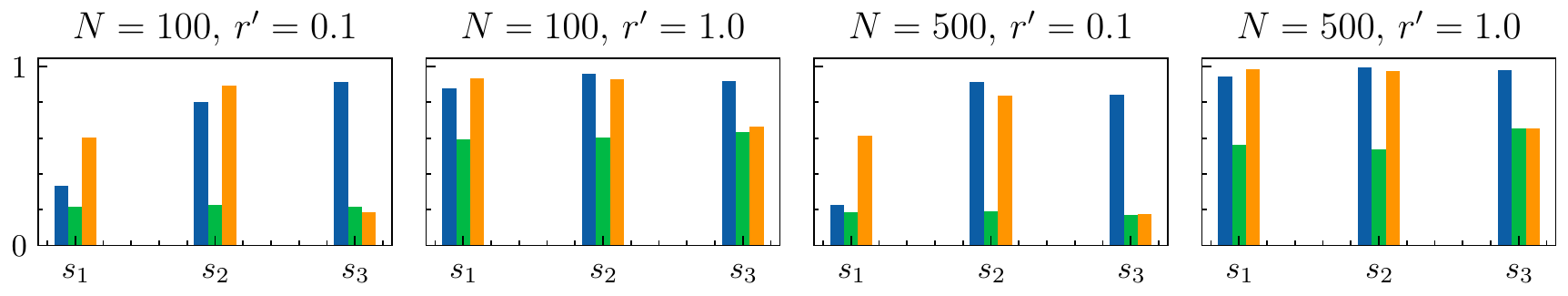}} \\
\subfloat[CelebA, SIG]{\includegraphics[width=8.8cm]{\impath/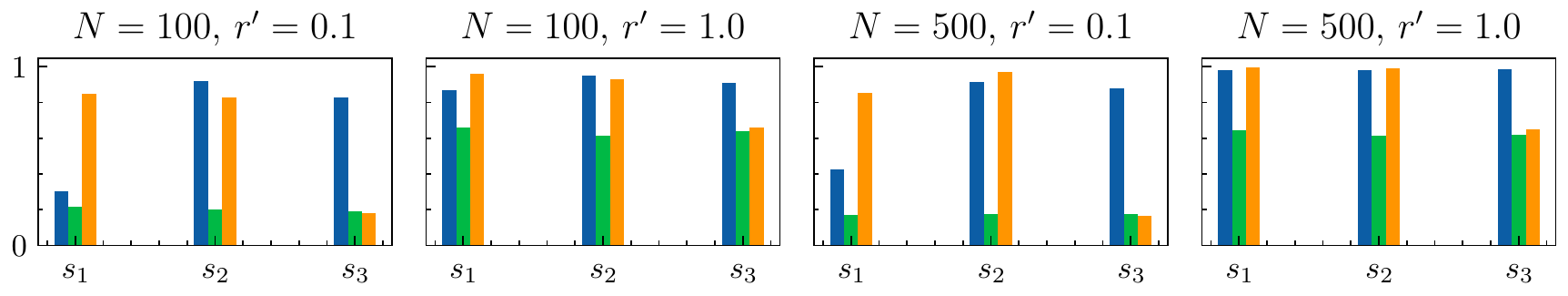}} \hfill 
\subfloat[CelebA, Warped]{\includegraphics[width=8.8cm]{\impath/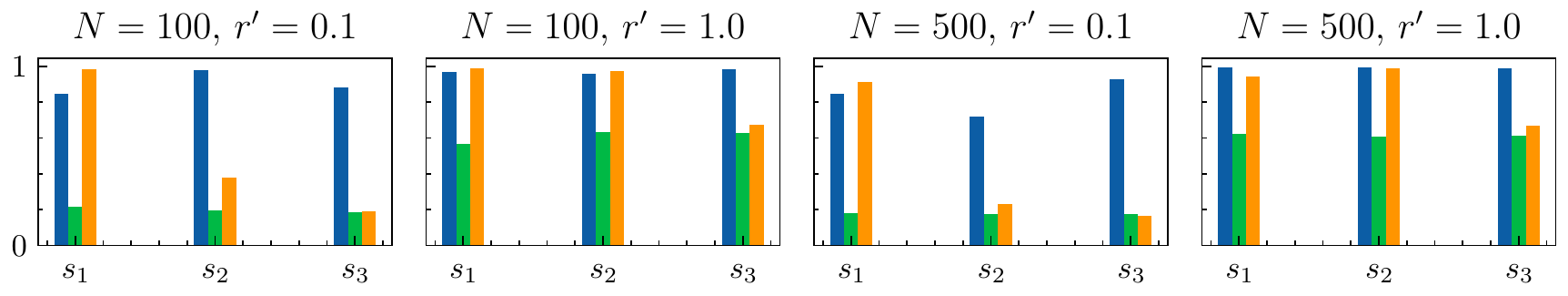}} \\
\subfloat{\includegraphics[height=0.5cm]{\impath/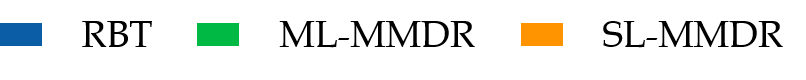}}
\caption{F1 scores of AC on the \{P-18\} models. X-axis: the level of features. Y-axis: the value of $F1$.}
\label{fig:ac_p18} 
\end{figure*}

\noindent\textbf{Activation Clustering.} As suggested in \cite{chen2019detecting}, we first adopt independent component analysis to reduce the dimensionality of the latent representations of inputs to obtain the vectors of length 20. Then we use k-means to cluster the reduced representations into two groups. The results of AC on the \{P-18\} models are shown in Fig. \ref{fig:ac_p18}. 

\begin{figure*}[!h]
\centering
\subfloat[CIFAR-10, Patched]{\includegraphics[width=8.8cm]{\impath/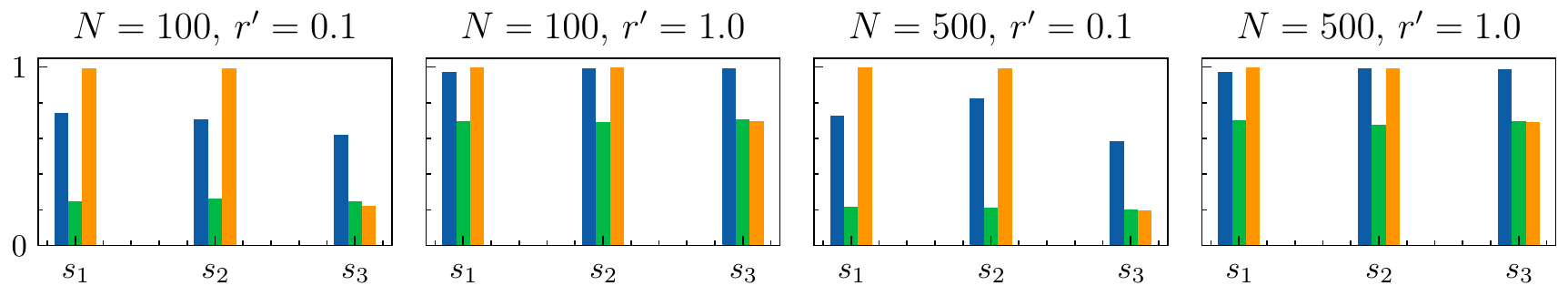}} \hfill
\subfloat[CIFAR-10, Blended]{\includegraphics[width=8.8cm]{\impath/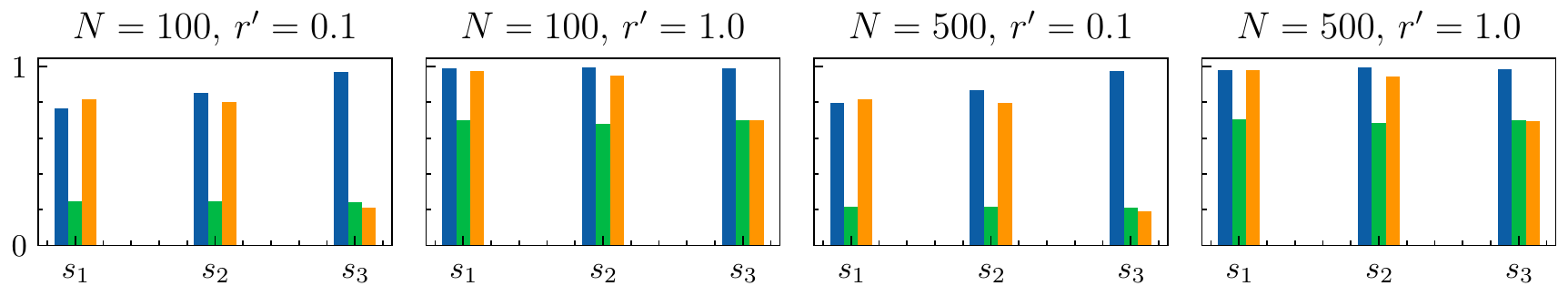}} \\
\subfloat[CIFAR-10, SIG]{\includegraphics[width=8.8cm]{\impath/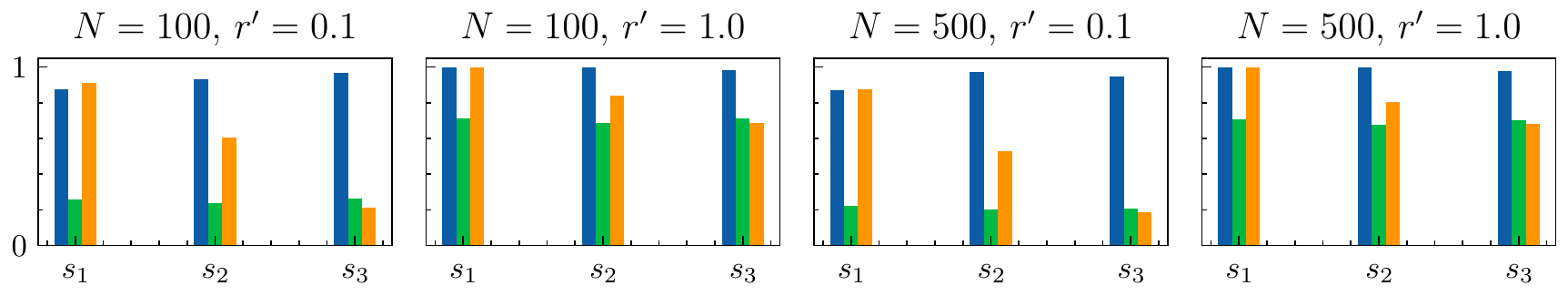}} \hfill 
\subfloat[CIFAR-10, Warped]{\includegraphics[width=8.8cm]{\impath/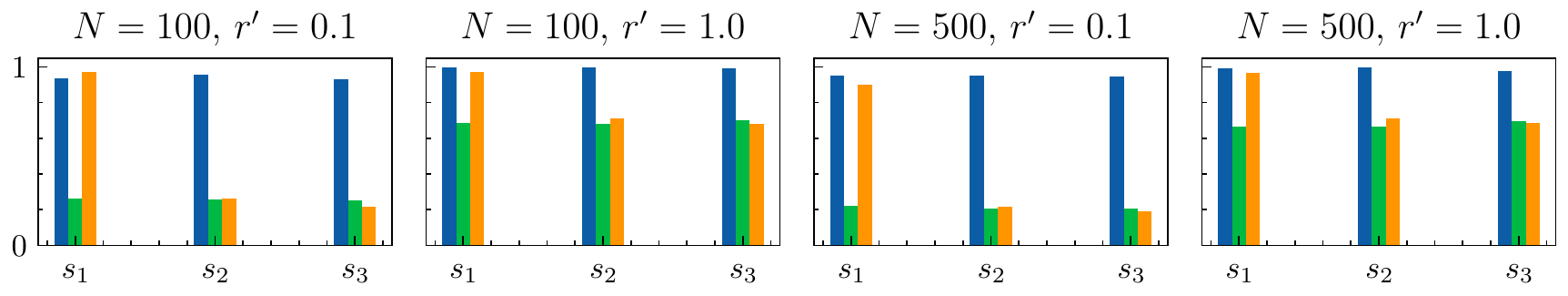}} \\
\subfloat[CelebA, Patched]{\includegraphics[width=8.8cm]{\impath/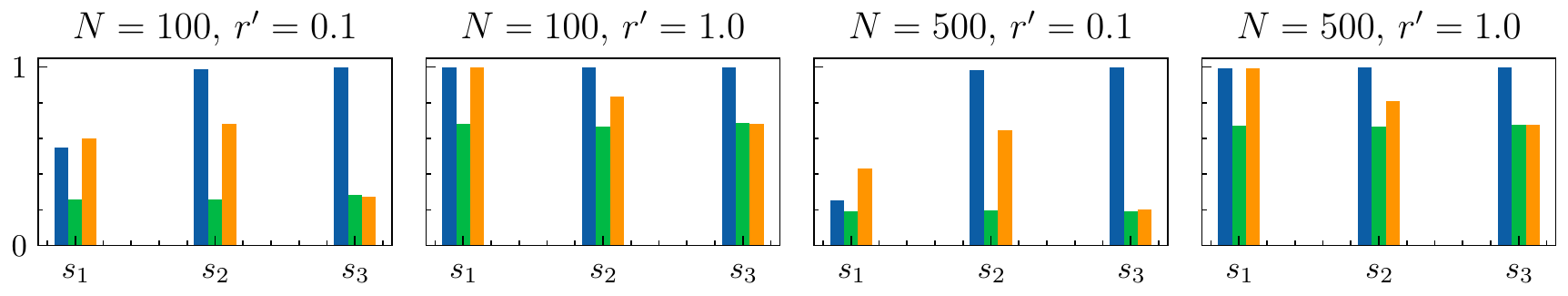}} \hfill
\subfloat[CelebA, Blended]{\includegraphics[width=8.8cm]{\impath/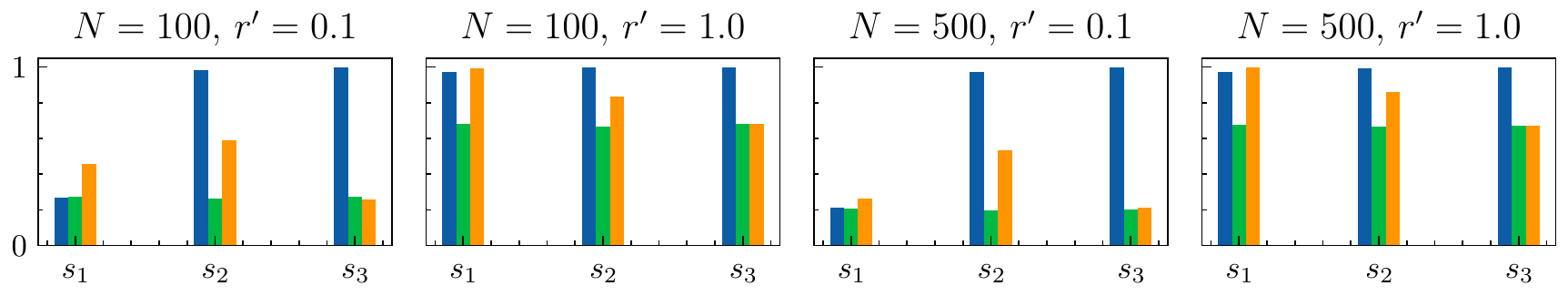}} \\
\subfloat[CelebA, SIG]{\includegraphics[width=8.8cm]{\impath/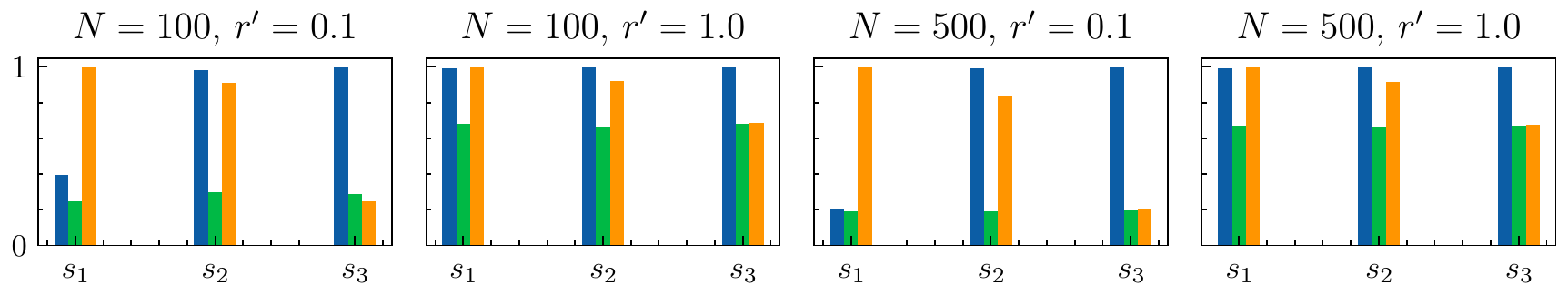}} \hfill 
\subfloat[CelebA, Warped]{\includegraphics[width=8.8cm]{\impath/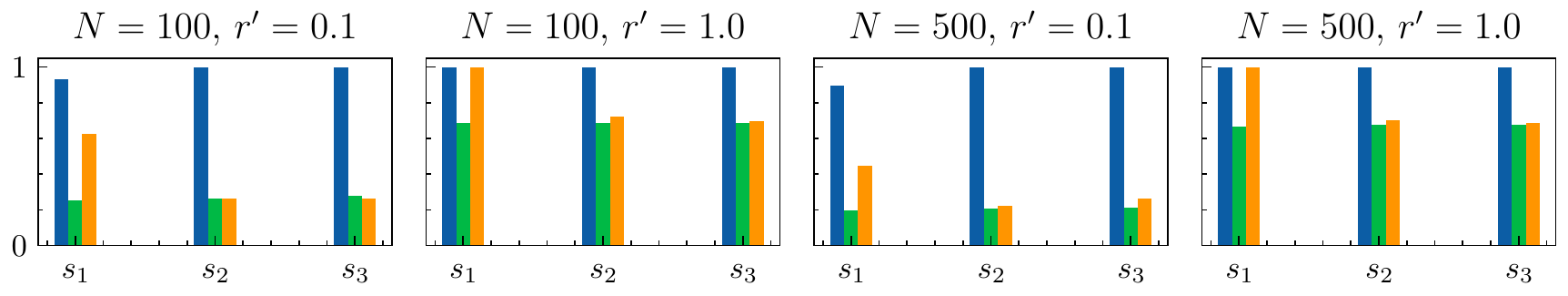}} \\
\subfloat{\includegraphics[height=0.5cm]{\impath/legend_defense.png}}
\caption{F1 scores of SS on the \{P-18\} models. X-axis: the level of features. Y-axis: the value of $F1$.}
\label{fig:ss_p18} 
\end{figure*}

\begin{figure*}[!ht]
\centering
\subfloat[CIFAR-10, Patched]{\includegraphics[width=8.8cm]{\impath/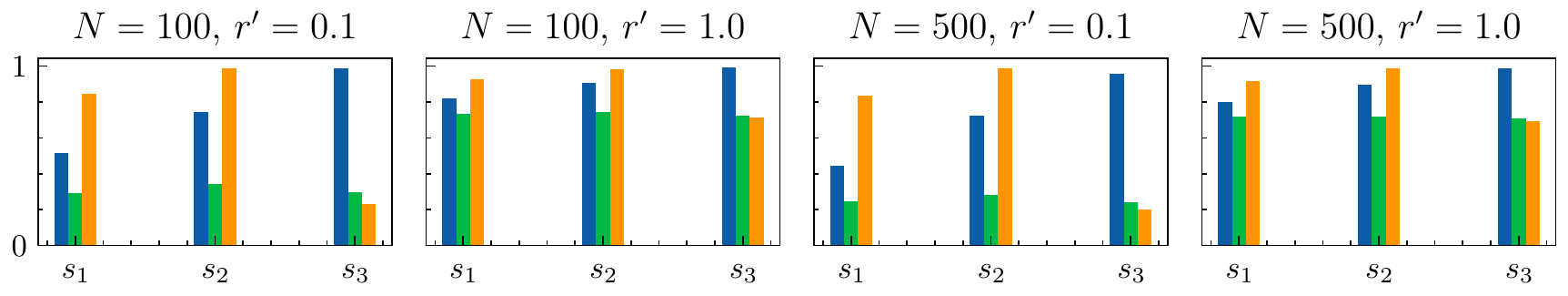}} \hfill
\subfloat[CIFAR-10, Blended]{\includegraphics[width=8.8cm]{\impath/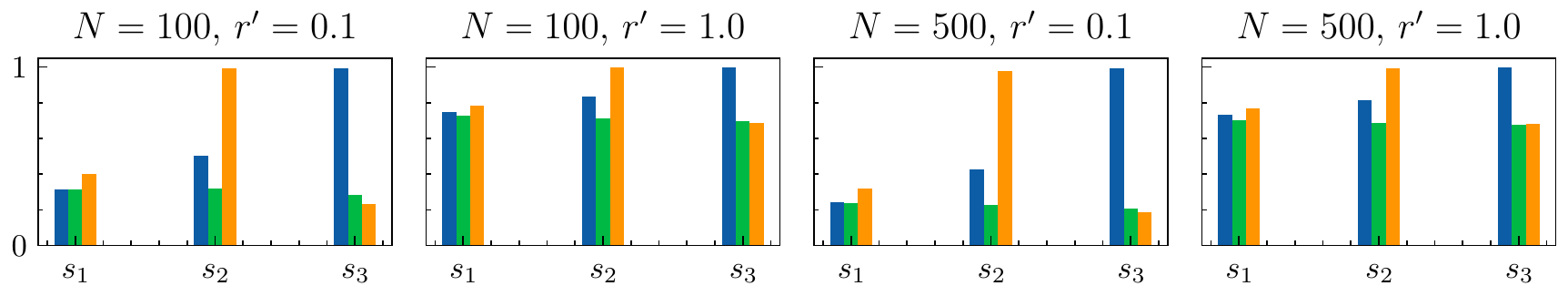}} \\
\subfloat[CIFAR-10, SIG]{\includegraphics[width=8.8cm]{\impath/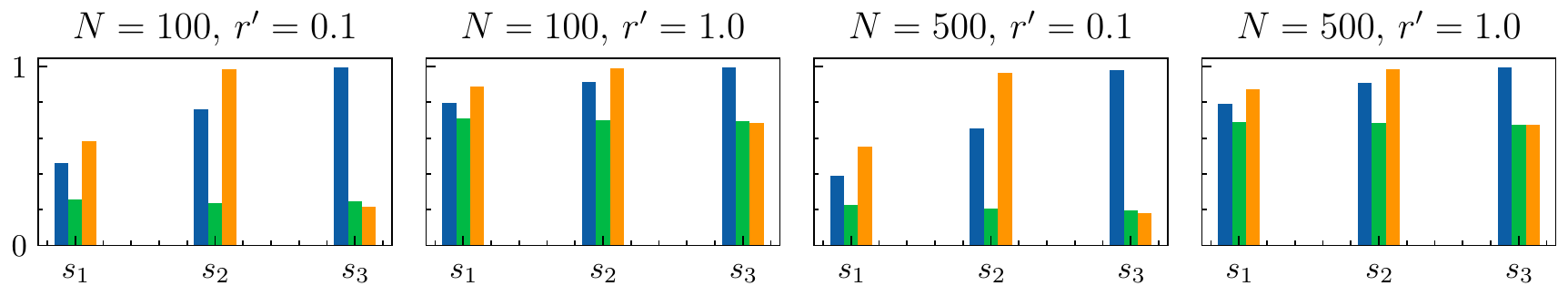}} \hfill 
\subfloat[CIFAR-10, Warped]{\includegraphics[width=8.8cm]{\impath/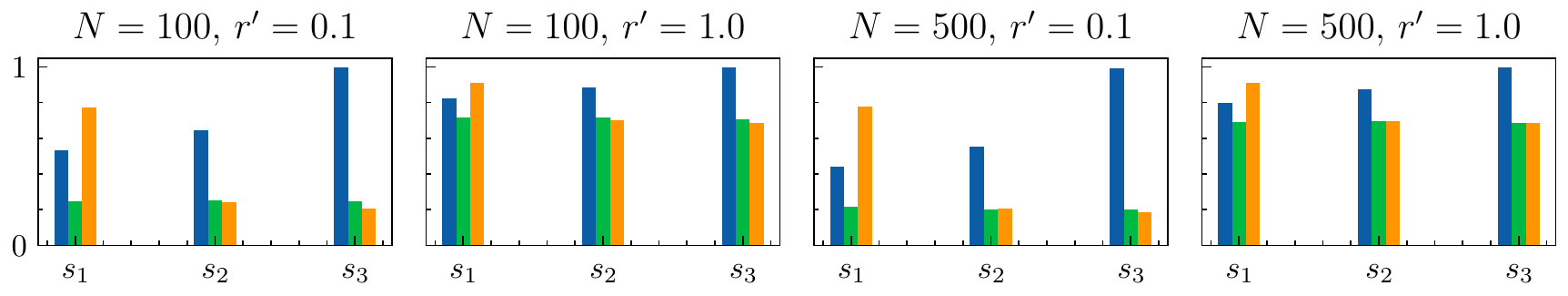}} \\
\subfloat[CelebA, Patched]{\includegraphics[width=8.8cm]{\impath/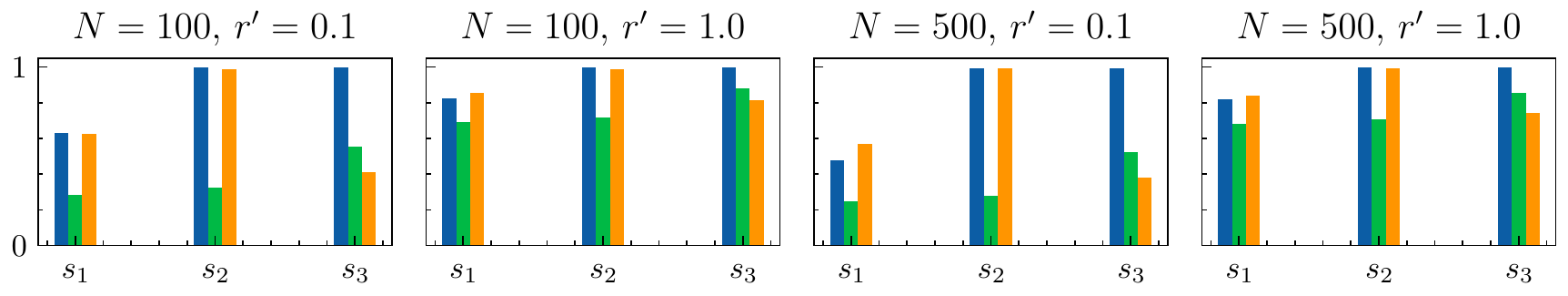}} \hfill
\subfloat[CelebA, Blended]{\includegraphics[width=8.8cm]{\impath/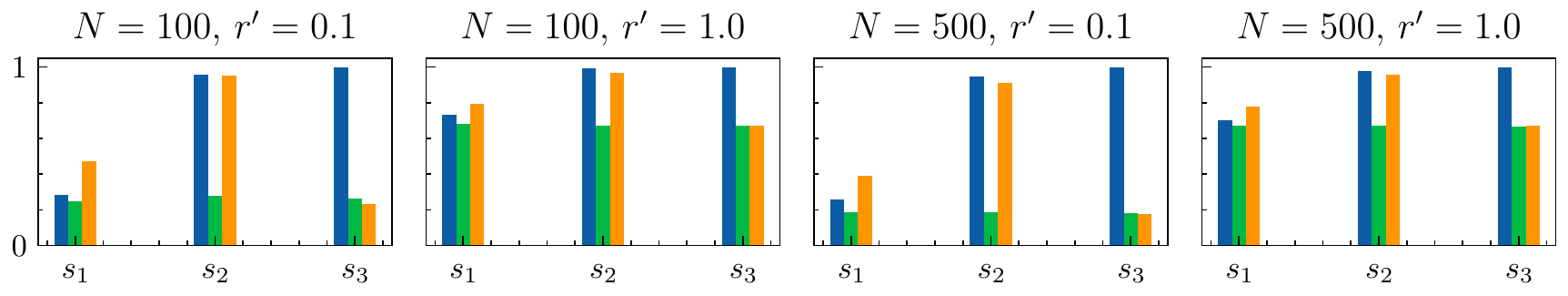}} \\
\subfloat[CelebA, SIG]{\includegraphics[width=8.8cm]{\impath/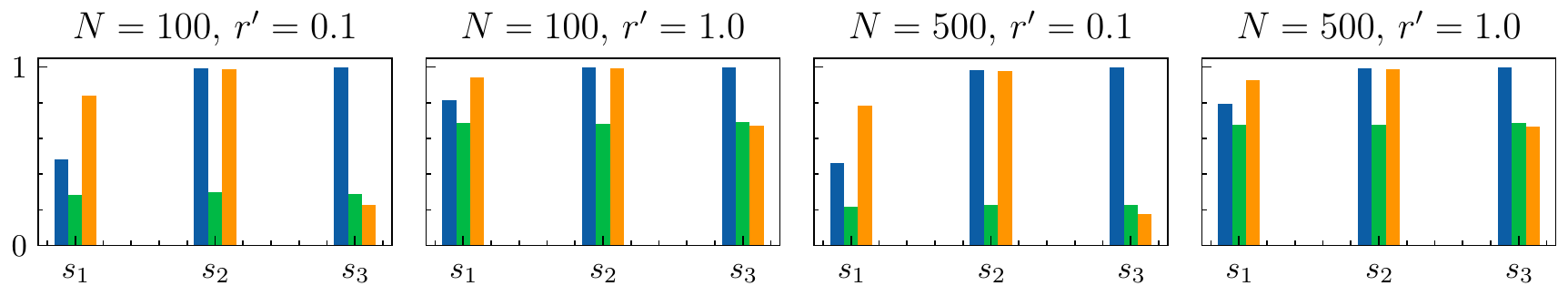}} \hfill 
\subfloat[CelebA, Warped]{\includegraphics[width=8.8cm]{\impath/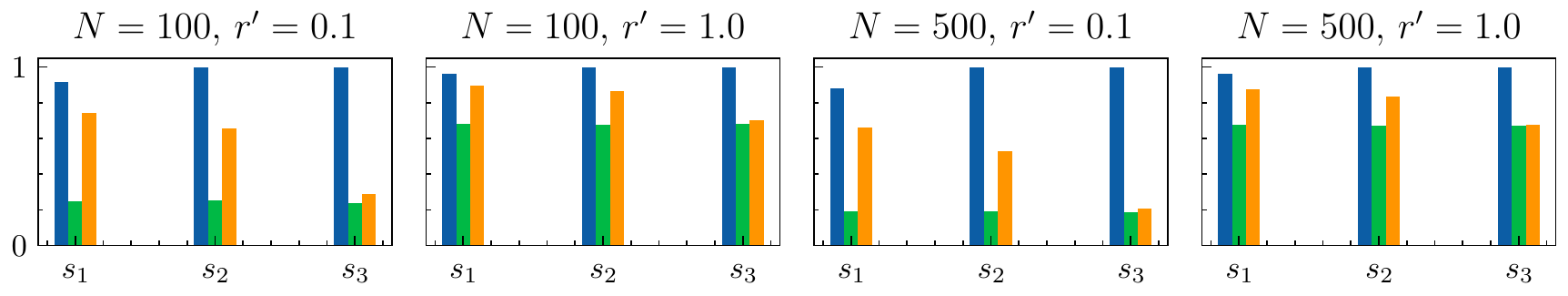}} \\
\subfloat{\includegraphics[height=0.5cm]{\impath/legend_defense.png}}
\caption{F1 scores of SR on the \{P-18\} models. X-axis: the level of features. Y-axis: the value of $F1$.}
\label{fig:sr_p18} 
\end{figure*}

\noindent\textbf{Spectral Signatures.} Following the same steps in \cite{tran2018spectral}, we first take singular value decomposition of the latent representations, and then use the top right vector to compute an outlier score for each input. The results of SS on the \{P-18\} models are shown in Fig. \ref{fig:ss_p18}.

\noindent\textbf{Subspace Reconstruction.} We choose 200 validated data and construct a subspace that can restore 90\% of the energy of these samples. Subsequently, for the representation of each input, we perform the projection and reconstruction steps with the learned subspace and compute the $l_2$ norm as the reconstruction loss for that sample. The max F1 score for each model is calculated by adjusting the threshold, and the results are shown in Fig. \ref{fig:sr_p18}.

\noindent\textbf{Summary.} The experimental results of the above three defense methods are similar and are summarized as follows:
\begin{itemize}
\item The representations at $s_1$ and $s_2$ levels can be used by these methods to detect malicious samples. For example, when $N=100$ and $r'=1.0$, the average F1 scores of AC at the three levels for RBT are 0.978, 0.991, and 0.948 on the \{CIFAR-10, P-18\} models, and 0.898, 0.963, and 0.939 on the \{CelebA, P-18\} models. This confirms the conclusion in Section \ref{sec:exp_cha}.
\item The performance of difference-based detection methods is dropped on the infected models trained with ML-MMDR. For example, when $N=500$ and $r'=1.0$, the average F1 values of AC at the three levels for ML-MMDR drop from 0.986, 0.997, 0.979 to 0.598, 0.546, 0.645 on the \{CIFAR-10, P-18\} models, and from 0.965, 0.975, 0.988 to 0.630, 0.618, 0.613 on the \{CelebA, P-18\} models. The results of SS and SR are similar.
\item The models trained with SL-MMDR can still be utilized by these defense methods to detect malicious samples. For example, when $N=500$ and $r'=1.0$, the average F1 scores of AC, SS, and SR at $s_0$ level for SL-MMDR are 0.962, 0.985, and 0.867 on the \{CIFAR-10, P-18\} models, and 0.977, 0.998, and 0.856 on the \{CelebA, P-18\} models. Sometimes, the values are higher than on the models trained with RBT. This indicates that constraining only the activations of the last hidden layer, which is adopted in the previous reduction methods \cite{tan2020bypassing,doan2021backdoor,ren2021simtrojan}, is not enough for bypassing backdoor detection algorithms. 
\end{itemize}

\subsection{Results of NC}
NC consists of two steps, i.e., trigger synthesis and neuron pruning, to complete the detection and mitigation of the backdoor. We conduct experiments on the above two steps separately.

\begin{figure*}[!ht]
\centering
\subfloat[CIFAR-10]{\includegraphics[width=8.8cm]{\impath/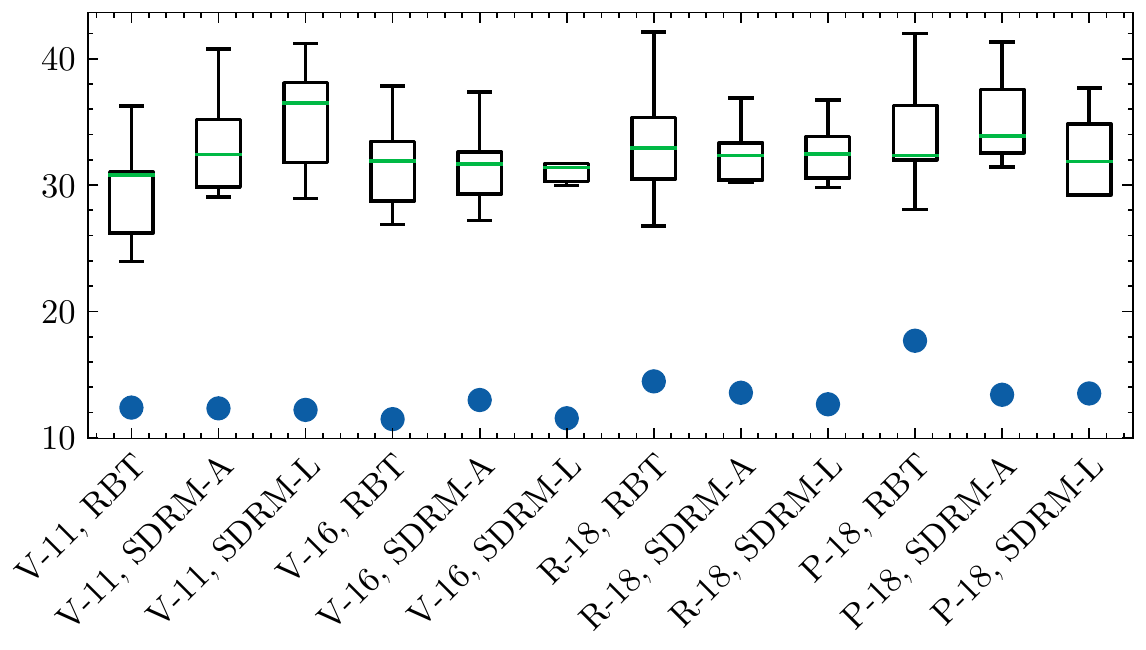}} \hfill
\subfloat[CelebA]{\includegraphics[width=8.8cm]{\impath/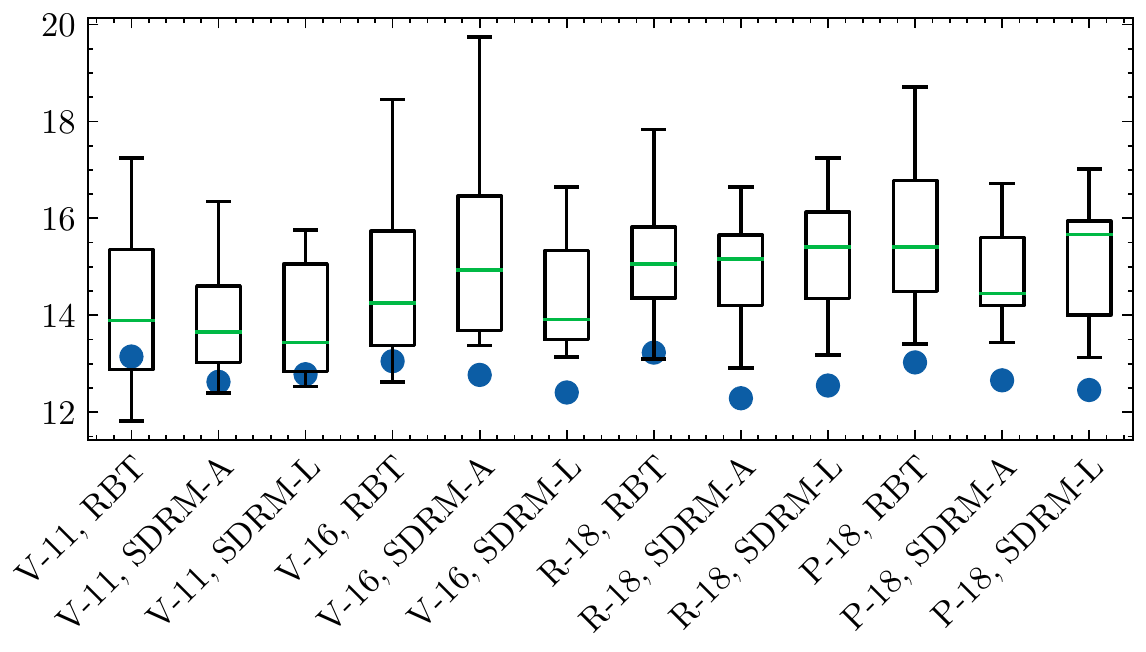}} \\
\subfloat{\includegraphics[height=0.5cm]{\impath/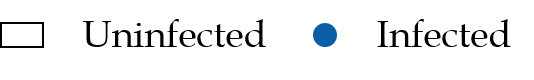}}
\caption{$l_1$ norm of triggers for infected and uninfected labels on the \{Patched\} models. X-axis: the model architecture and the backdoor training method. Y-axis: the $l_1$ norm of trigger.}
\label{fig:nc_p} 
\end{figure*}

\begin{figure}[!ht]
\centering
\includegraphics[width=8.0cm]{\impath/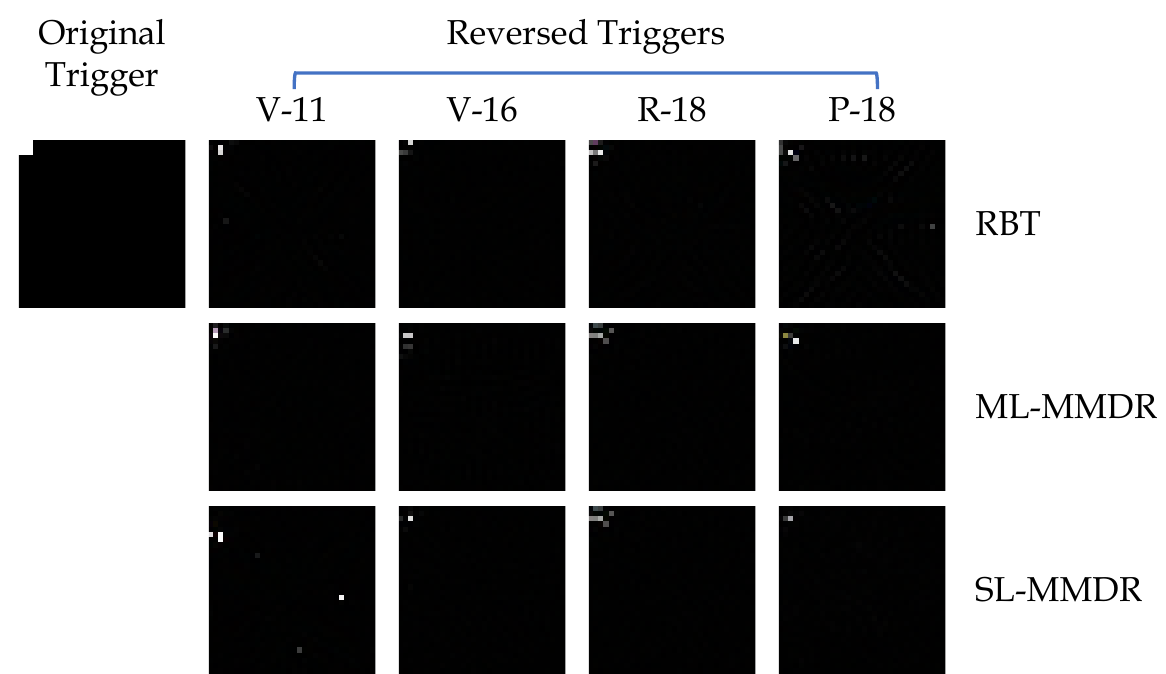}
\caption{Reversed triggers on the \{CIFAR-10, Patched\} models.}
\label{fig:rev_tri}
\end{figure}

\noindent\textbf{Trigger Synthesis.} Wang et al. \cite{wang2019neural} defined a generic form of backdoor sample generation with a 3D trigger pattern and a 2D location mask. Then they formulated an optimization problem to reverse a pattern and a location for each category. The final synthetic trigger is selected by calculating the $l_1$ norm of all candidate masks. Because the form defined above is mainly used to reverse the patching-based attack \cite{gu2017badnets}, we conduct experiments on the \{Patched\} models. The results are shown in Fig. \ref{fig:nc_p}, and some of the reversed triggers on the \{CIFAR-10, Patched\} models are shown in Fig. \ref{fig:rev_tri}.

As see, whether it is from the distinguishability of the $l_1$ norm or from the intuitive feeling, the infected models trained with ML-MMDR can still be used to reverse the triggers. This indicates that the trigger synthesis relies on a different principle from the difference-based methods to defend against backdoor attacks. However, this method seems to be easily affected by the dataset, such as the performance on the \{CelebA, Patched\} models is not as good as on the \{CIFAR-10, Patched\} models.

\begin{figure*}[!ht]
\centering
\subfloat[CIFAR-10, Patched]{\includegraphics[width=8.8cm]{\impath/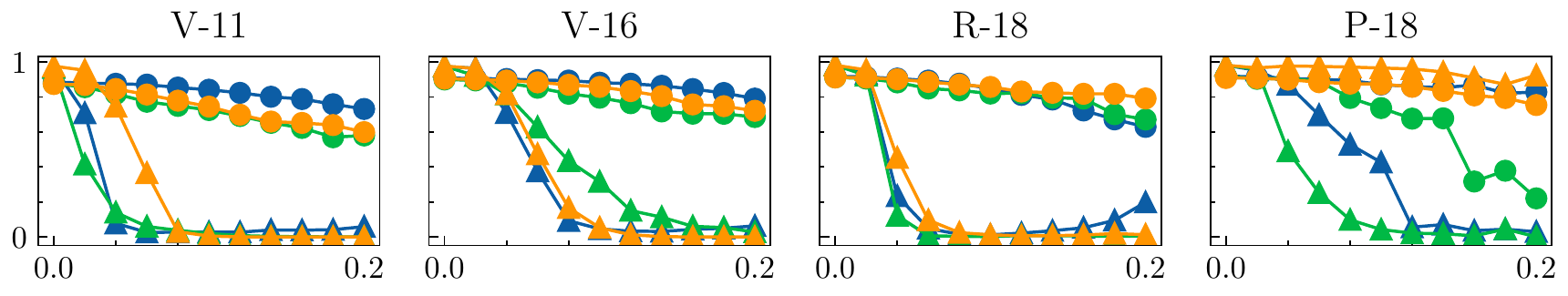}} \hfill
\subfloat[CIFAR-10, Blended]{\includegraphics[width=8.8cm]{\impath/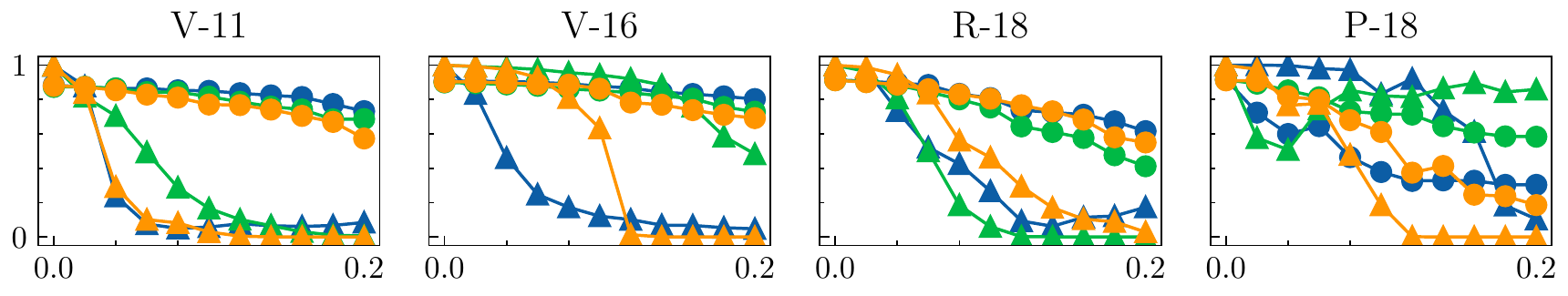}} \\
\subfloat[CIFAR-10, SIG]{\includegraphics[width=8.8cm]{\impath/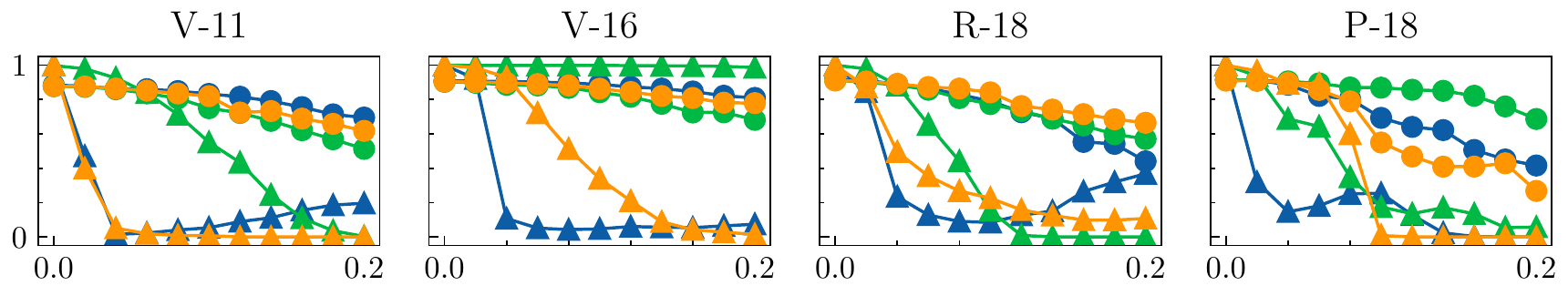}} \hfill 
\subfloat[CIFAR-10, Warped]{\includegraphics[width=8.8cm]{\impath/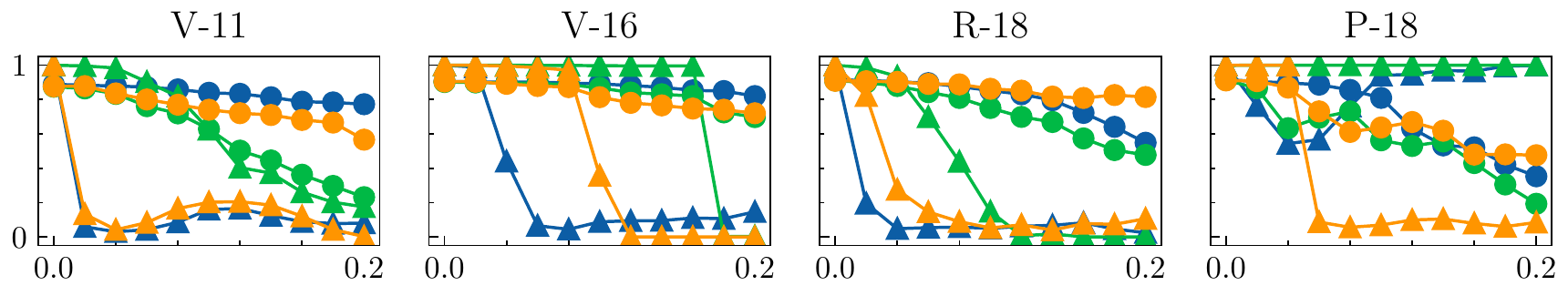}} \\
\subfloat[CelebA, Patched]{\includegraphics[width=8.8cm]{\impath/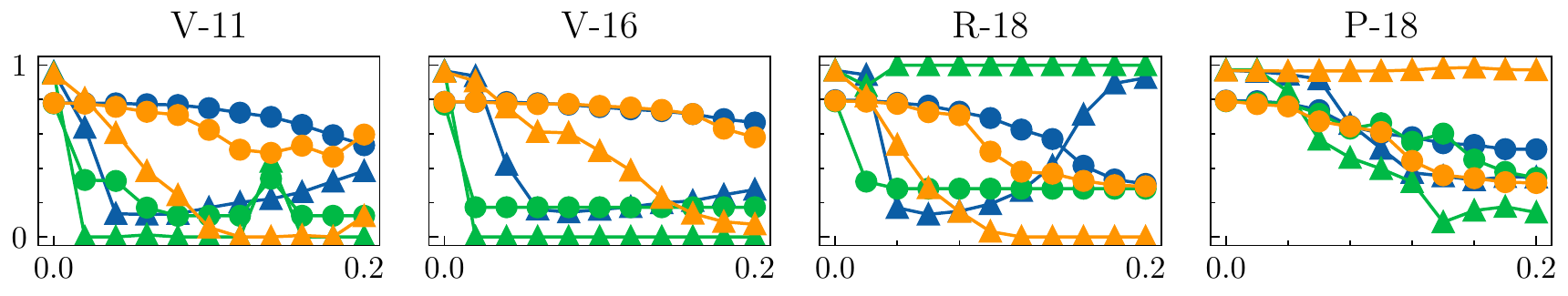}} \hfill
\subfloat[CelebA, Blended]{\includegraphics[width=8.8cm]{\impath/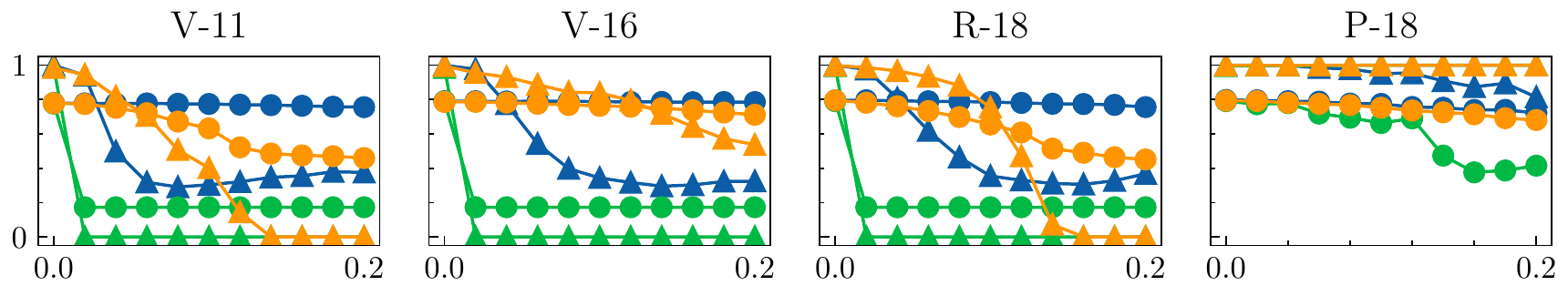}} \\
\subfloat[CelebA, SIG]{\includegraphics[width=8.8cm]{\impath/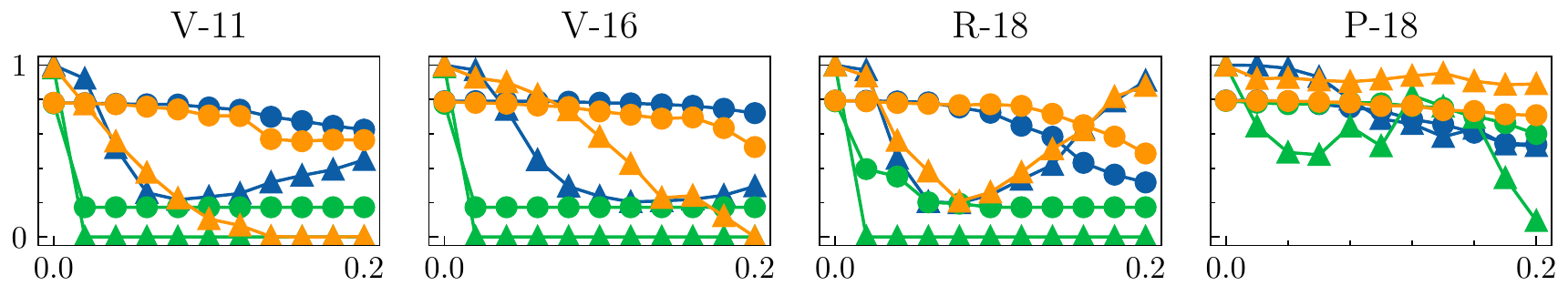}} \hfill 
\subfloat[CelebA, Warped]{\includegraphics[width=8.8cm]{\impath/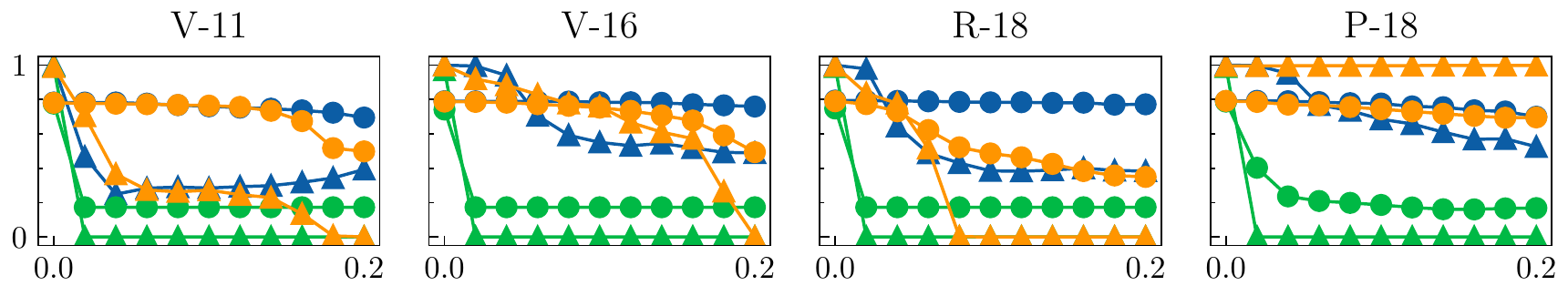}} \\
\subfloat{\includegraphics[height=0.5cm]{\impath/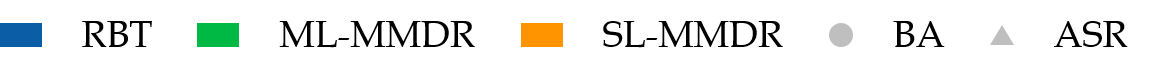}}
\caption{BA and ASR of the infected models when pruning the trigger-related neurons. X-axis: the ratio of neurons pruned. Y-axis: the benign accuracy and the attack success rate.}
\label{fig:np} 
\end{figure*}

\noindent\textbf{Neuron Pruning.} After obtaining the reversed trigger, Wang et al. \cite{wang2019neural} used it to patch the infected model by neuron pruning. To rigorously test the proposed method, we use the real trigger instead of the reversed one for pruning. The results are shown in Fig. \ref{fig:np}.

It can be seen that the curves of BA and ASR are affected by ML-MMDR. For example, the two curves almost overlap on the \{CIFAR-10, V-11, Warped\} models, showing similar downward trends. On the \{CelebA\} models, in most cases, BA and ASR quickly drop to low values simultaneously. The results indicate that the performance of the neuron pruning is reduced on the models trained with ML-MMDR compared to on the models trained with RBT. We think these results are reasonable because neuron pruning essentially assumes that the backdoor behavior is encoded into some neurons, which is similar to the assumption of the distributional differences.

Interestingly, we find that in some cases, ASR declines at the beginning and rises later as the pruning ratio increases, such as on the \{CelebA, R-18, Patched\} model trained with RBT. We believe there are two possible reasons:
\begin{itemize}
\item Neuro pruning preferentially masks the neurons with large activation values on malicious samples, but this does not mean that the model would not classify the input into the target category $t$, especially when the pruning ratio is large. 
\item Backdoor attacks may leave a bias in the infected model during training. This bias would play a dominant role when the model's functionality is greatly broken, causing the model to classify arbitrary inputs into the backdoor target $t$.
\end{itemize}

\subsection{Ablation Study on Different Kernels}
\begin{figure*}[!ht]
\centering
\subfloat[AC]{\includegraphics[width=8.8cm]{\impath/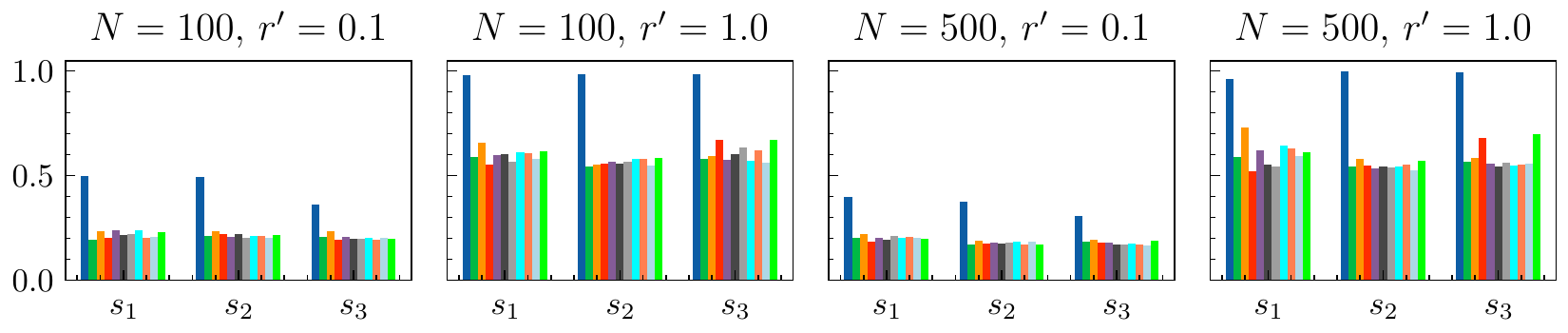}} \hfill
\subfloat[SS]{\includegraphics[width=8.8cm]{\impath/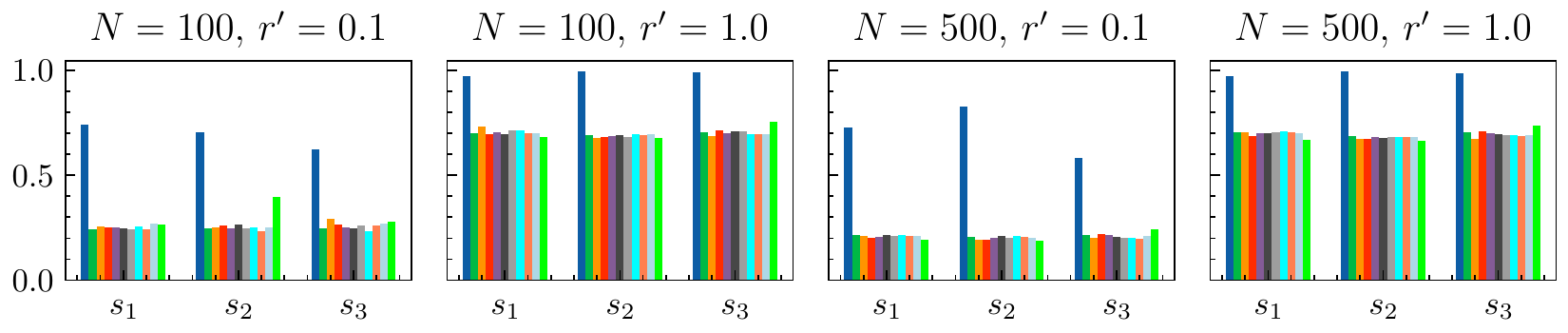}} \\
\subfloat[SR]{\includegraphics[width=8.8cm]{\impath/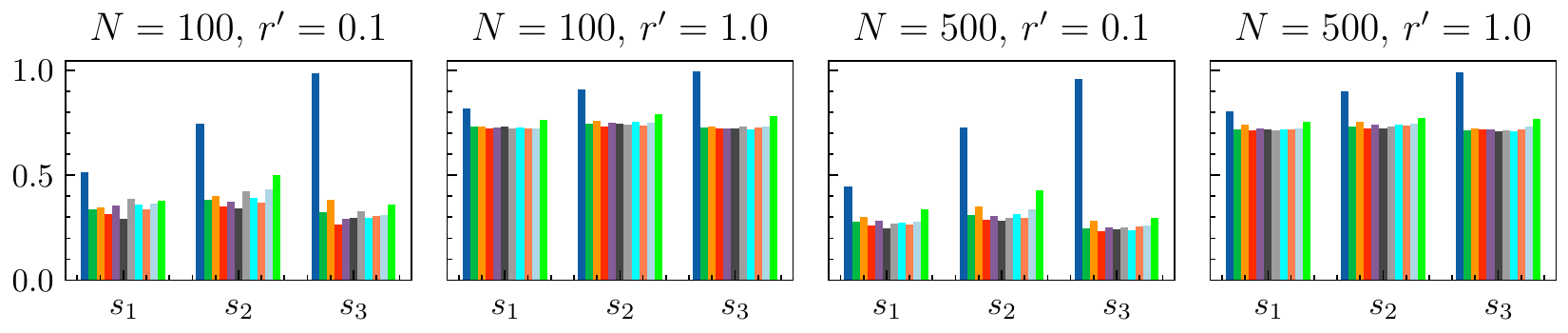}} \\
\subfloat{\includegraphics[height=0.5cm]{\impath/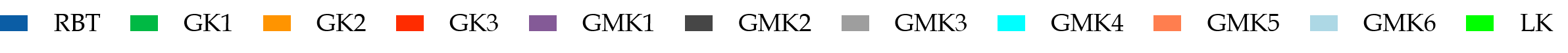}}
\caption{F1 scores of AC, SS, SR on the \{CIFAR-10, P-18, Patched\} models with different kernels.}
\label{fig:kernel_c10_p18_p} 
\end{figure*}

Considering that the choice of a kernel in MMD may have an impact on the training of the backdoored model, we conduct experiments on different kernels, including Gaussian Kernel (GK), Gaussian Mixture Kernel (GMK), and Linear Kernel (LK). The detailed settings are shown in TABLE \ref{tab:kernels}. The results of three difference-based defense methods are shown in Fig. \ref{fig:kernel_c10_p18_p}.

\begin{table}[!ht] 
\centering 
\caption{Settings of different kernels. $\sigma$: the standard deviation.} 
\label{tab:kernels}
\begin{tabular}{l|l} 
\toprule
Kernel & Parameter                                \\ \midrule
GK1    & $\sigma = 1/2$                           \\
GK2    & $\sigma = 1$                             \\
GK3    & $\sigma = 2$                             \\
GMK1   & $\sigma = [1/2, 1, 2]$                   \\
GMK2   & $\sigma = [1/4, 1/2, 1, 2, 4]$           \\
GMK3   & $\sigma = [1/8, 1/4, 1/2, 1, 2, 4, 8]$   \\
GMK4   & $\sigma = [1/3, 1, 3]$                   \\
GMK5   & $\sigma = [1/9, 1/3, 1, 3, 9]$           \\
GMK6   & $\sigma = [1/27, 1/9, 1/3, 1, 3, 9, 27]$ \\
LK     & - \\
\bottomrule
\end{tabular}
\end{table}

Two observations can be seen from the figure. First of all, no matter which kernel is used, ML-MMDR significantly reduces the detection effects. Second, GK or GMK is a better choice than LK. We believe that the reason is that these two types of kernels map the original dimensions to infinite dimensions, which makes the measurement of MMD more accurate.

\section{Discussion} \label{sec:dis}
One of the trends of backdoor attacks is to be more concealed, and this concealment is reflected in two aspects. On the one hand, the attacker hopes that the constructed malicious samples can evade human perception, so some researchers have proposed the invisible attacks \cite{zhong2020backdoor,li2020invisible} and the label-consistent attack \cite{turner2019label}. On the other hand, the attack method needs to be machine imperceptible, meaning that it needs to escape various backdoor defense methods. Some methods have been proposed to achieve this goal, including new trigger forms \cite{nguyen2021wanet} or backdoor training approaches \cite{tan2020bypassing}. Our work also belongs to one of these. However, it is not sufficient to consider machine perception or human perception alone, because the defender may set up multiple defense methods. Constructing an attack that can bypass defense methods with different principles requires further research. 

The proposed method in this paper reduces the distributional differences by adding a constraint to the loss function during the training of a backdoored model. It requires the attacker to have permission to train the model. Another more relaxed setting is that the attacker only provides data, and the user trains the model on the data. Can the distributional differences still be reduced in this case? We argue the answer is positive. However, by combining the optimized triggers \cite{liu2017neural} and MMD, we make a preliminary attempt and find it is not easy. We optimize a trigger on a trained benign model, so that adding the trigger to a bengin sample can not only force the model's output into a specified category, but also minimize the distributional differences of the latent representations. Subsequently, we use this optimized trigger to construct a poisoning training set and train an infected model on these data with RBT. The new model can always optimize parameters that are easier for completing the task, but are less concealed. We plan it in future work.

\section{Related Work} \label{sec:rel_wor}
\subsection{Backdoor Attacks}
Since Gu et al. \cite{gu2017badnets} first explored the backdoor vulnerability in deep learning models, many variants have been proposed. From the method of injecting the hidden threat, backdoor attacks can be roughly divided into two categories, i.e., poisoning-based attacks and non-poisoning-based attacks. Poisoning-based methods  \cite{chen2017targeted,gu2017badnets,liu2017trojaning,turner2019label,li2020invisible,liu2020reflection,nguyen2021wanet} execute the Trojan horse implantation by mixing a small number of malicious samples into the training data and learning an infected model from the mixed dataset. Constructing more concealed and effective malicious samples is the research focus of this type of attack. The primary method \cite{gu2017badnets} uses a static local patch as the trigger. To obtain better attack performance, Liu et al. \cite{liu2017trojaning} built a Trojan method, in which the trigger is optimized rather than pre-defined. Zhong et al. \cite{zhong2020backdoor} argued that the triggers of the previous attacks \cite{gu2017badnets,liu2017trojaning} are all visually visible, which undermines the concealment of the backdoor. Then, they proposed adding an imperceptible perturbation mask to the bengin image to generate its backdoor adversary. However, these attacks ignore that the inconsistency of the instance and its label can increase the risk of disclosure. Turner et al. \cite{turner2019label} first pointed out this problem and proposed the label-consistent backdoor attacks by leveraging adversarial examples and generative models. In addition to the above digital attacks, some studies focus on the physical world. Chen et al. \cite{chen2017targeted} adopted a pair of glasses as the physical trigger to fool a face recognition system. Li et al. \cite{li2020rethinking} suggested that physical transformations should be considered when training a backdoored model.

Non-poisoning-based methods \cite{dumford2018backdooring,kurita2020weight,rakin2020tbt,tang2020embarrassingly,wang2020backdoor} encode the backdoor functionality into deep models by transfer learning or weight perturbations. Dumford and Scheirer \cite{dumford2018backdooring} first explored the possibility of injecting the backdoor without poisoning, where they proposed to modify the model's parameters directly. Unlike \cite{dumford2018backdooring}, Tang et al. \cite{tang2020embarrassingly} proposed a training-free method by inserting a malicious module into the target model instead of perturbing the parameters. 

The method we proposed in this paper can be used to enhance the above attacks to escape difference-based backdoor detection. As shown in \ref{equ:mlmmdr}, ML-MMDR adds a constraint item to the loss function without affecting the regular backdoor training.

\subsection{Backdoor Defenses}
Several defense approaches have been proposed to defend against these attacks, such as backdoor detection \cite{chen2019detecting,ficco2021malware}, trigger synthesis \cite{chen2019deepinspect,wang2019neural}, model reconstruction \cite{liu2017neural,li2021neural}, and model diagnosis \cite{xu2019detecting,kolouri2020universal}. Among them, a vast number of defense methods \cite{chen2019detecting,liu2018fine,tran2018spectral,chou2020sentinet,jin2020unified,soremekun2020exposing,hayase2021spectre} distinguish backdoor samples from the clean ones by exploiting the distributional differences between clean and adversarial representations. Three typical methods have been introduced above, we here add some others. Chou et al. \cite{chou2020sentinet} leveraged the power of visualization tools, such as Grad-CAM \cite{selvaraju2017grad}, to inspect the backdoor behavior of malicious models. Soremekun et al. \cite{soremekun2020exposing} proposed a detection method similar to the activation clustering \cite{chen2019detecting}, which uses t-SNE as the dimensionality reduction and the mean-shift as the clustering algorithm. Hayase et al. \cite{hayase2021spectre} argued that the previous defense \cite{tran2018spectral} works only when the spectral signature is large and proposed a method, SPECTRE, using robust covariance estimation to amplify the signature of malicious data.

However, our work demonstrates that the performance of difference-based detection methods can be greatly reduced when using infected models trained with ML-MMDR. This proves that there is a need to study more powerful defense methods.

\section{Conclusion} \label{sec:con}
Our work studies the distributional differences between the features of clean and adversarial inputs in the infected model, which is the hypothesis of many detection methods. We identify that the distributional differences of multiple levels are all large enough to be used to distinguish inputs. Therefore, we propose a multi-level MMD reduction method and demonstrate that the differences can be considerably reduced without harming the attack intensity. Finally, the experimental results of three difference-based defense methods, i.e., the activation clustering, the spectral signatures, and the subspace reconstruction, indicate that the defense effects decrease drastically as the differences reduce. The proposed method can enhance existing attacks to escape backdoor detection algorithms.

\section*{Acknowledgment}
The work is partially supported by the National Natural Science Foundation of China under grant No.U19B2044 and No.61836011.

\bibliographystyle{IEEEtran}
\bibliography{./references.bib}

\begin{thebibliography}{10}
\providecommand{\url}[1]{#1}
\csname url@samestyle\endcsname
\providecommand{\newblock}{\relax}
\providecommand{\bibinfo}[2]{#2}
\providecommand{\BIBentrySTDinterwordspacing}{\spaceskip=0pt\relax}
\providecommand{\BIBentryALTinterwordstretchfactor}{4}
\providecommand{\BIBentryALTinterwordspacing}{\spaceskip=\fontdimen2\font plus
\BIBentryALTinterwordstretchfactor\fontdimen3\font minus
  \fontdimen4\font\relax}
\providecommand{\BIBforeignlanguage}[2]{{%
\expandafter\ifx\csname l@#1\endcsname\relax
\typeout{** WARNING: IEEEtran.bst: No hyphenation pattern has been}%
\typeout{** loaded for the language `#1'. Using the pattern for}%
\typeout{** the default language instead.}%
\else
\language=\csname l@#1\endcsname
\fi
#2}}
\providecommand{\BIBdecl}{\relax}
\BIBdecl

\bibitem{girshick2015fast}
R.~Girshick, ``Fast r-cnn,'' in \emph{Proceedings of the IEEE international
  conference on computer vision}, 2015, pp. 1440--1448.

\bibitem{long2015fully}
J.~Long, E.~Shelhamer, and T.~Darrell, ``Fully convolutional networks for
  semantic segmentation,'' in \emph{Proceedings of the IEEE conference on
  computer vision and pattern recognition}, 2015, pp. 3431--3440.

\bibitem{redmon2018yolov3}
J.~Redmon and A.~Farhadi, ``Yolov3: An incremental improvement,'' \emph{arXiv
  preprint arXiv:1804.02767}, 2018.

\bibitem{yin2019online}
J.~Yin, P.~Xia, and J.~He, ``Online hard region mining for semantic
  segmentation,'' \emph{Neural Processing Letters}, vol.~50, no.~3, pp.
  2665--2679, 2019.

\bibitem{xia2020boosting}
P.~Xia, J.~He, and J.~Yin, ``Boosting image caption generation with feature
  fusion module,'' \emph{Multimedia Tools and Applications}, vol.~79, no.~33,
  pp. 24\,225--24\,239, 2020.

\bibitem{sutskever2014sequence}
I.~Sutskever, O.~Vinyals, and Q.~V. Le, ``Sequence to sequence learning with
  neural networks,'' \emph{arXiv preprint arXiv:1409.3215}, 2014.

\bibitem{chen2016enhanced}
Q.~Chen, X.~Zhu, Z.~Ling, S.~Wei, H.~Jiang, and D.~Inkpen, ``Enhanced lstm for
  natural language inference,'' \emph{arXiv preprint arXiv:1609.06038}, 2016.

\bibitem{devlin2018bert}
J.~Devlin, M.-W. Chang, K.~Lee, and K.~Toutanova, ``Bert: Pre-training of deep
  bidirectional transformers for language understanding,'' \emph{arXiv preprint
  arXiv:1810.04805}, 2018.

\bibitem{chan2016listen}
W.~Chan, N.~Jaitly, Q.~Le, and O.~Vinyals, ``Listen, attend and spell: A neural
  network for large vocabulary conversational speech recognition,'' in
  \emph{2016 IEEE International Conference on Acoustics, Speech and Signal
  Processing (ICASSP)}.\hskip 1em plus 0.5em minus 0.4em\relax IEEE, 2016, pp.
  4960--4964.

\bibitem{silver2016mastering}
D.~Silver, A.~Huang, C.~J. Maddison, A.~Guez, L.~Sifre, G.~Van Den~Driessche,
  J.~Schrittwieser, I.~Antonoglou, V.~Panneershelvam, M.~Lanctot \emph{et~al.},
  ``Mastering the game of go with deep neural networks and tree search,''
  \emph{nature}, vol. 529, no. 7587, pp. 484--489, 2016.

\bibitem{jumper2021highly}
J.~Jumper, R.~Evans, A.~Pritzel, T.~Green, M.~Figurnov, O.~Ronneberger,
  K.~Tunyasuvunakool, R.~Bates, A.~{\v{Z}}{\'\i}dek, A.~Potapenko
  \emph{et~al.}, ``Highly accurate protein structure prediction with
  alphafold,'' \emph{Nature}, vol. 596, no. 7873, pp. 583--589, 2021.

\bibitem{brown2020language}
T.~Brown, B.~Mann, N.~Ryder, M.~Subbiah, J.~D. Kaplan, P.~Dhariwal,
  A.~Neelakantan, P.~Shyam, G.~Sastry, A.~Askell \emph{et~al.}, ``Language
  models are few-shot learners,'' \emph{Advances in neural information
  processing systems}, vol.~33, pp. 1877--1901, 2020.

\bibitem{chen2019detecting}
B.~Chen, W.~Carvalho, N.~Baracaldo, H.~Ludwig, B.~Edwards, T.~Lee, I.~Molloy,
  and B.~Srivastava, ``Detecting backdoor attacks on deep neural networks by
  activation clustering,'' in \emph{SafeAI@ AAAI}, 2019.

\bibitem{tran2018spectral}
B.~Tran, J.~Li, and A.~Madry, ``Spectral signatures in backdoor attacks,''
  \emph{Advances in neural information processing systems}, vol.~31, 2018.

\bibitem{chen2017targeted}
X.~Chen, C.~Liu, B.~Li, K.~Lu, and D.~Song, ``Targeted backdoor attacks on deep
  learning systems using data poisoning,'' \emph{arXiv preprint
  arXiv:1712.05526}, 2017.

\bibitem{gu2017badnets}
T.~Gu, B.~Dolan-Gavitt, and S.~Garg, ``Badnets: Identifying vulnerabilities in
  the machine learning model supply chain,'' \emph{arXiv preprint
  arXiv:1708.06733}, 2017.

\bibitem{liu2017trojaning}
Y.~Liu, S.~Ma, Y.~Aafer, W.-C. Lee, J.~Zhai, W.~Wang, and X.~Zhang, ``Trojaning
  attack on neural networks,'' 2017.

\bibitem{xue2020one}
M.~Xue, C.~He, J.~Wang, and W.~Liu, ``One-to-n \& n-to-one: Two advanced
  backdoor attacks against deep learning models,'' \emph{IEEE Transactions on
  Dependable and Secure Computing}, 2020.

\bibitem{jin2020unified}
K.~Jin, T.~Zhang, C.~Shen, Y.~Chen, M.~Fan, C.~Lin, and T.~Liu, ``A unified
  framework for analyzing and detecting malicious examples of dnn models,''
  \emph{arXiv preprint arXiv:2006.14871}, 2020.

\bibitem{hayase2021spectre}
J.~Hayase, W.~Kong, R.~Somani, and S.~Oh, ``Spectre: Defending against backdoor
  attacks using robust statistics,'' \emph{arXiv preprint arXiv:2104.11315},
  2021.

\bibitem{tan2020bypassing}
T.~J.~L. Tan and R.~Shokri, ``Bypassing backdoor detection algorithms in deep
  learning,'' in \emph{2020 IEEE European Symposium on Security and Privacy
  (EuroS\&P)}.\hskip 1em plus 0.5em minus 0.4em\relax IEEE, 2020, pp. 175--183.

\bibitem{doan2021backdoor}
K.~Doan, Y.~Lao, and P.~Li, ``Backdoor attack with imperceptible input and
  latent modification,'' \emph{Advances in Neural Information Processing
  Systems}, vol.~34, 2021.

\bibitem{ren2021simtrojan}
Y.~Ren, L.~Li, and J.~Zhou, ``Simtrojan: Stealthy backdoor attack,'' in
  \emph{2021 IEEE International Conference on Image Processing (ICIP)}.\hskip
  1em plus 0.5em minus 0.4em\relax IEEE, 2021, pp. 819--823.

\bibitem{gretton2012kernel}
A.~Gretton, K.~M. Borgwardt, M.~J. Rasch, B.~Sch{\"o}lkopf, and A.~Smola, ``A
  kernel two-sample test,'' \emph{The Journal of Machine Learning Research},
  vol.~13, no.~1, pp. 723--773, 2012.

\bibitem{szekely2013energy}
G.~J. Sz{\'e}kely and M.~L. Rizzo, ``Energy statistics: A class of statistics
  based on distances,'' \emph{Journal of statistical planning and inference},
  vol. 143, no.~8, pp. 1249--1272, 2013.

\bibitem{kolouri2019generalized}
S.~Kolouri, K.~Nadjahi, U.~Simsekli, R.~Badeau, and G.~Rohde, ``Generalized
  sliced wasserstein distances,'' \emph{Advances in Neural Information
  Processing Systems}, vol.~32, 2019.

\bibitem{barni2019new}
M.~Barni, K.~Kallas, and B.~Tondi, ``A new backdoor attack in cnns by training
  set corruption without label poisoning,'' in \emph{2019 IEEE International
  Conference on Image Processing (ICIP)}.\hskip 1em plus 0.5em minus
  0.4em\relax IEEE, 2019, pp. 101--105.

\bibitem{villani2009optimal}
C.~Villani, \emph{Optimal transport: old and new}.\hskip 1em plus 0.5em minus
  0.4em\relax Springer, 2009, vol. 338.

\bibitem{nguyen2021wanet}
A.~Nguyen and A.~Tran, ``Wanet--imperceptible warping-based backdoor attack,''
  \emph{arXiv preprint arXiv:2102.10369}, 2021.

\bibitem{liu2015deep}
Z.~Liu, P.~Luo, X.~Wang, and X.~Tang, ``Deep learning face attributes in the
  wild,'' in \emph{Proceedings of the IEEE international conference on computer
  vision}, 2015, pp. 3730--3738.

\bibitem{javaheripi2020cleann}
M.~Javaheripi, M.~Samragh, G.~Fields, T.~Javidi, and F.~Koushanfar, ``Cleann:
  Accelerated trojan shield for embedded neural networks,'' in \emph{2020
  IEEE/ACM International Conference On Computer Aided Design (ICCAD)}.\hskip
  1em plus 0.5em minus 0.4em\relax IEEE, 2020, pp. 1--9.

\bibitem{wang2019neural}
B.~Wang, Y.~Yao, S.~Shan, H.~Li, B.~Viswanath, H.~Zheng, and B.~Y. Zhao,
  ``Neural cleanse: Identifying and mitigating backdoor attacks in neural
  networks,'' in \emph{2019 IEEE Symposium on Security and Privacy (SP)}.\hskip
  1em plus 0.5em minus 0.4em\relax IEEE, 2019, pp. 707--723.

\bibitem{krizhevsky2009learning}
A.~Krizhevsky, G.~Hinton \emph{et~al.}, ``Learning multiple layers of features
  from tiny images,'' 2009.

\bibitem{salem2020dynamic}
A.~Salem, R.~Wen, M.~Backes, S.~Ma, and Y.~Zhang, ``Dynamic backdoor attacks
  against machine learning models,'' \emph{arXiv preprint arXiv:2003.03675},
  2020.

\bibitem{simonyan2014very}
K.~Simonyan and A.~Zisserman, ``Very deep convolutional networks for
  large-scale image recognition,'' \emph{arXiv preprint arXiv:1409.1556}, 2014.

\bibitem{he2016deep}
K.~He, X.~Zhang, S.~Ren, and J.~Sun, ``Deep residual learning for image
  recognition,'' in \emph{Proceedings of the IEEE conference on computer vision
  and pattern recognition}, 2016, pp. 770--778.

\bibitem{he2016identity}
------, ``Identity mappings in deep residual networks,'' in \emph{European
  conference on computer vision}.\hskip 1em plus 0.5em minus 0.4em\relax
  Springer, 2016, pp. 630--645.

\bibitem{paszke2017automatic}
A.~Paszke, S.~Gross, S.~Chintala, G.~Chanan, E.~Yang, Z.~DeVito, Z.~Lin,
  A.~Desmaison, L.~Antiga, and A.~Lerer, ``Automatic differentiation in
  pytorch,'' 2017.

\bibitem{zhong2020backdoor}
H.~Zhong, C.~Liao, A.~C. Squicciarini, S.~Zhu, and D.~Miller, ``Backdoor
  embedding in convolutional neural network models via invisible
  perturbation,'' in \emph{Proceedings of the Tenth ACM Conference on Data and
  Application Security and Privacy}, 2020, pp. 97--108.

\bibitem{li2020invisible}
S.~Li, M.~Xue, B.~Z.~H. Zhao, H.~Zhu, and X.~Zhang, ``Invisible backdoor
  attacks on deep neural networks via steganography and regularization,''
  \emph{IEEE Transactions on Dependable and Secure Computing}, vol.~18, no.~5,
  pp. 2088--2105, 2020.

\bibitem{turner2019label}
A.~Turner, D.~Tsipras, and A.~Madry, ``Label-consistent backdoor attacks,''
  \emph{arXiv preprint arXiv:1912.02771}, 2019.

\bibitem{liu2017neural}
Y.~Liu, Y.~Xie, and A.~Srivastava, ``Neural trojans,'' in \emph{2017 IEEE
  International Conference on Computer Design (ICCD)}.\hskip 1em plus 0.5em
  minus 0.4em\relax IEEE, 2017, pp. 45--48.

\bibitem{liu2020reflection}
Y.~Liu, X.~Ma, J.~Bailey, and F.~Lu, ``Reflection backdoor: A natural backdoor
  attack on deep neural networks,'' in \emph{European Conference on Computer
  Vision}.\hskip 1em plus 0.5em minus 0.4em\relax Springer, 2020, pp. 182--199.

\bibitem{li2020rethinking}
Y.~Li, T.~Zhai, B.~Wu, Y.~Jiang, Z.~Li, and S.~Xia, ``Rethinking the trigger of
  backdoor attack,'' \emph{arXiv preprint arXiv:2004.04692}, 2020.

\bibitem{dumford2018backdooring}
J.~Dumford and W.~Scheirer, ``Backdooring convolutional neural networks via
  targeted weight perturbations,'' in \emph{2020 IEEE International Joint
  Conference on Biometrics (IJCB)}.\hskip 1em plus 0.5em minus 0.4em\relax
  IEEE, 2018, pp. 1--9.

\bibitem{kurita2020weight}
K.~Kurita, P.~Michel, and G.~Neubig, ``Weight poisoning attacks on pre-trained
  models,'' \emph{arXiv preprint arXiv:2004.06660}, 2020.

\bibitem{rakin2020tbt}
A.~S. Rakin, Z.~He, and D.~Fan, ``Tbt: Targeted neural network attack with bit
  trojan,'' in \emph{Proceedings of the IEEE/CVF Conference on Computer Vision
  and Pattern Recognition}, 2020, pp. 13\,198--13\,207.

\bibitem{tang2020embarrassingly}
R.~Tang, M.~Du, N.~Liu, F.~Yang, and X.~Hu, ``An embarrassingly simple approach
  for trojan attack in deep neural networks,'' in \emph{Proceedings of the 26th
  ACM SIGKDD International Conference on Knowledge Discovery \& Data Mining},
  2020, pp. 218--228.

\bibitem{wang2020backdoor}
S.~Wang, S.~Nepal, C.~Rudolph, M.~Grobler, S.~Chen, and T.~Chen, ``Backdoor
  attacks against transfer learning with pre-trained deep learning models,''
  \emph{IEEE Transactions on Services Computing}, 2020.

\bibitem{ficco2021malware}
M.~Ficco, ``Malware analysis by combining multiple detectors and observation
  windows,'' \emph{IEEE Transactions on Computers}, 2021.

\bibitem{chen2019deepinspect}
H.~Chen, C.~Fu, J.~Zhao, and F.~Koushanfar, ``Deepinspect: A black-box trojan
  detection and mitigation framework for deep neural networks.'' in
  \emph{IJCAI}, 2019, pp. 4658--4664.

\bibitem{li2021neural}
Y.~Li, X.~Lyu, N.~Koren, L.~Lyu, B.~Li, and X.~Ma, ``Neural attention
  distillation: Erasing backdoor triggers from deep neural networks,''
  \emph{arXiv preprint arXiv:2101.05930}, 2021.

\bibitem{xu2019detecting}
X.~Xu, Q.~Wang, H.~Li, N.~Borisov, C.~A. Gunter, and B.~Li, ``Detecting ai
  trojans using meta neural analysis,'' \emph{arXiv preprint arXiv:1910.03137},
  2019.

\bibitem{kolouri2020universal}
S.~Kolouri, A.~Saha, H.~Pirsiavash, and H.~Hoffmann, ``Universal litmus
  patterns: Revealing backdoor attacks in cnns,'' in \emph{Proceedings of the
  IEEE/CVF Conference on Computer Vision and Pattern Recognition}, 2020, pp.
  301--310.

\bibitem{liu2018fine}
K.~Liu, B.~Dolan-Gavitt, and S.~Garg, ``Fine-pruning: Defending against
  backdooring attacks on deep neural networks,'' in \emph{International
  Symposium on Research in Attacks, Intrusions, and Defenses}.\hskip 1em plus
  0.5em minus 0.4em\relax Springer, 2018, pp. 273--294.

\bibitem{chou2020sentinet}
E.~Chou, F.~Tram{\`e}r, and G.~Pellegrino, ``Sentinet: Detecting localized
  universal attacks against deep learning systems,'' in \emph{2020 IEEE
  Security and Privacy Workshops (SPW)}.\hskip 1em plus 0.5em minus 0.4em\relax
  IEEE, 2020, pp. 48--54.

\bibitem{soremekun2020exposing}
E.~Soremekun, S.~Udeshi, S.~Chattopadhyay, and A.~Zeller, ``Exposing backdoors
  in robust machine learning models,'' \emph{arXiv preprint arXiv:2003.00865},
  2020.

\bibitem{selvaraju2017grad}
R.~R. Selvaraju, M.~Cogswell, A.~Das, R.~Vedantam, D.~Parikh, and D.~Batra,
  ``Grad-cam: Visual explanations from deep networks via gradient-based
  localization,'' in \emph{Proceedings of the IEEE international conference on
  computer vision}, 2017, pp. 618--626.

\end{thebibliography}

\clearpage
\setcounter{table}{0}
\setcounter{figure}{0}
\appendices
\section{DNN Architectures} \label{app:dnn_arc}
The details of the DNN architectures and the selected layers are shown in TABLE \ref{tab:dnn_arc}. For V-11 and V-16, we always choose the ReLU layers to extract the latent representations. For R-18 or P-18, we use the RBB or PBB blocks to extract the latent representations.

\begin{table}[!h] 
\centering 
\caption{DNN architectures used in this paper. Conv: a convolutional layer. BN: a batch normalization layer. ReLU: a rectified linear activation layer. MP: a max-pooling layer. GAP: a global average pooling layer. FC: a fully connected layer. RBB: a basic block for ResNet. PBB: a basic block for PreActResNet. The \hl{underline} indicates that the outputs of that layer or block are used as the latent representations of inputs.}
\label{tab:dnn_arc}
\begin{tabular}{c|c|c|c|c} 
\toprule
Layer & V-11      & V-16      & R-18                      & P-18                      \\ \midrule
1     & Conv      & Conv      & Conv                      & Conv                      \\ 
2     & BN        & BN        & BN                        & BN                        \\
3     & ReLU      & ReLU      & ReLU                      & ReLU                      \\ \cline{4-5}
4     & MP        & Conv      & \multirow{6}{*}{RBB}      & \multirow{6}{*}{PBB}      \\
5     & Conv      & BN        &                           &                           \\
6     & BN        & ReLU      &                           &                           \\
7     & ReLU      & MP        &                           &                           \\
8     & MP        & Conv      &                           &                           \\
9     & Conv      & BN        &                           &                           \\ \cline{4-5}
10    & BN        & ReLU      & \multirow{6}{*}{RBB}      & \multirow{6}{*}{PBB}      \\
11    & ReLU      & Conv      &                           &                           \\
12    & Conv      & BN        &                           &                           \\ 
13    & BN        & ReLU      &                           &                           \\
14    & \hl{ReLU} & MP        &                           &                           \\
15    & MP        & Conv      &                           &                           \\ \cline{4-5}
16    & Conv      & BN        & \multirow{6}{*}{RBB}      & \multirow{6}{*}{PBB}      \\
17    & BN        & ReLU      &                           &                           \\
18    & ReLU      & Conv      &                           &                           \\
19    & Conv      & BN        &                           &                           \\
20    & BN        & ReLU      &                           &                           \\
21    & \hl{ReLU} & Conv      &                           &                           \\ \cline{4-5}
22    & MP        & BN        & \multirow{6}{*}{\hl{RBB}} & \multirow{6}{*}{\hl{PBB}} \\ 
23    & Conv      & \hl{ReLU} &                           &                           \\
24    & BN        & MP        &                           &                           \\
25    & ReLU      & Conv      &                           &                           \\
26    & Conv      & BN        &                           &                           \\
27    & BN        & ReLU      &                           &                           \\ \cline{4-5}
28    & \hl{ReLU} & Conv      & \multirow{6}{*}{RBB}      & \multirow{6}{*}{PBB}      \\
29    & GAP       & BN        &                           &                           \\
30    & FC        & ReLU      &                           &                           \\
31    &           & Conv      &                           &                           \\
32    &           & BN        &                           &                           \\
33    &           & \hl{ReLU} &                           &                           \\ \cline{4-5}
34    &           & MP        & \multirow{6}{*}{\hl{RBB}} & \multirow{6}{*}{\hl{PBB}} \\ 
35    &           & Conv      &                           &                           \\
36    &           & BN        &                           &                           \\
37    &           & ReLU      &                           &                           \\
38    &           & Conv      &                           &                           \\
39    &           & BN        &                           &                           \\ \cline{4-5}
40    &           & ReLU      & \multirow{6}{*}{RBB}      & \multirow{6}{*}{PBB}      \\
41    &           & Conv      &                           &                           \\
42    &           & BN        &                           &                           \\
43    &           & \hl{ReLU} &                           &                           \\
44    &           & GAP       &                           &                           \\
45    &           & FC        &                           &                           \\ \cline{4-5}
46    &           &           & \multirow{6}{*}{\hl{RBB}} & \multirow{6}{*}{\hl{PBB}} \\ 
47    &           &           &                           &                           \\
48    &           &           &                           &                           \\
49    &           &           &                           &                           \\
50    &           &           &                           &                           \\
51    &           &           &                           &                           \\ \cline{4-5}
52    &           &           & GAP                       & GAP                       \\
53    &           &           & FC                        & FC                        \\
\bottomrule
\end{tabular}
\label{tab:structure}
\end{table}

\section{Quantification of the Distributional Differences} \label{app:qua_dif}
MMD, ED, and SWD are used to quantify the distributional differences between the latent representations of benign and malicious samples at the three levels. The specific distance values on CIFAR-10 and CelebA are shown in TABLE \ref{tab:dis_c10} and TABLE \ref{tab:dis_ce}, respectively.

\begin{table*}[!ht] 
\centering 
\caption{MMD, ED, and SWD values on CIFAR-10 between the latent representations of benign and malicious samples at the three levels. All values are averaged over sets of 500 inputs sampled randomly from the particular data.}
\label{tab:dis_c10}
\begin{tabular}{c|c||P{\colwid}|P{\colwid}|P{\colwid}||P{\colwid}|P{\colwid}|P{\colwid}||P{\colwid}|P{\colwid}|P{\colwid}} 
\toprule
\multirow{2}{*}{Model} & \multirow{2}{*}{Method} & \multicolumn{3}{c||}{$\mmd$} & \multicolumn{3}{c||}{$\ed$} & \multicolumn{3}{c}{$\swd$} \\ 
                      &             & $s_1$ & $s_2$ & $s_3$ & $s_1$ & $s_2$ & $s_3$ & $s_1$ & $s_2$ & $s_3$ \\ \midrule \midrule
\multirow{3}{*}{V-11} & Patched     & 0.41 & 0.84 & 2.88 & 778.82 & 973.58 & 606.35 & 167.14 & 271.81 & 130.11 \\
                      & Intra-class & 0.02 & 0.02 & 0.01 & 11.17 & 6.45 & 1.00 & 21.19 & 7.86 & 3.13 \\
                      & Inter-class & 0.18 & 0.44 & 2.47 & 280.59 & 460.08 & 622.32 & 180.37 & 208.98 & 32.67 \\ \midrule
\multirow{3}{*}{V-16} & Patched     & 0.59 & 1.86 & 3.27 & 1286.90 & 2627.58 & 384.35 & 318.80 & 679.74 & 111.44 \\
                      & Intra-class & 0.01 & 0.01 & 0.01 & 10.38 & 3.44 & 0.28 & 11.32 & 10.84 & 3.39 \\
                      & Inter-class & 0.27 & 0.91 & 3.95 & 521.48 & 1017.54 & 861.37 & 361.89 & 375.39 & 47.60 \\ \midrule
\multirow{3}{*}{R-18} & Patched     & 0.29 & 0.58 & 1.85 & 4898.08 & 4594.22 & 7626.06 & 467.06 & 475.38 & 87.39 \\
                      & Intra-class & 0.02 & 0.01 & 0.01 & 110.74 & 34.85 & 19.56 & 41.87 & 15.68 & 26.11 \\
                      & Inter-class & 0.16 & 0.25 & 1.06 & 2407.12 & 1726.27 & 3794.22 & 480.62 & 312.22 & 109.04 \\ \midrule
\multirow{3}{*}{R-18} & Patched     & 0.27 & 0.38 & 1.11 & 3541.53 & 1819.97 & 3507.75 & 136.28 & 36.06 & 260.36 \\
                      & Intra-class & 0.02 & 0.01 & 0.01 & 92.33 & 25.61 & 8.83 & 77.79 & 24.00 & 8.85 \\
                      & Inter-class & 0.16 & 0.29 & 2.06 & 1794.53 & 1292.70 & 7961.95 & 472.66 & 154.34 & 274.56 \\ \midrule \midrule
\multirow{3}{*}{V-11} & Blended     & 0.26 & 0.90 & 3.02 & 445.24 & 994.06 & 573.97 & 142.29 & 228.55 & 129.32 \\
                      & Intra-class & 0.01 & 0.02 & 0.01 & 10.67 & 5.26 & 0.91 & 17.43 & 6.90 & 4.43 \\
                      & Inter-class & 0.19 & 0.43 & 2.46 & 322.78 & 429.84 & 631.29 & 191.69 & 239.53 & 29.38 \\ \midrule
\multirow{3}{*}{V-16} & Blended     & 0.41 & 1.64 & 3.86 & 798.74 & 1958.26 & 445.55 & 194.21 & 431.69 & 141.87 \\
                      & Intra-class & 0.01 & 0.02 & 0.01 & 10.80 & 5.30 & 1.15 & 12.86 & 19.20 & 3.28 \\
                      & Inter-class & 0.24 & 1.01 & 4.19 & 453.56 & 1156.00 & 887.87 & 285.58 & 473.22 & 65.75 \\ \midrule
\multirow{3}{*}{R-18} & Blended     & 0.28 & 0.31 & 1.09 & 4251.33 & 2112.81 & 3429.00 & 191.57 & 138.60 & 50.17 \\
                      & Intra-class & 0.01 & 0.01 & 0.01 & 90.34 & 44.57 & 13.39 & 66.76 & 15.31 & 33.18 \\
                      & Inter-class & 0.19 & 0.26 & 1.12 & 2835.14 & 1802.27 & 4079.09 & 1135.98 & 356.98 & 109.47 \\ \midrule
\multirow{3}{*}{P-18} & Blended     & 0.28 & 0.39 & 2.54 & 3402.49 & 1823.00 & 12990.96 & 708.26 & 106.44 & 272.66 \\
                      & Intra-class & 0.01 & 0.01 & 0.01 & 79.71 & 27.83 & 14.88 & 34.93 & 21.38 & 7.07 \\
                      & Inter-class & 0.19 & 0.32 & 2.41 & 2214.61 & 1548.64 & 10264.65 & 309.86 & 141.03 & 34.96 \\ \midrule \midrule
\multirow{3}{*}{V-11} & SIG         & 0.29 & 0.92 & 3.33 & 503.06 & 1075.74 & 719.73 & 172.31 & 356.56 & 112.83 \\
                      & Intra-class & 0.02 & 0.02 & 0.01 & 11.21 & 6.48 & 0.98 & 13.37 & 9.17 & 3.37 \\
                      & Inter-class & 0.22 & 0.43 & 2.55 & 352.16 & 436.90 & 657.50 & 202.55 & 217.84 & 22.85 \\ \midrule
\multirow{3}{*}{V-16} & SIG         & 0.56 & 1.87 & 3.49 & 1219.64 & 2656.29 & 308.19 & 483.49 & 769.85 & 116.86 \\
                      & Intra-class & 0.01 & 0.01 & 0.01 & 10.99 & 4.38 & 0.56 & 12.49 & 7.35 & 4.18 \\
                      & Inter-class & 0.24 & 1.00 & 4.21 & 460.41 & 1121.27 & 867.11 & 310.03 & 394.09 & 38.28 \\ \midrule
\multirow{3}{*}{R-18} & SIG         & 0.37 & 0.45 & 1.52 & 5910.28 & 3190.55 & 4969.67 & 774.35 & 147.18 & 132.84 \\
                      & Intra-class & 0.01 & 0.01 & 0.02 & 94.78 & 38.04 & 20.07 & 113.43 & 24.77 & 39.25 \\
                      & Inter-class & 0.19 & 0.27 & 1.20 & 2777.41 & 1872.69 & 4622.49 & 1185.90 & 229.39 & 44.91 \\ \midrule
\multirow{3}{*}{P-18} & SIG         & 0.33 & 0.50 & 2.22 & 4421.40 & 2556.99 & 10010.73 & 281.72 & 277.23 & 75.30 \\
                      & Intra-class & 0.01 & 0.01 & 0.01 & 71.52 & 26.03 & 7.74 & 44.37 & 13.68 & 6.89 \\
                      & Inter-class & 0.17 & 0.31 & 2.60 & 2017.53 & 1549.07 & 12497.88 & 48.41 & 205.11 & 374.96 \\ \midrule \midrule
\multirow{3}{*}{V-11} & Warped      & 0.77 & 1.49 & 3.65 & 1659.84 & 2215.77 & 720.80 & 1160.18 & 1037.38 & 152.64 \\
                      & Intra-class & 0.01 & 0.01 & 0.01 & 9.80 & 3.95 & 1.07 & 17.51 & 5.25 & 2.68 \\
                      & Inter-class & 0.22 & 0.43 & 2.54 & 341.63 & 380.34 & 664.04 & 210.73 & 203.33 & 33.79 \\ \midrule
\multirow{3}{*}{V-16} & Warped      & 1.21 & 1.72 & 3.98 & 3114.98 & 2334.75 & 439.79 & 2312.96 & 927.90 & 141.02 \\
                      & Intra-class & 0.01 & 0.02 & 0.01 & 9.19 & 4.45 & 0.78 & 9.65 & 12.80 & 1.79 \\
                      & Inter-class & 0.27 & 0.93 & 3.96 & 443.33 & 979.75 & 799.54 & 340.94 & 368.47 & 56.23 \\ \midrule
\multirow{3}{*}{R-18} & Warped      & 0.46 & 0.59 & 1.03 & 7807.47 & 4156.52 & 2668.61 & 503.87 & 125.95 & 609.97 \\
                      & Intra-class & 0.01 & 0.02 & 0.02 & 79.31 & 43.34 & 21.76 & 50.00 & 27.71 & 23.54 \\
                      & Inter-class & 0.17 & 0.28 & 1.21 & 2535.64 & 1908.70 & 4732.28 & 703.90 & 435.46 & 60.81 \\ \midrule
\multirow{3}{*}{P-18} & Warped     & 0.42 & 0.55 & 1.93 & 5818.77 & 2655.92 & 6626.69 & 230.71 & 319.87 & 52.46 \\
                      & Intra-class & 0.01 & 0.02 & 0.01 & 72.00 & 34.00 & 15.78 & 50.25 & 32.63 & 9.80 \\
                      & Inter-class & 0.20 & 0.31 & 2.42 & 2346.60 & 1477.92 & 10500.92 & 155.89 & 88.97 & 292.53 \\ \midrule
\bottomrule	                 
\end{tabular}
\end{table*}

\begin{table*}[!ht] 
\centering 
\caption{MMD, ED, and SWD values on CelebA between the latent representations of benign and malicious samples at the three levels. All values are averaged over sets of 500 inputs sampled randomly from the particular data.}
\label{tab:dis_ce}
\begin{tabular}{c|c||P{\colwid}|P{\colwid}|P{\colwid}||P{\colwid}|P{\colwid}|P{\colwid}||P{\colwid}|P{\colwid}|P{\colwid}} 
\toprule
\multirow{2}{*}{Model} & \multirow{2}{*}{Method} & \multicolumn{3}{c||}{$\mmd$} & \multicolumn{3}{c||}{$\ed$} & \multicolumn{3}{c}{$\swd$} \\ 
                      &             & $s_1$ & $s_2$ & $s_3$ & $s_1$ & $s_2$ & $s_3$ & $s_1$ & $s_2$ & $s_3$ \\ \midrule \midrule
\multirow{3}{*}{V-11} & Patched     & 0.58 & 2.25 & 3.04 & 932.95 & 2066.36 & 888.92 & 169.11 & 293.66 & 205.00 \\
                      & Intra-class & 0.02 & 0.02 & 0.02 & 9.64 & 2.59 & 1.18 & 15.27 & 12.72 & 8.68 \\
                      & Inter-class & 0.09 & 0.54 & 1.47 & 117.69 & 275.41 & 190.44 & 8.92 & 52.58 & 22.93 \\ \midrule
\multirow{3}{*}{V-16} & Patched     & 1.32 & 3.54 & 3.31 & 2540.82 & 3320.80 & 806.02 & 450.21 & 460.55 & 207.84 \\
                      & Intra-class & 0.02 & 0.02 & 0.03 & 13.26 & 3.57 & 1.40 & 9.77 & 17.00 & 12.25 \\
                      & Inter-class & 0.24 & 1.33 & 1.63 & 314.71 & 516.73 & 226.58 & 44.44 & 112.95 & 26.27 \\ \midrule
\multirow{3}{*}{R-18} & Patched     & 0.27 & 1.26 & 3.61 & 3117.11 & 7926.51 & 19089.21 & 371.52 & 809.13 & 3412.75 \\
                      & Intra-class & 0.02 & 0.02 & 0.02 & 83.23 & 31.93 & 8.35 & 20.14 & 26.54 & 34.68 \\
                      & Inter-class & 0.08 & 0.40 & 1.18 & 794.15 & 1783.98 & 1881.09 & 29.32 & 141.38 & 43.78 \\ \midrule
\multirow{3}{*}{P-18} & Patched     & 0.29 & 1.49 & 2.98 & 1395.52 & 3931.52 & 10782.47 & 196.65 & 335.78 & 121.51 \\
                      & Intra-class & 0.02 & 0.02 & 0.01 & 31.60 & 13.52 & 3.61 & 23.65 & 21.54 & 9.48 \\
                      & Inter-class & 0.08 & 0.43 & 1.35 & 306.56 & 732.31 & 1853.09 & 84.68 & 87.12 & 22.68 \\ \midrule \midrule
\multirow{3}{*}{V-11} & Blended     & 0.25 & 1.81 & 3.59 & 352.43 & 1332.43 & 1123.55 & 131.30 & 150.84 & 247.53 \\
                      & Intra-class & 0.02 & 0.02 & 0.02 & 15.71 & 2.47 & 0.93 & 19.18 & 7.81 & 8.74 \\
                      & Inter-class & 0.09 & 0.61 & 1.32 & 122.77 & 320.88 & 173.47 & 9.76 & 46.22 & 15.29 \\ \midrule
\multirow{3}{*}{V-16} & Blended     & 0.80 & 3.49 & 3.96 & 1279.55 & 2406.84 & 912.84 & 255.41 & 315.77 & 203.79 \\
                      & Intra-class & 0.02 & 0.01 & 0.01 & 10.05 & 1.33 & 0.28 & 13.99 & 8.63 & 5.08 \\
                      & Inter-class & 0.24 & 1.16 & 1.55 & 323.90 & 397.31 & 212.21 & 25.64 & 117.19 & 16.69 \\ \midrule
\multirow{3}{*}{R-18} & Blended     & 0.20 & 0.74 & 3.91 & 2128.99 & 4083.16 & 21899.05 & 126.18 & 463.06 & 4165.82 \\
                      & Intra-class & 0.02 & 0.02 & 0.02 & 104.46 & 28.57 & 9.34 & 67.40 & 28.77 & 27.17 \\
                      & Inter-class & 0.09 & 0.35 & 1.30 & 888.29 & 1635.75 & 2082.18 & 61.34 & 118.59 & 29.75 \\ \midrule
\multirow{3}{*}{P-18} & Blended     & 0.22 & 1.04 & 4.65 & 971.04 & 2226.99 & 27127.98 & 93.00 & 265.71 & 410.14 \\
                      & Intra-class & 0.02 & 0.02 & 0.01 & 35.05 & 9.81 & 1.63 & 33.99 & 8.50 & 5.28 \\
                      & Inter-class & 0.07 & 0.40 & 1.47 & 297.17 & 684.72 & 1883.19 & 31.88 & 64.16 & 52.16 \\ \midrule \midrule
\multirow{3}{*}{V-11} & SIG         & 0.28 & 2.13 & 3.74 & 420.24 & 1876.40 & 1398.77 & 81.00 & 336.93 & 305.82 \\
                      & Intra-class & 0.02 & 0.02 & 0.01 & 9.69 & 2.15 & 0.16 & 9.24 & 8.82 & 3.33 \\
                      & Inter-class & 0.09 & 0.58 & 1.37 & 116.33 & 297.75 & 182.32 & 9.71 & 40.21 & 16.81 \\ \midrule
\multirow{3}{*}{V-16} & SIG         & 1.14 & 4.02 & 3.32 & 1986.21 & 3170.06 & 625.41 & 493.58 & 468.55 & 149.46 \\
                      & Intra-class & 0.02 & 0.02 & 0.02 & 9.30 & 2.33 & 0.88 & 10.58 & 11.34 & 7.65 \\
                      & Inter-class & 0.25 & 1.47 & 1.60 & 314.24 & 464.11 & 230.23 & 27.35 & 121.42 & 23.92 \\ \midrule
\multirow{3}{*}{R-18} & SIG         & 0.27 & 1.15 & 4.30 & 2942.51 & 7034.58 & 31957.46 & 320.26 & 766.79 & 5457.98 \\
                      & Intra-class & 0.02 & 0.02 & 0.01 & 89.73 & 39.39 & 4.98 & 35.29 & 31.60 & 33.97 \\
                      & Inter-class & 0.08 & 0.38 & 1.19 & 782.21 & 1712.78 & 1790.68 & 48.84 & 200.99 & 69.90 \\ \midrule
\multirow{3}{*}{P-18} & SIG         & 0.27 & 1.33 & 4.33 & 1380.08 & 3530.54 & 26558.91 & 243.13 & 480.76 & 485.03 \\
                      & Intra-class & 0.02 & 0.02 & 0.03 & 34.52 & 15.87 & 14.26 & 48.92 & 10.75 & 5.45 \\
                      & Inter-class & 0.07 & 0.43 & 1.46 & 302.50 & 801.63 & 1681.01 & 68.84 & 37.40 & 108.35 \\ \midrule \midrule
\multirow{3}{*}{V-11} & Warped      & 0.82 & 2.72 & 4.10 & 1383.38 & 2240.92 & 1138.15 & 630.73 & 357.68 & 241.28 \\
                      & Intra-class & 0.02 & 0.02 & 0.01 & 9.58 & 2.80 & 0.31 & 17.76 & 11.73 & 4.04 \\
                      & Inter-class & 0.09 & 0.58 & 1.30 & 120.40 & 297.94 & 168.84 & 11.43 & 30.95 & 6.32 \\ \midrule
\multirow{3}{*}{V-16} & Warped      & 1.93 & 4.29 & 4.13 & 3259.40 & 2034.82 & 555.86 & 743.10 & 117.84 & 114.63 \\
                      & Intra-class & 0.02 & 0.02 & 0.01 & 8.17 & 1.50 & 0.39 & 8.53 & 8.18 & 5.87 \\
                      & Inter-class & 0.23 & 1.19 & 1.68 & 297.12 & 445.75 & 255.00 & 18.21 & 54.99 & 8.92 \\ \midrule
\multirow{3}{*}{R-18} & Warped      & 0.41 & 1.64 & 4.75 & 4863.88 & 10773.26 & 27248.32 & 415.60 & 1075.88 & 4853.27 \\
                      & Intra-class & 0.02 & 0.02 & 0.02 & 89.28 & 30.22 & 12.61 & 38.00 & 23.76 & 67.28 \\
                      & Inter-class & 0.09 & 0.36 & 1.19 & 845.40 & 1642.50 & 1870.43 & 31.34 & 144.72 & 69.24 \\ \midrule
\multirow{3}{*}{P-18} & Warped      & 0.51 & 2.68 & 4.82 & 2687.21 & 9386.98 & 20266.48 & 259.45 & 545.32 & 225.31 \\
                      & Intra-class & 0.02 & 0.02 & 0.02 & 37.62 & 10.47 & 4.46 & 44.28 & 8.19 & 9.45 \\
                      & Inter-class & 0.08 & 0.44 & 1.57 & 313.12 & 760.52 & 1873.21 & 48.90 & 58.20 & 71.26 \\ \midrule
\bottomrule	                 
\end{tabular}
\end{table*}

\section{Results of ML-MMDR} \label{app:res_mlmmdr}
\begin{figure*}[!ht]
\centering
\subfloat[CIFAR-10, Patched]{\includegraphics[width=8.8cm]{\impath/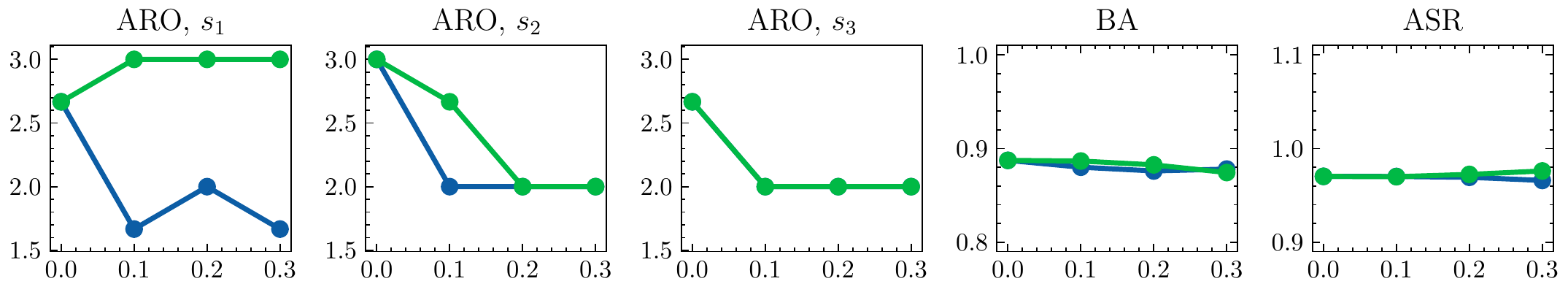}} \hfill
\subfloat[CIFAR-10, Blended]{\includegraphics[width=8.8cm]{\impath/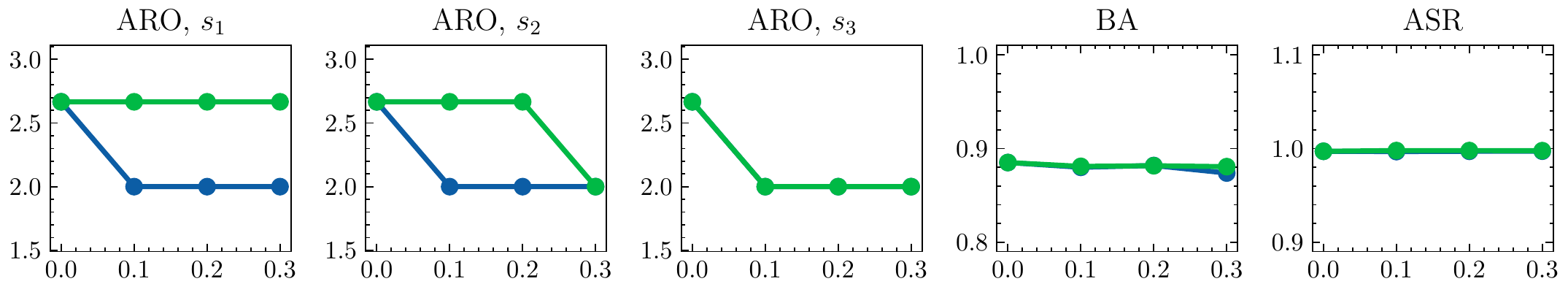}} \\
\subfloat[CIFAR-10, SIG]{\includegraphics[width=8.8cm]{\impath/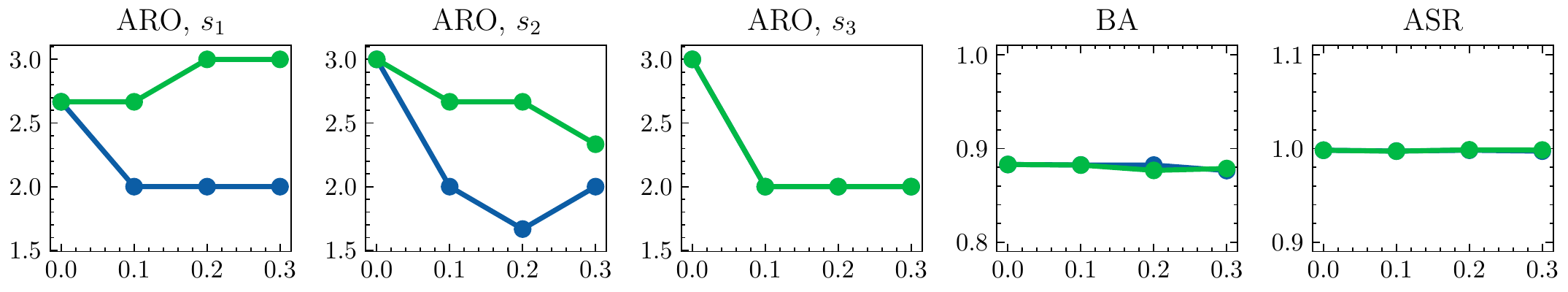}} \hfill 
\subfloat[CIFAR-10, Warped]{\includegraphics[width=8.8cm]{\impath/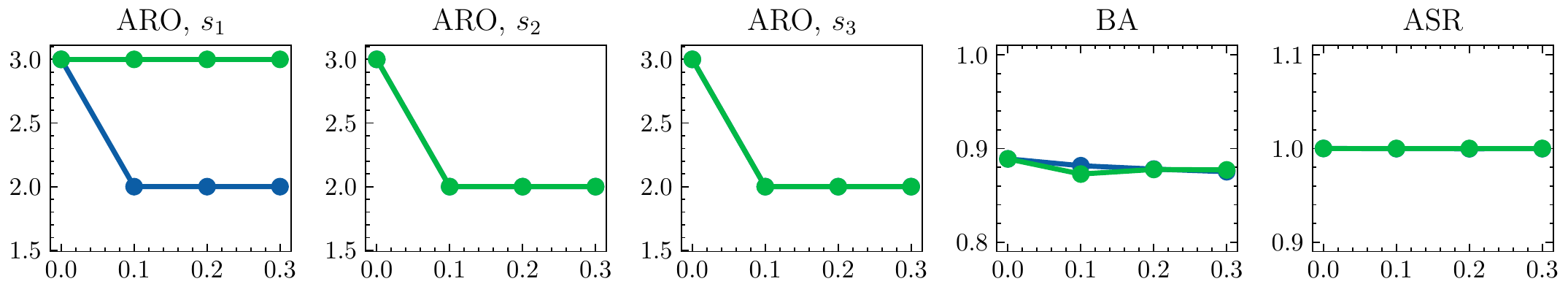}} \\
\subfloat[CelebA, Patched]{\includegraphics[width=8.8cm]{\impath/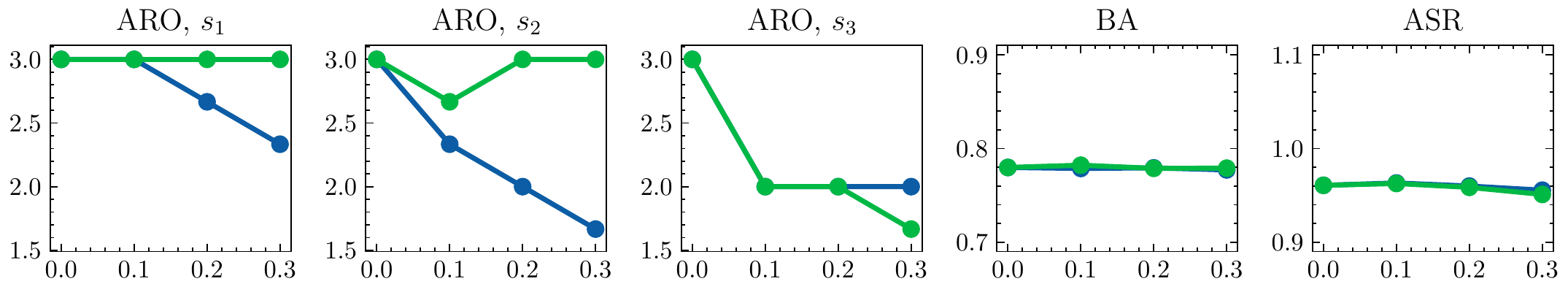}} \hfill
\subfloat[CelebA, Blended]{\includegraphics[width=8.8cm]{\impath/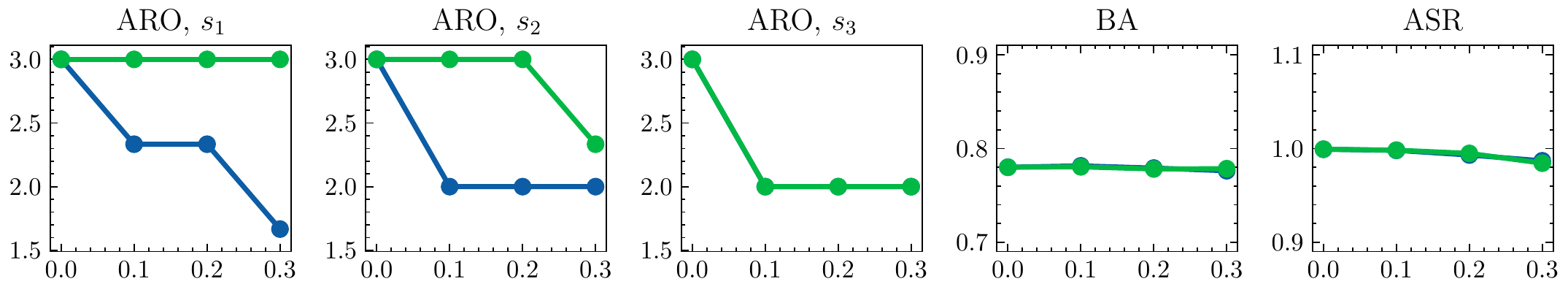}} \\
\subfloat[CelebA, SIG]{\includegraphics[width=8.8cm]{\impath/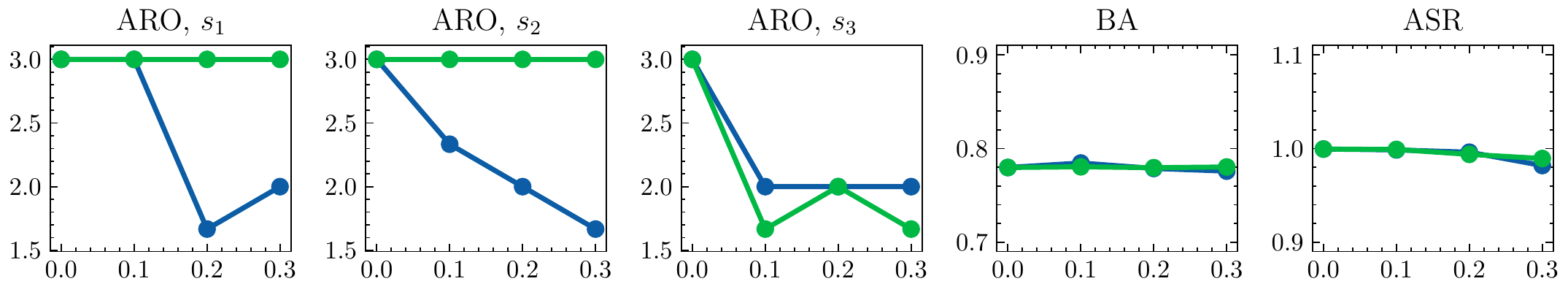}} \hfill 
\subfloat[CelebA, Warped]{\includegraphics[width=8.8cm]{\impath/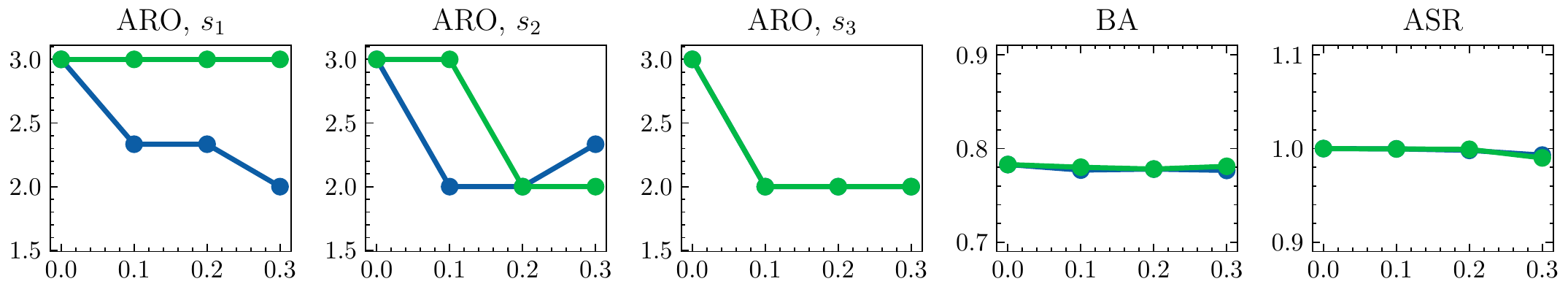}} \\
\subfloat{\includegraphics[height=0.5cm]{\impath/legend_mlmmdr.png}}
\caption{Results of SL-MMDR and ML-MMDR on the \{V-11\} models with different $\lambda$. X-axis: the value of $\lambda$. Y-axis: the value of each indicator.}
\label{fig:mlmmdr_v11} 
\end{figure*}

\begin{figure*}[!ht]
\centering
\subfloat[CIFAR-10, Patched]{\includegraphics[width=8.8cm]{\impath/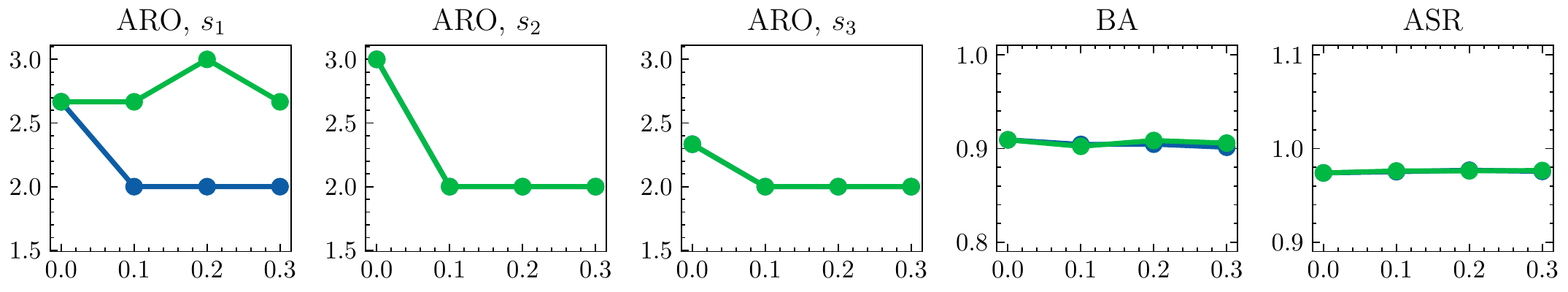}} \hfill
\subfloat[CIFAR-10, Blended]{\includegraphics[width=8.8cm]{\impath/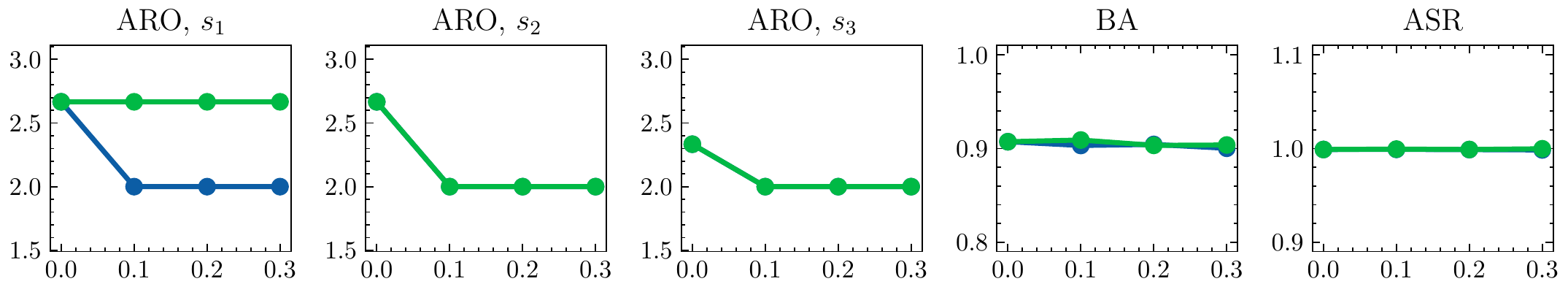}} \\
\subfloat[CIFAR-10, SIG]{\includegraphics[width=8.8cm]{\impath/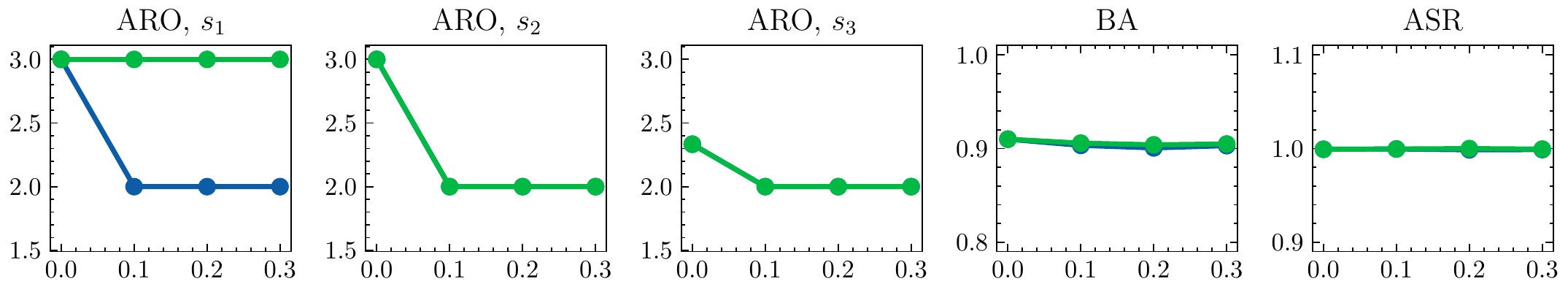}} \hfill 
\subfloat[CIFAR-10, Warped]{\includegraphics[width=8.8cm]{\impath/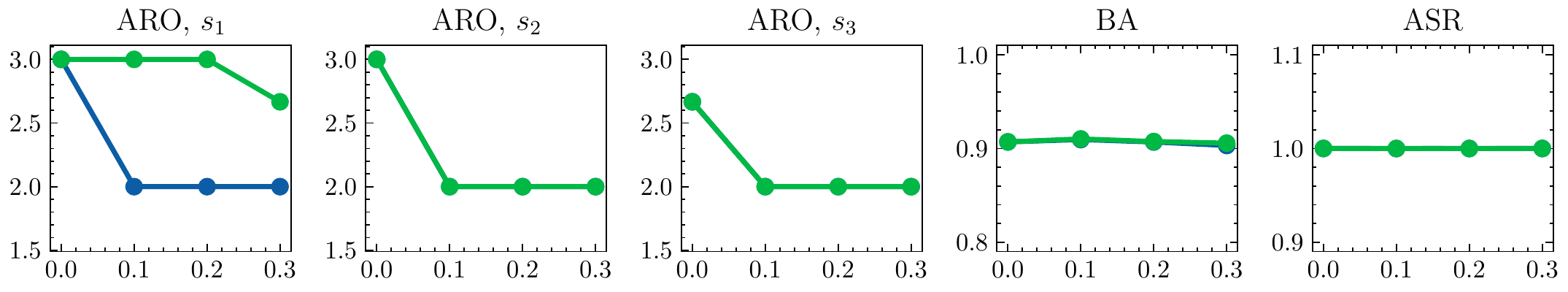}} \\
\subfloat[CelebA, Patched]{\includegraphics[width=8.8cm]{\impath/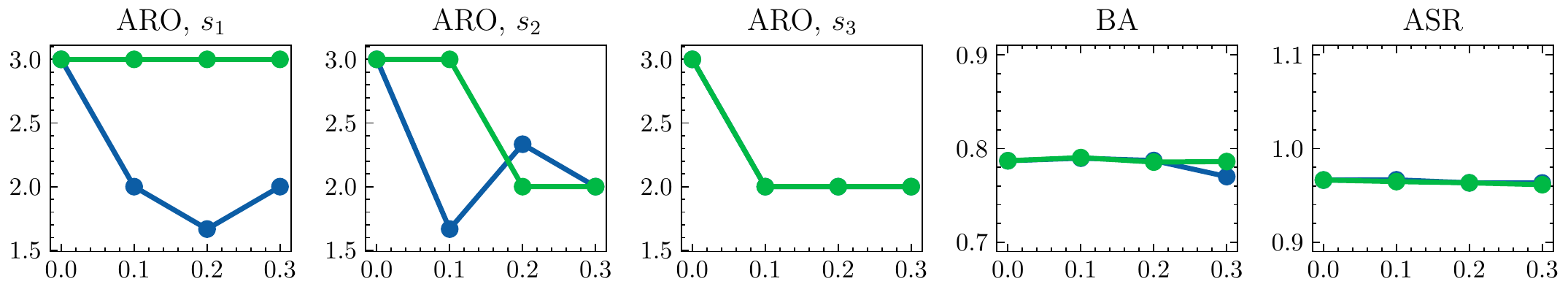}} \hfill
\subfloat[CelebA, Blended]{\includegraphics[width=8.8cm]{\impath/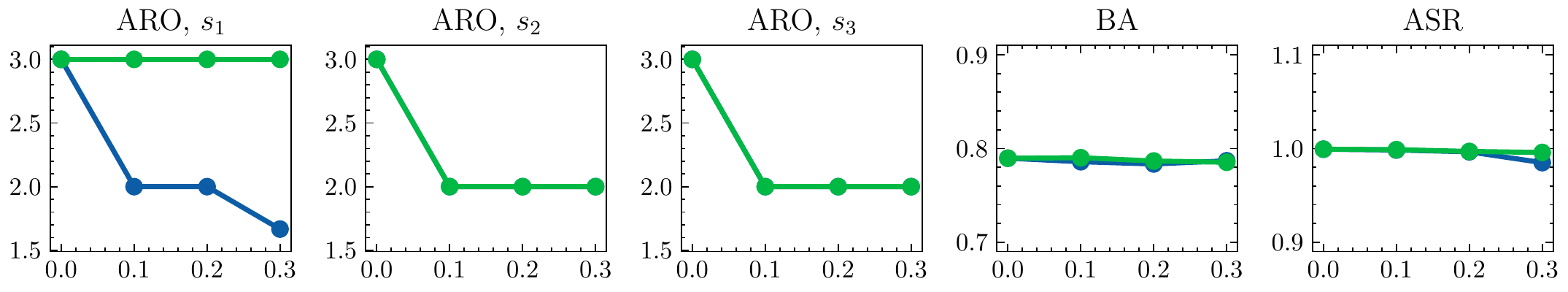}} \\
\subfloat[CelebA, SIG]{\includegraphics[width=8.8cm]{\impath/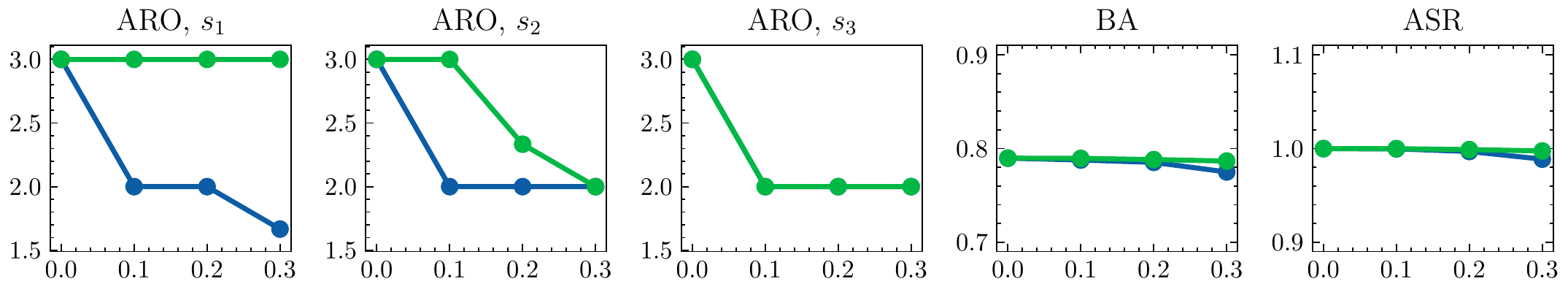}} \hfill 
\subfloat[CelebA, Warped]{\includegraphics[width=8.8cm]{\impath/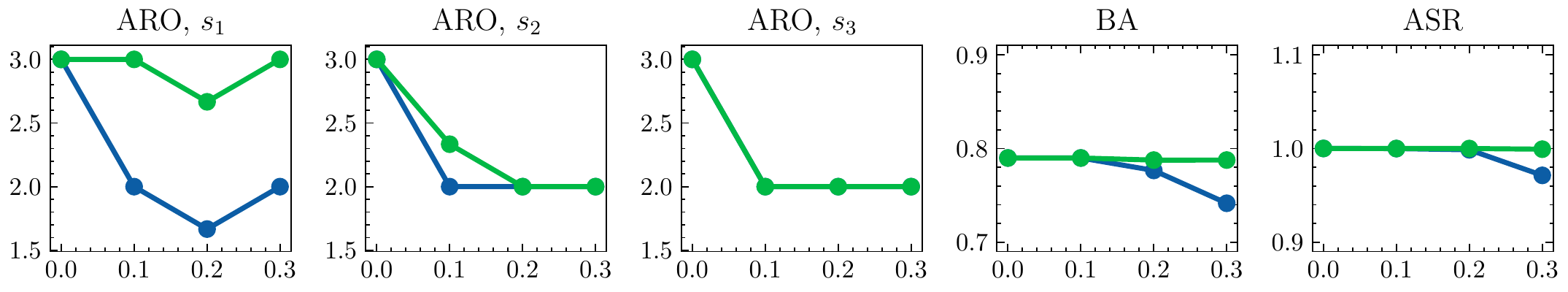}} \\
\subfloat{\includegraphics[height=0.5cm]{\impath/legend_mlmmdr.png}}
\caption{Results of SL-MMDR and ML-MMDR on the \{V-16\} models with different $\lambda$. X-axis: the value of $\lambda$. Y-axis: the value of each indicator.}
\label{fig:mlmmdr_v16} 
\end{figure*}

\begin{figure*}[!ht]
\centering
\subfloat[CIFAR-10, Patched]{\includegraphics[width=8.8cm]{\impath/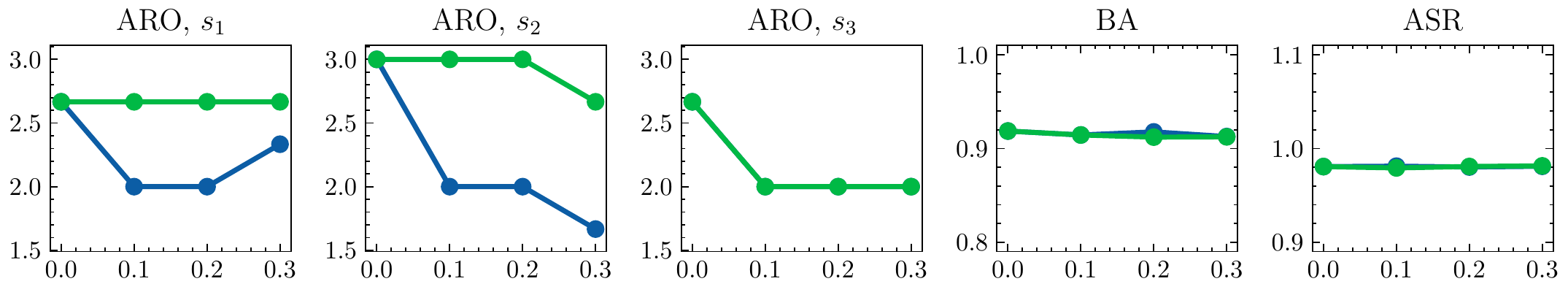}} \hfill
\subfloat[CIFAR-10, Blended]{\includegraphics[width=8.8cm]{\impath/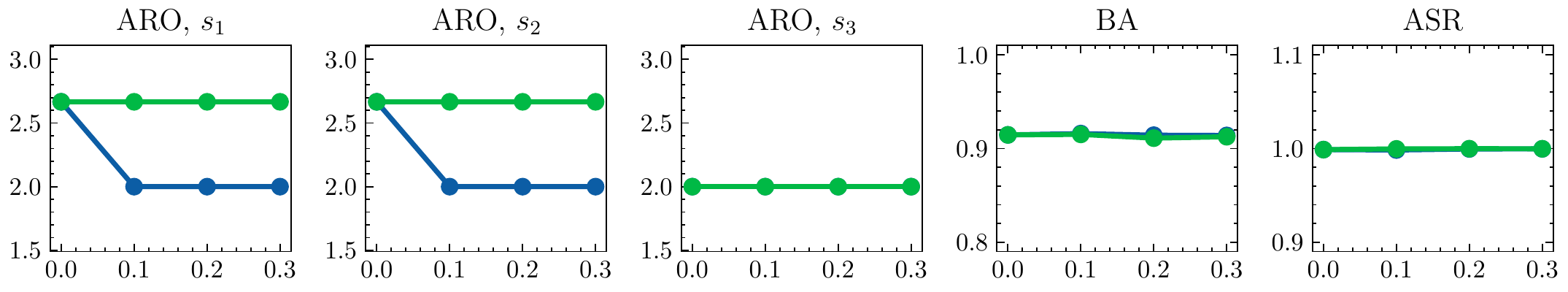}} \\
\subfloat[CIFAR-10, SIG]{\includegraphics[width=8.8cm]{\impath/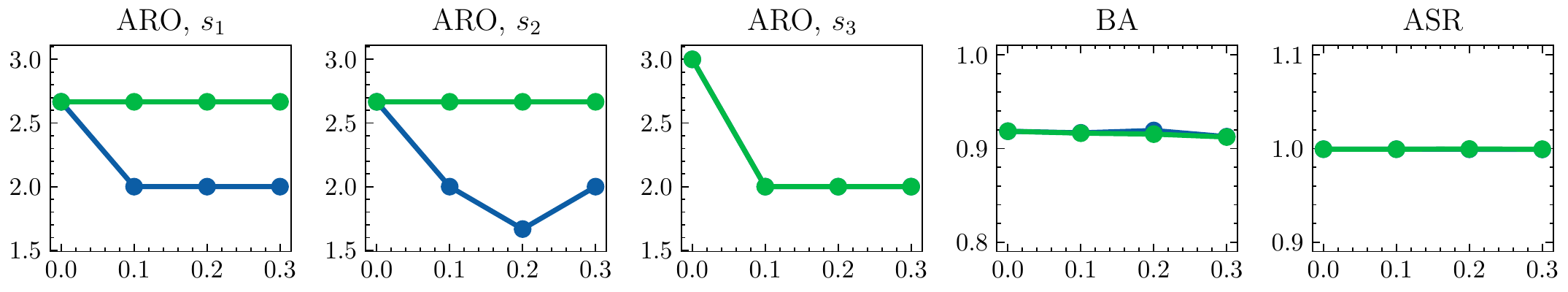}} \hfill 
\subfloat[CIFAR-10, Warped]{\includegraphics[width=8.8cm]{\impath/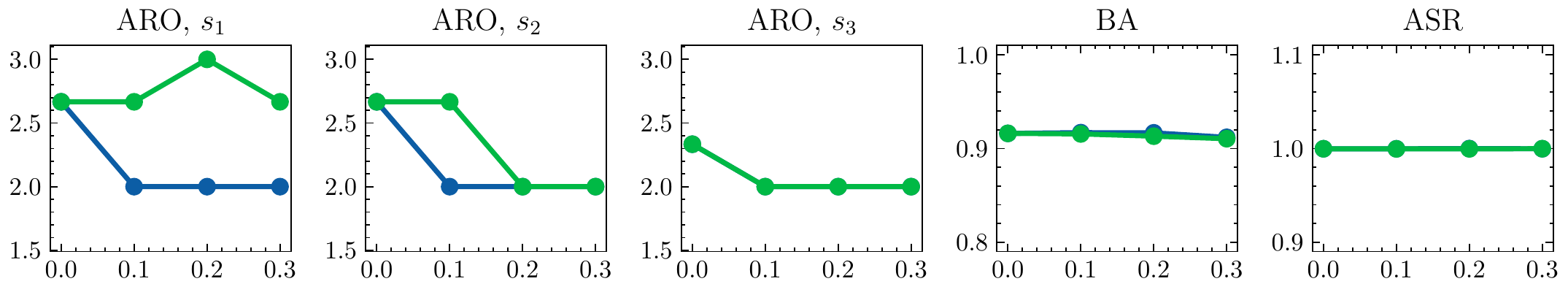}} \\
\subfloat[CelebA, Patched]{\includegraphics[width=8.8cm]{\impath/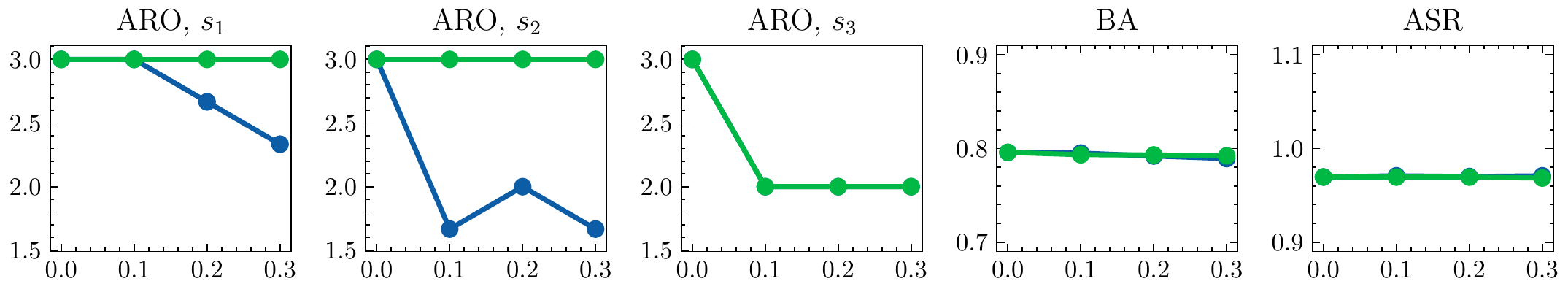}} \hfill
\subfloat[CelebA, Blended]{\includegraphics[width=8.8cm]{\impath/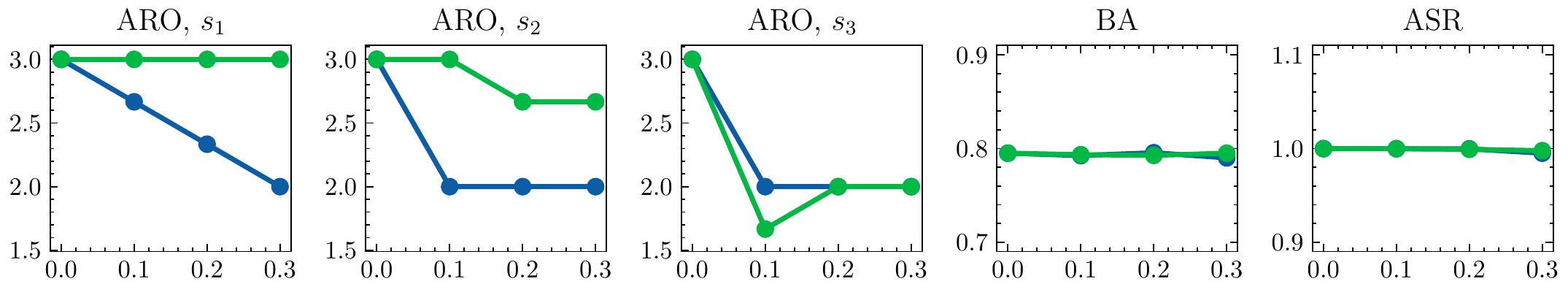}} \\
\subfloat[CelebA, SIG]{\includegraphics[width=8.8cm]{\impath/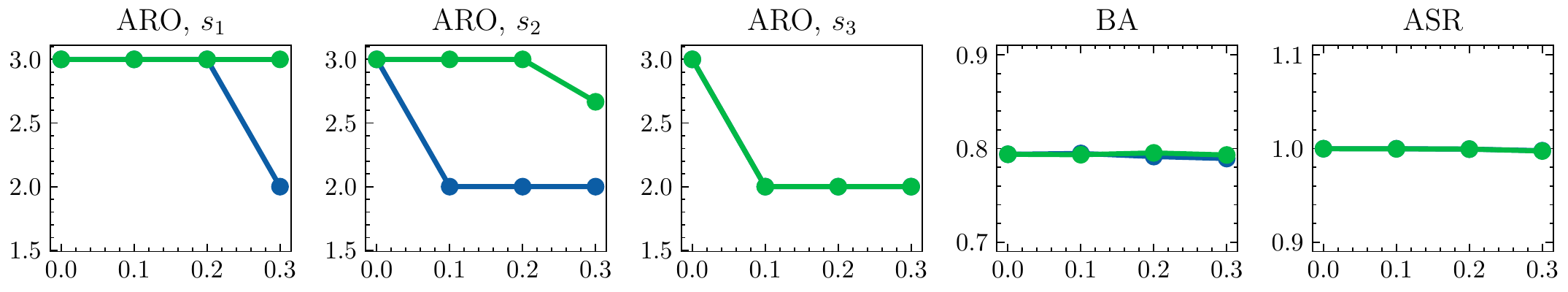}} \hfill 
\subfloat[CelebA, Warped]{\includegraphics[width=8.8cm]{\impath/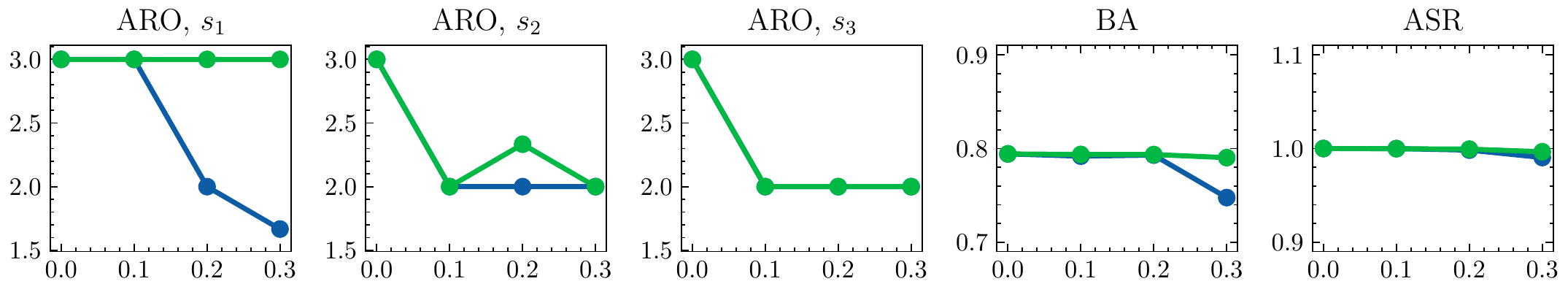}} \\
\subfloat{\includegraphics[height=0.5cm]{\impath/legend_mlmmdr.png}}
\caption{Results of SL-MMDR and ML-MMDR on the \{R-18\} models with different $\lambda$. X-axis: the value of $\lambda$. Y-axis: the value of each indicator.}
\label{fig:mlmmdr_r18} 
\end{figure*}

The results of SL-MMDR and ML-MMDR on the \{V-11\}, \{V-16\}, and \{R-18\} models with different $\lambda$ are shown in Fig. \ref{fig:mlmmdr_v11}, Fig. \ref{fig:mlmmdr_v16}, and Fig. \ref{fig:mlmmdr_r18}, respectively.

\section{Performance of AC, SS, and SR} \label{app:per_def}
\begin{figure*}[!ht]
\centering
\subfloat[CIFAR-10, Patched]{\includegraphics[width=8.8cm]{\impath/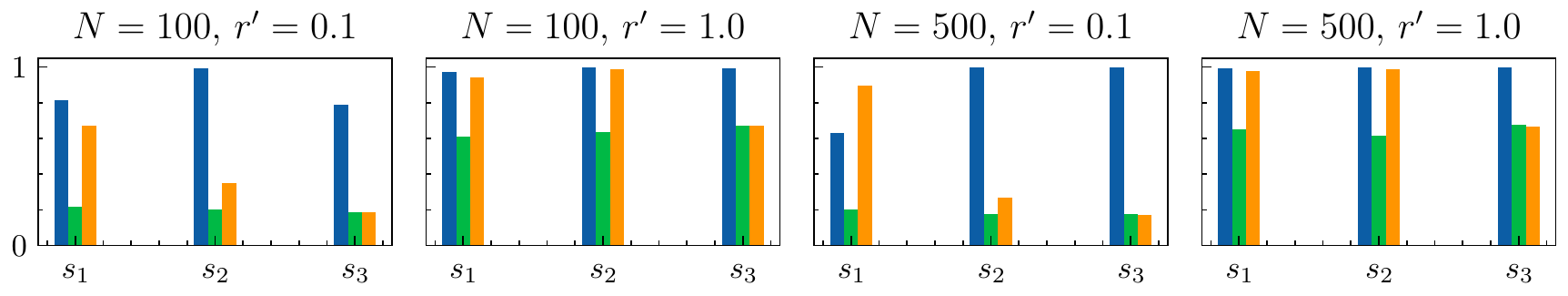}} \hfill
\subfloat[CIFAR-10, Blended]{\includegraphics[width=8.8cm]{\impath/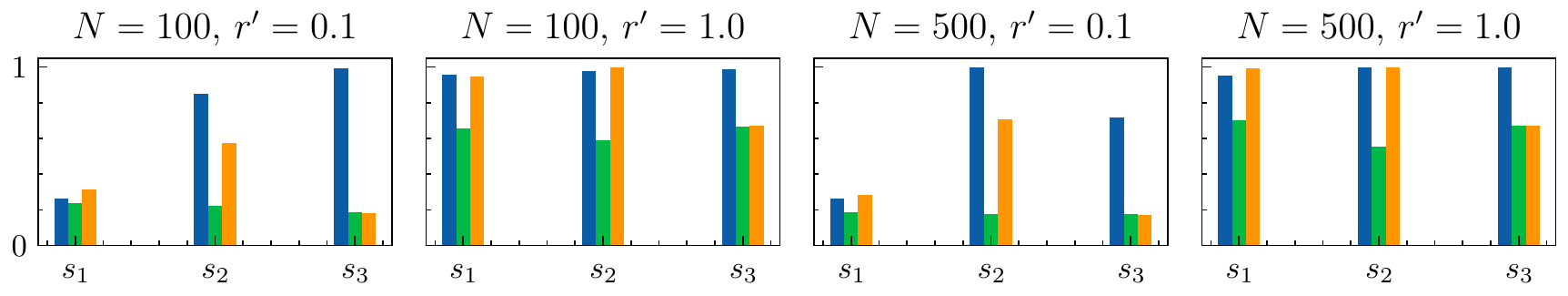}} \\
\subfloat[CIFAR-10, SIG]{\includegraphics[width=8.8cm]{\impath/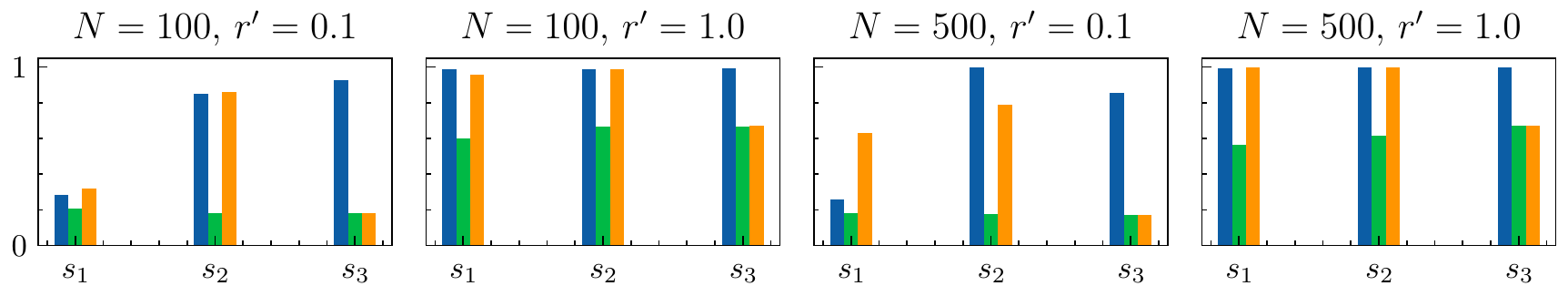}} \hfill 
\subfloat[CIFAR-10, Warped]{\includegraphics[width=8.8cm]{\impath/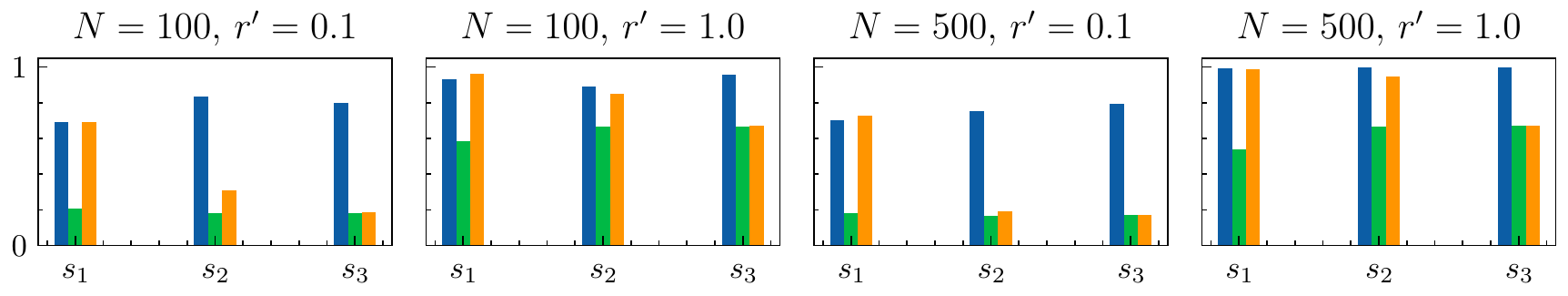}} \\
\subfloat[CelebA, Patched]{\includegraphics[width=8.8cm]{\impath/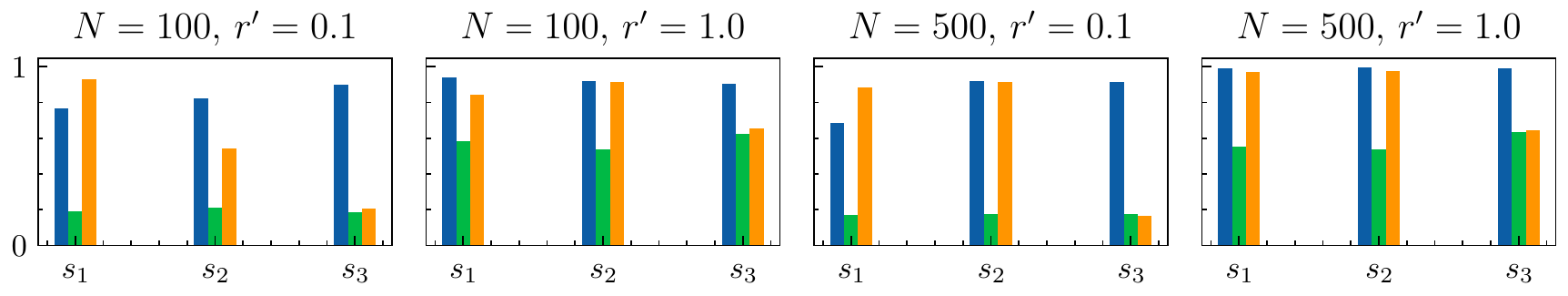}} \hfill
\subfloat[CelebA, Blended]{\includegraphics[width=8.8cm]{\impath/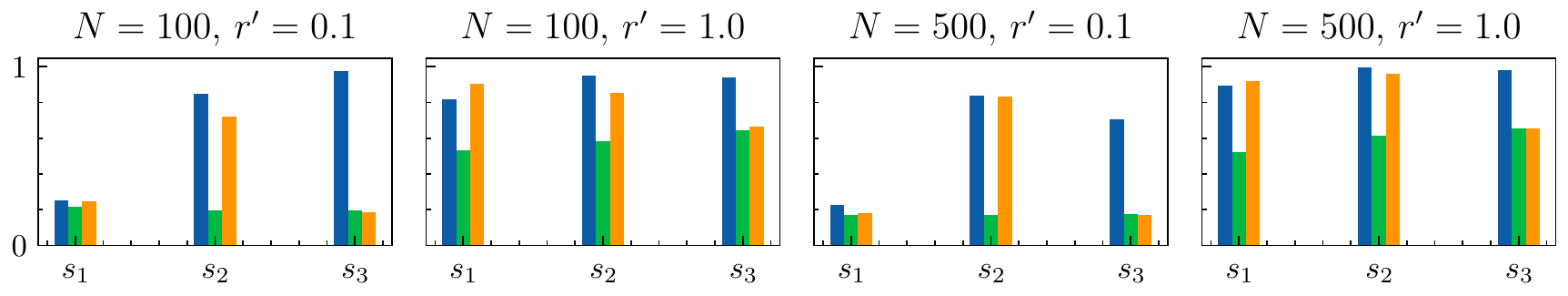}} \\
\subfloat[CelebA, SIG]{\includegraphics[width=8.8cm]{\impath/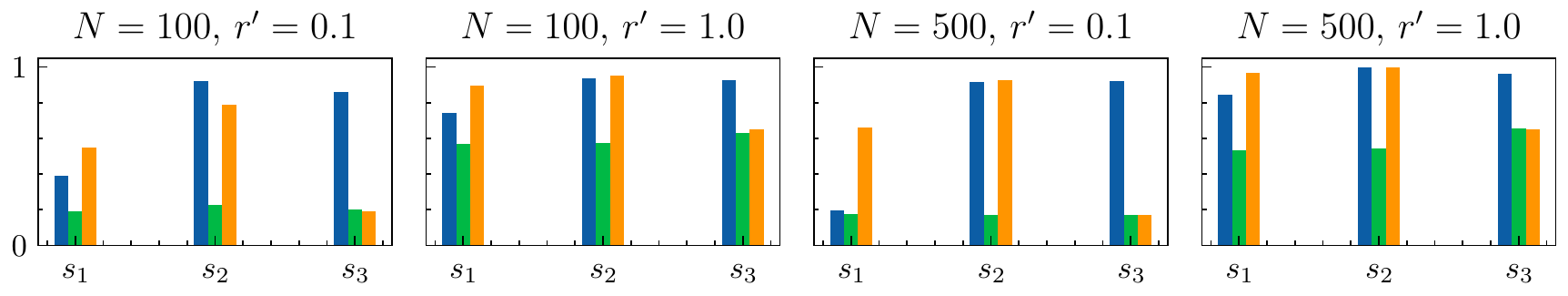}} \hfill 
\subfloat[CelebA, Warped]{\includegraphics[width=8.8cm]{\impath/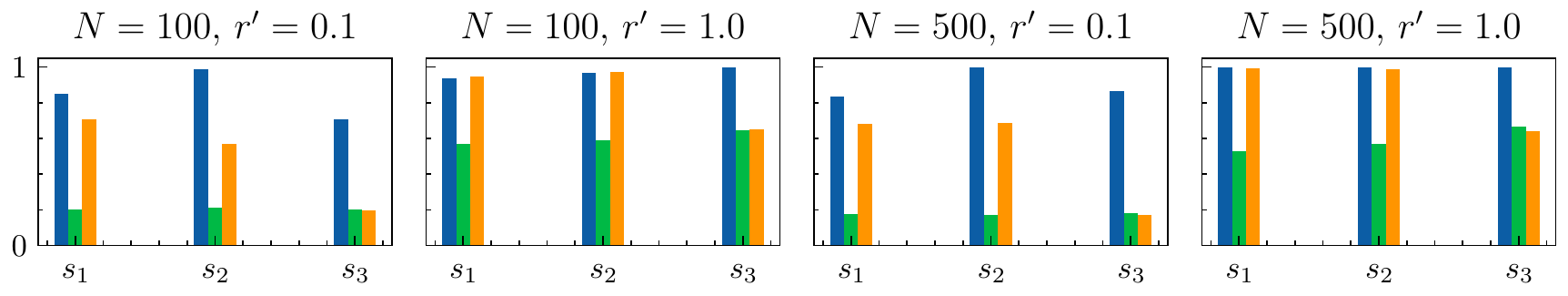}} \\
\subfloat{\includegraphics[height=0.5cm]{\impath/legend_defense.png}}
\caption{F1 scores of AC on the \{V-11\} models. X-axis: the level of features. Y-axis: the value of $F1$.}
\label{fig:ac_v11} 
\end{figure*}

\begin{figure*}[!ht]
\centering
\subfloat[CIFAR-10, Patched]{\includegraphics[width=8.8cm]{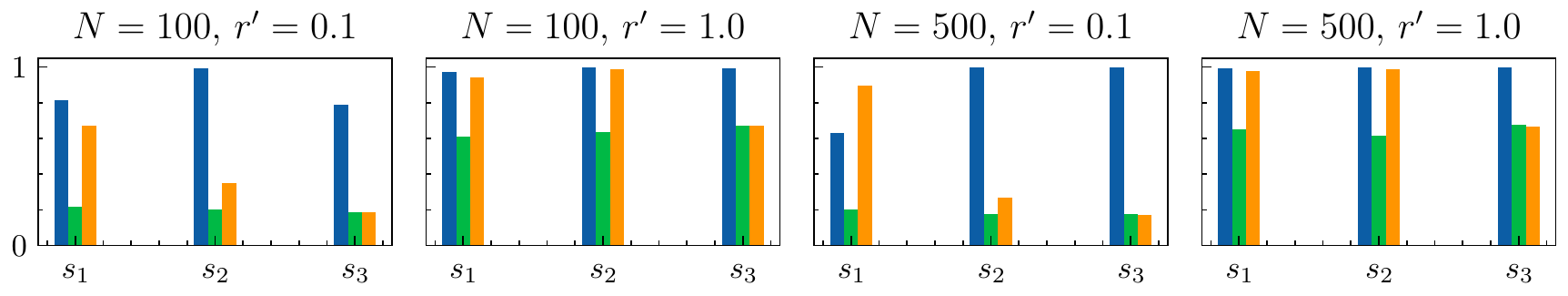}} \hfill
\subfloat[CIFAR-10, Blended]{\includegraphics[width=8.8cm]{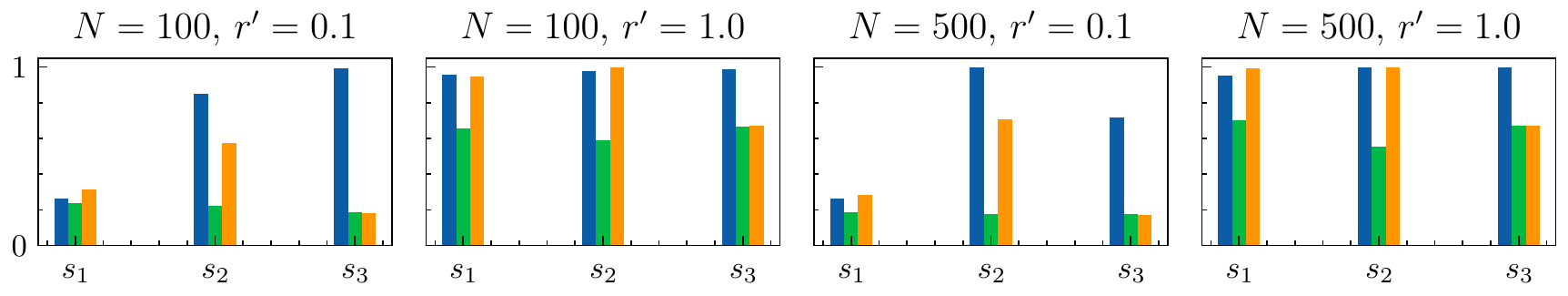}} \\
\subfloat[CIFAR-10, SIG]{\includegraphics[width=8.8cm]{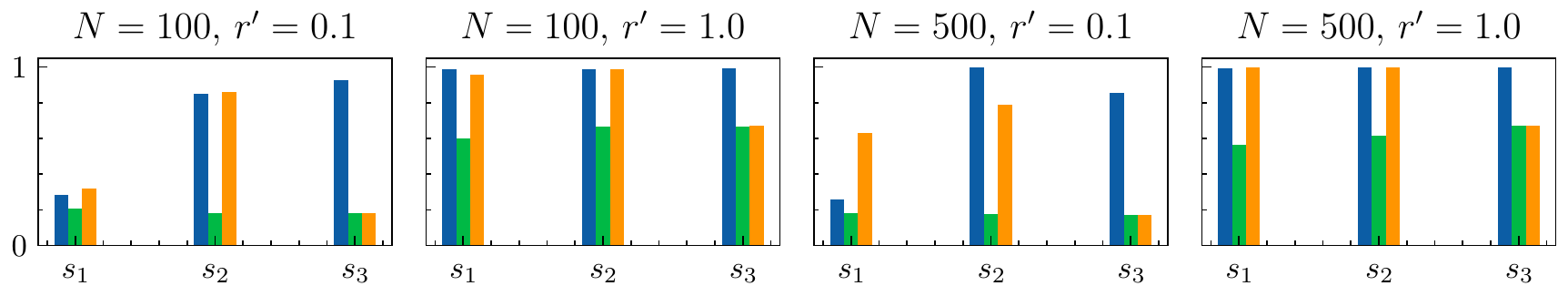}} \hfill 
\subfloat[CIFAR-10, Warped]{\includegraphics[width=8.8cm]{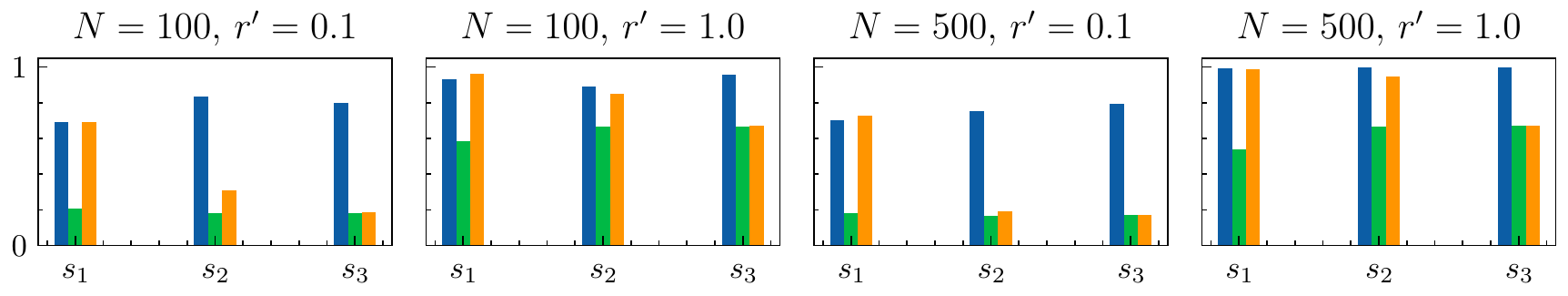}} \\
\subfloat[CelebA, Patched]{\includegraphics[width=8.8cm]{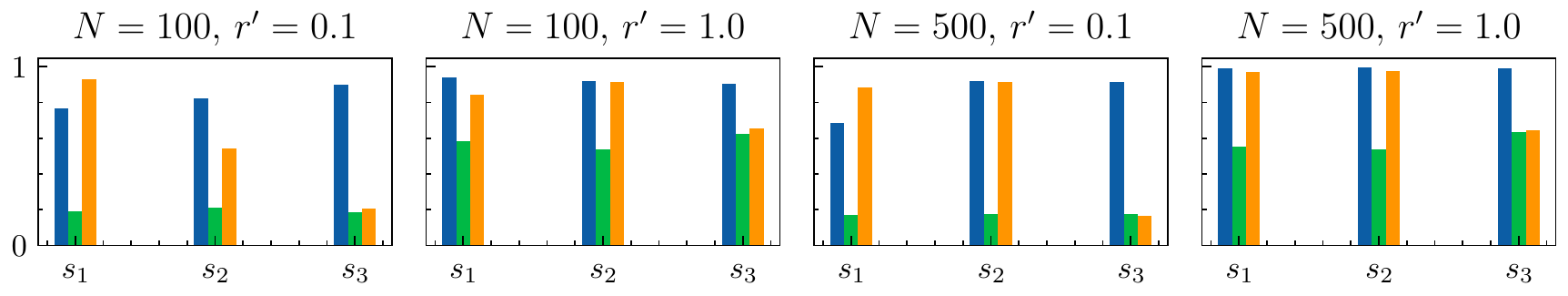}} \hfill
\subfloat[CelebA, Blended]{\includegraphics[width=8.8cm]{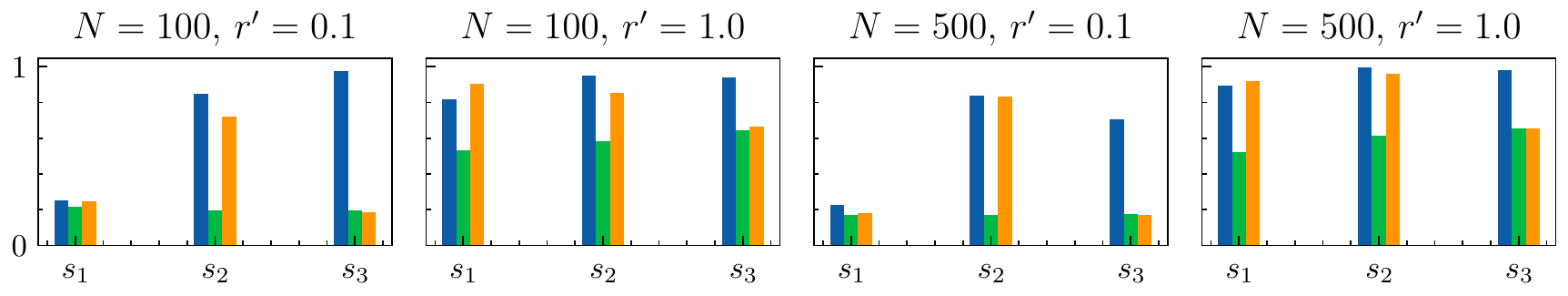}} \\
\subfloat[CelebA, SIG]{\includegraphics[width=8.8cm]{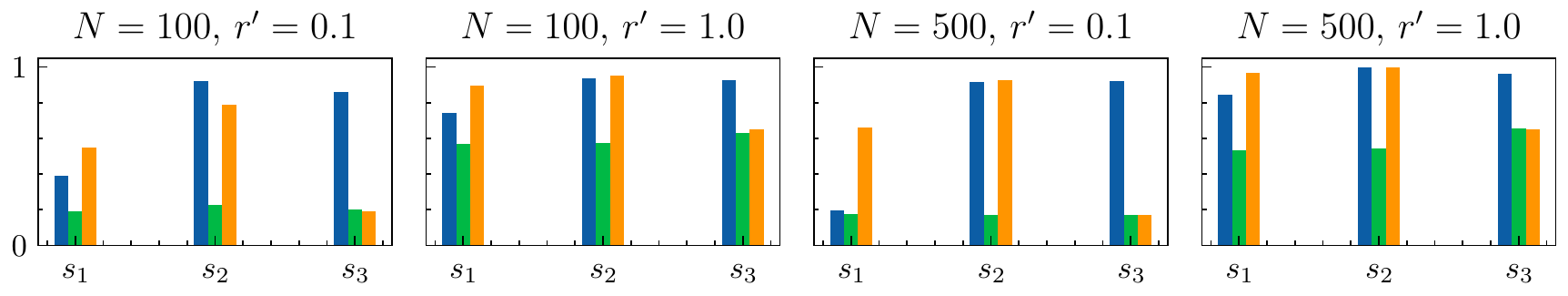}} \hfill 
\subfloat[CelebA, Warped]{\includegraphics[width=8.8cm]{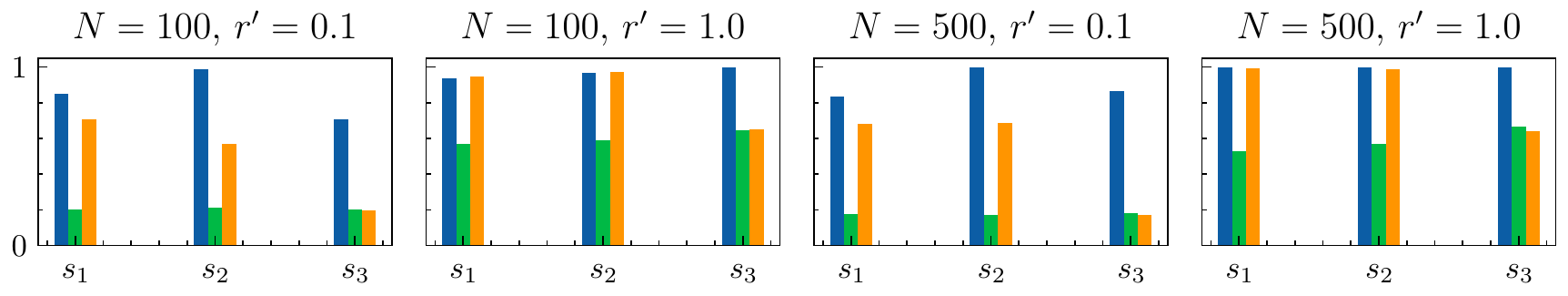}} \\
\subfloat{\includegraphics[height=0.5cm]{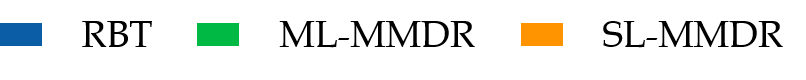}}
\caption{F1 scores of AC on the \{V-16\} models. X-axis: the level of features. Y-axis: the value of $F1$.}
\label{fig:ac_v16} 
\end{figure*}

\begin{figure*}[!ht]
\centering
\subfloat[CIFAR-10, Patched]{\includegraphics[width=8.8cm]{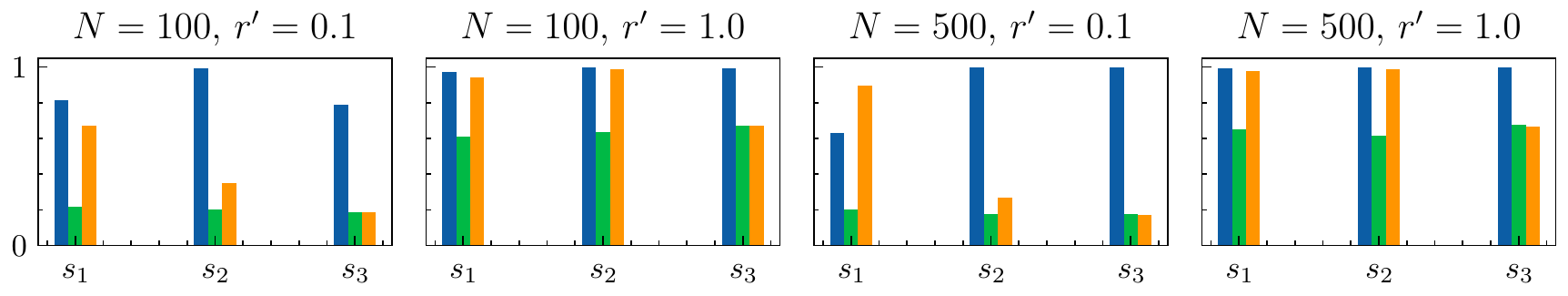}} \hfill
\subfloat[CIFAR-10, Blended]{\includegraphics[width=8.8cm]{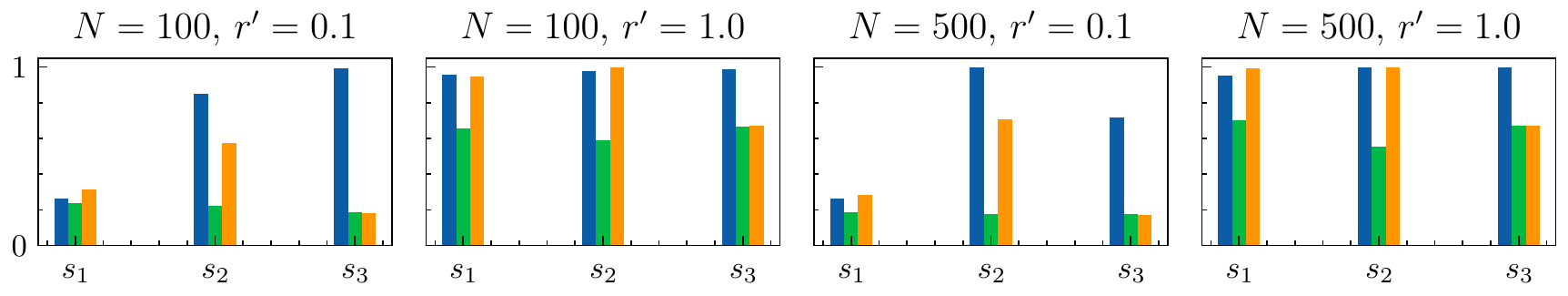}} \\
\subfloat[CIFAR-10, SIG]{\includegraphics[width=8.8cm]{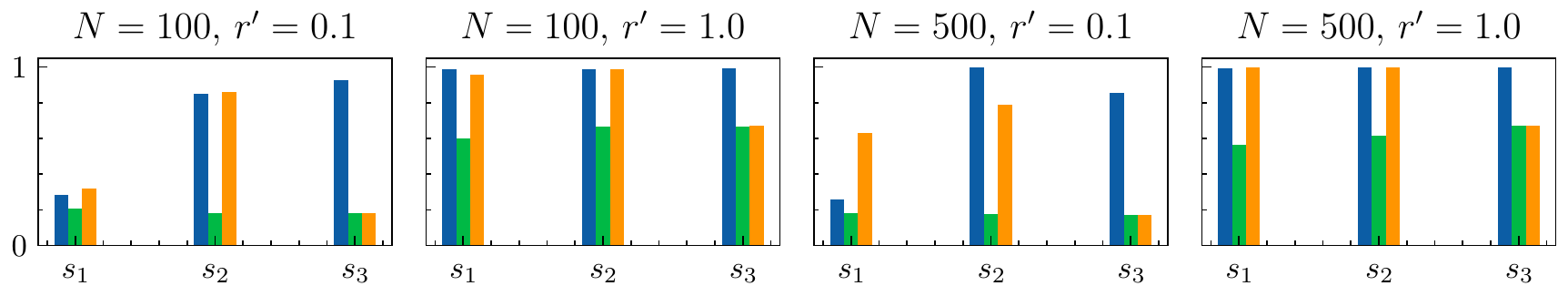}} \hfill 
\subfloat[CIFAR-10, Warped]{\includegraphics[width=8.8cm]{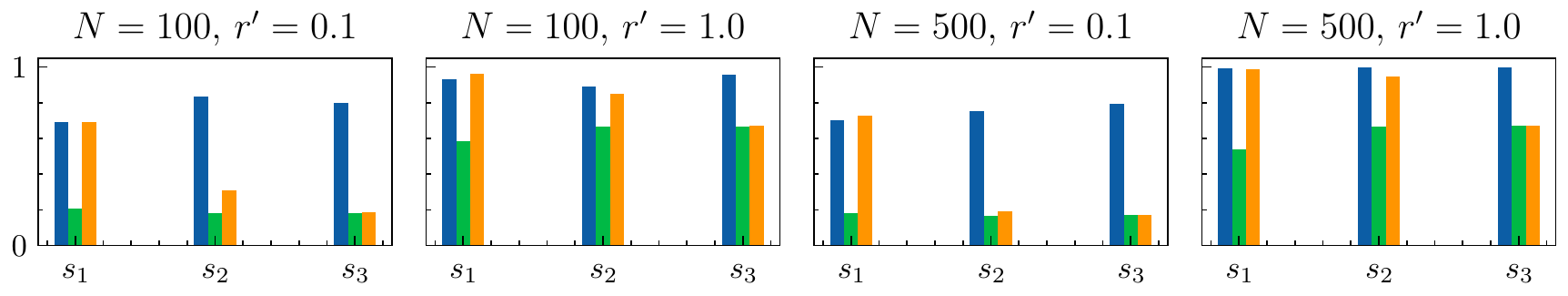}} \\
\subfloat[CelebA, Patched]{\includegraphics[width=8.8cm]{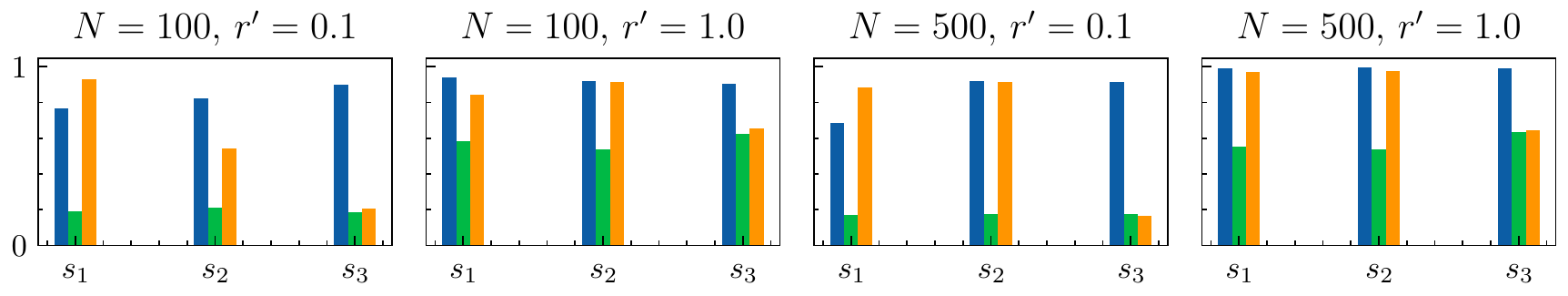}} \hfill
\subfloat[CelebA, Blended]{\includegraphics[width=8.8cm]{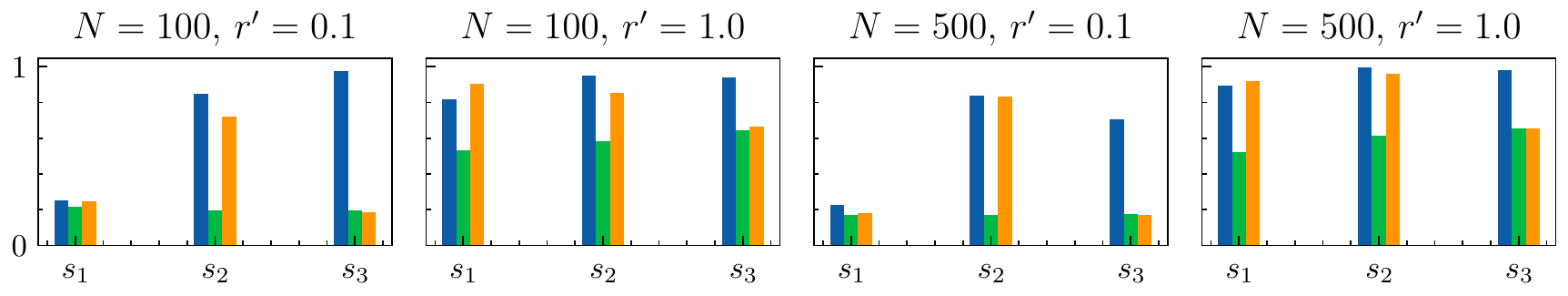}} \\
\subfloat[CelebA, SIG]{\includegraphics[width=8.8cm]{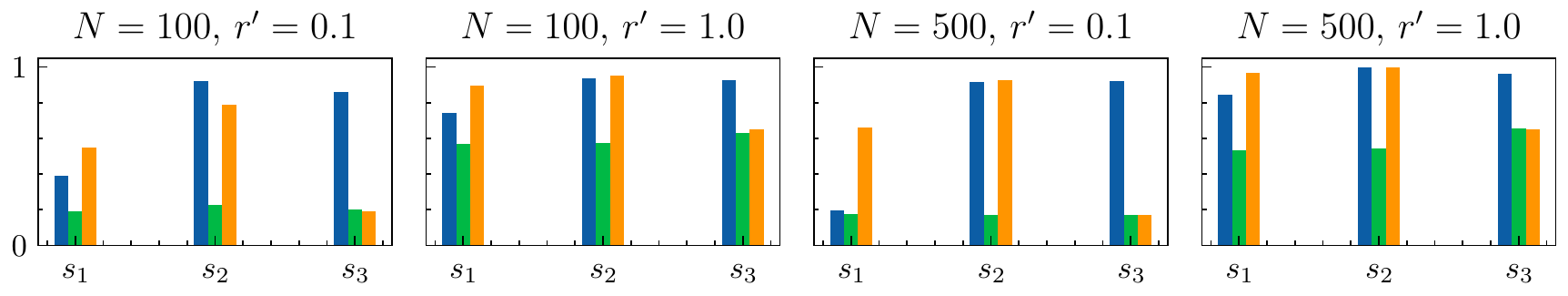}} \hfill 
\subfloat[CelebA, Warped]{\includegraphics[width=8.8cm]{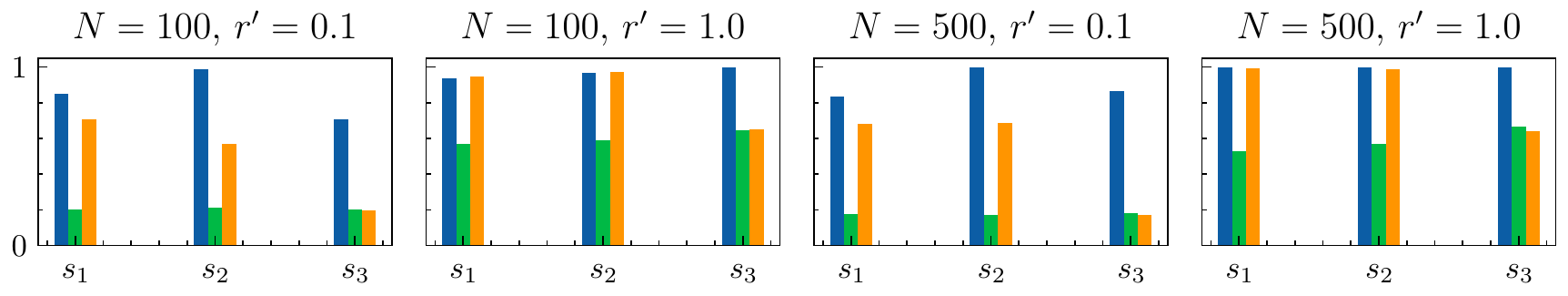}} \\
\subfloat{\includegraphics[height=0.5cm]{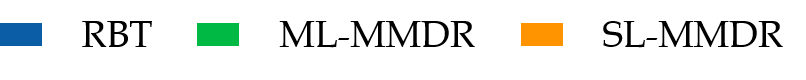}}
\caption{F1 scores of AC on the \{R-18\} models. X-axis: the level of features. Y-axis: the value of $F1$.}
\label{fig:ac_r18} 
\end{figure*}

\begin{figure*}[!ht]
\centering
\subfloat[CIFAR-10, Patched]{\includegraphics[width=8.8cm]{\impath/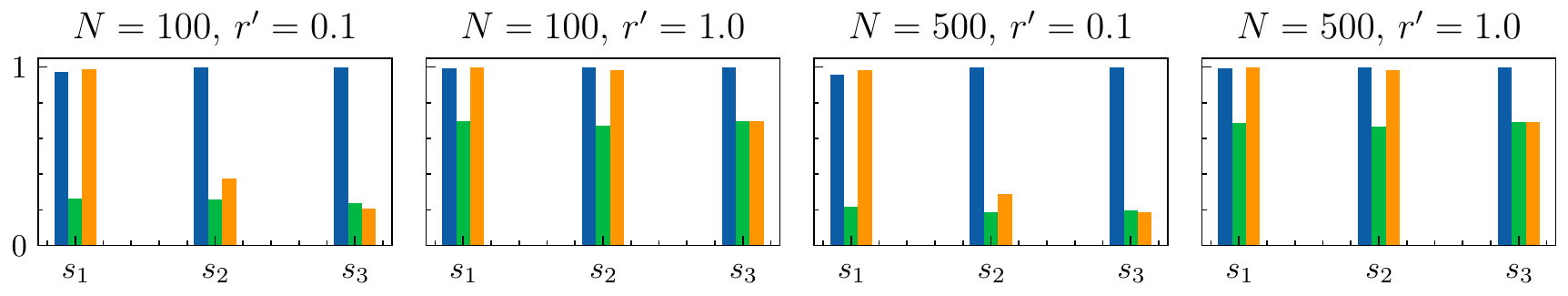}} \hfill
\subfloat[CIFAR-10, Blended]{\includegraphics[width=8.8cm]{\impath/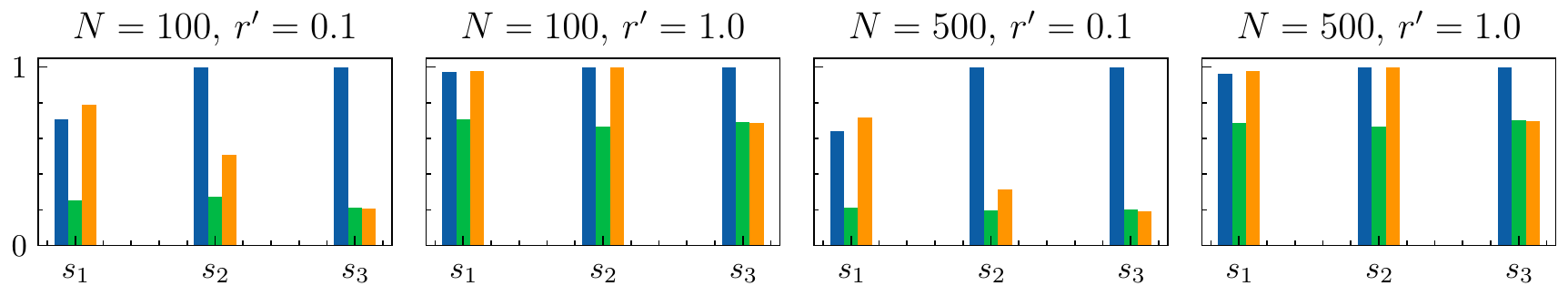}} \\
\subfloat[CIFAR-10, SIG]{\includegraphics[width=8.8cm]{\impath/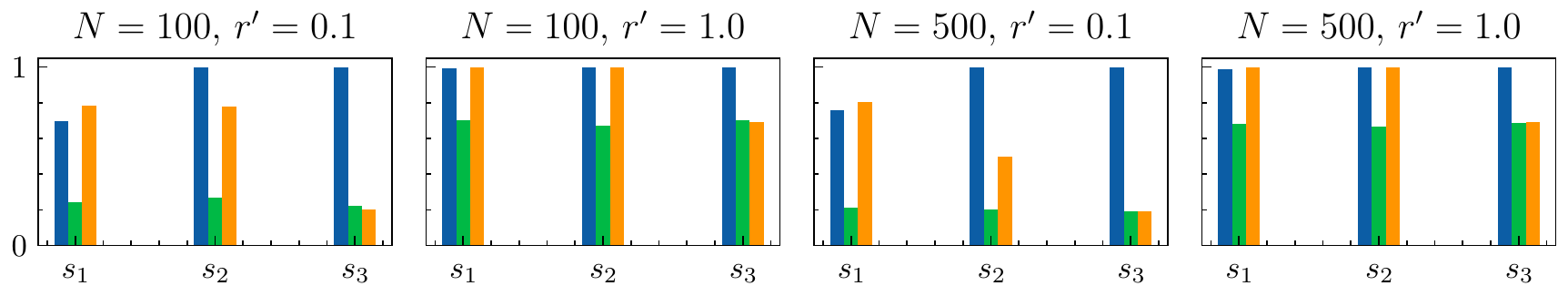}} \hfill 
\subfloat[CIFAR-10, Warped]{\includegraphics[width=8.8cm]{\impath/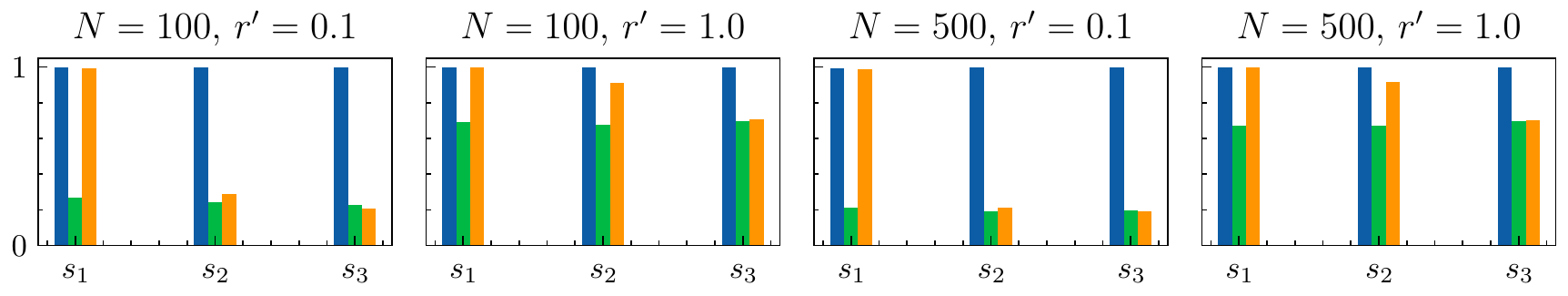}} \\
\subfloat[CelebA, Patched]{\includegraphics[width=8.8cm]{\impath/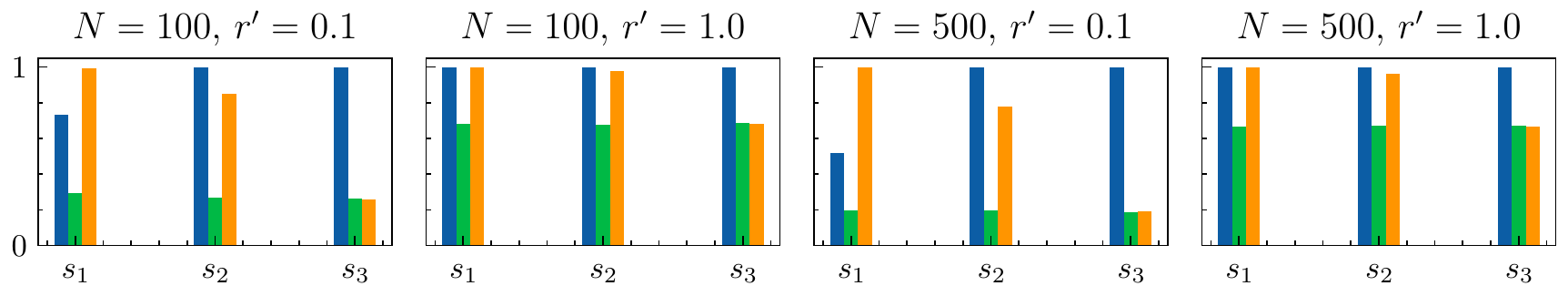}} \hfill
\subfloat[CelebA, Blended]{\includegraphics[width=8.8cm]{\impath/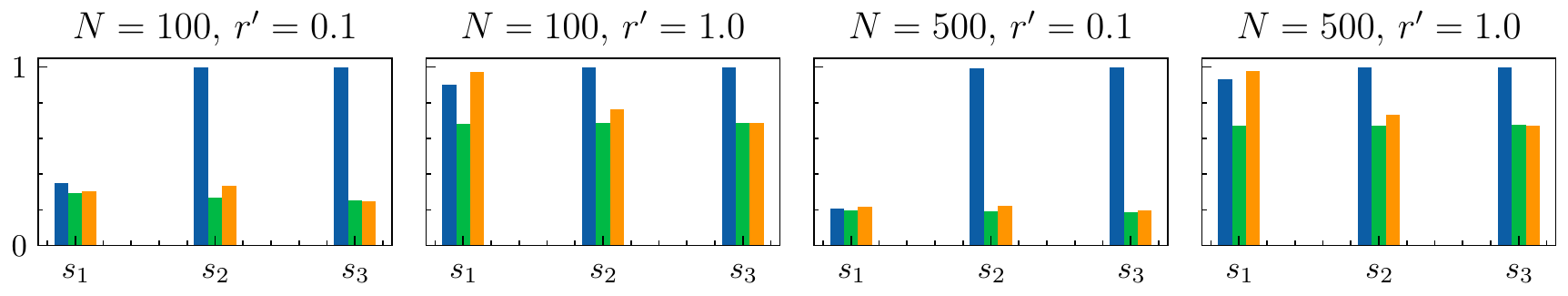}} \\
\subfloat[CelebA, SIG]{\includegraphics[width=8.8cm]{\impath/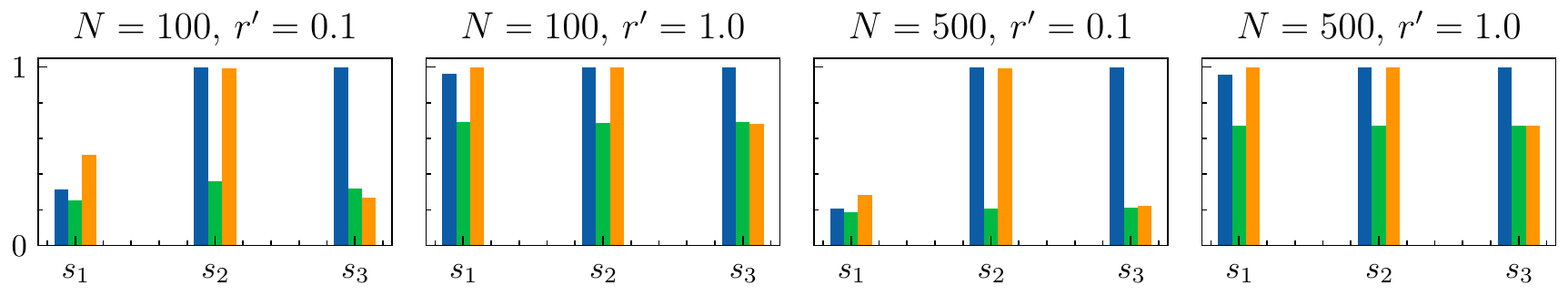}} \hfill 
\subfloat[CelebA, Warped]{\includegraphics[width=8.8cm]{\impath/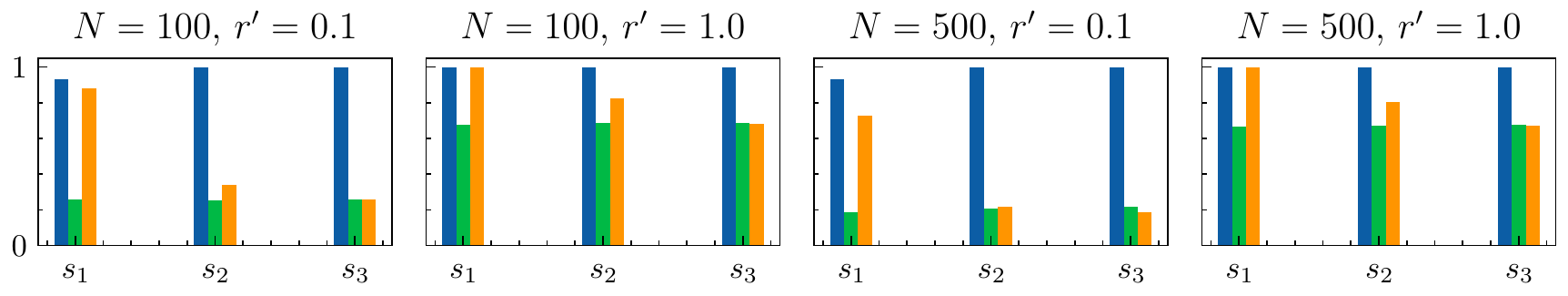}} \\
\subfloat{\includegraphics[height=0.5cm]{\impath/legend_defense.png}}
\caption{F1 scores of SS on the \{V-11\} models. X-axis: the level of features. Y-axis: the value of $F1$.}
\label{fig:ss_v11} 
\end{figure*}

\begin{figure*}[!ht]
\centering
\subfloat[CIFAR-10, Patched]{\includegraphics[width=8.8cm]{\impath/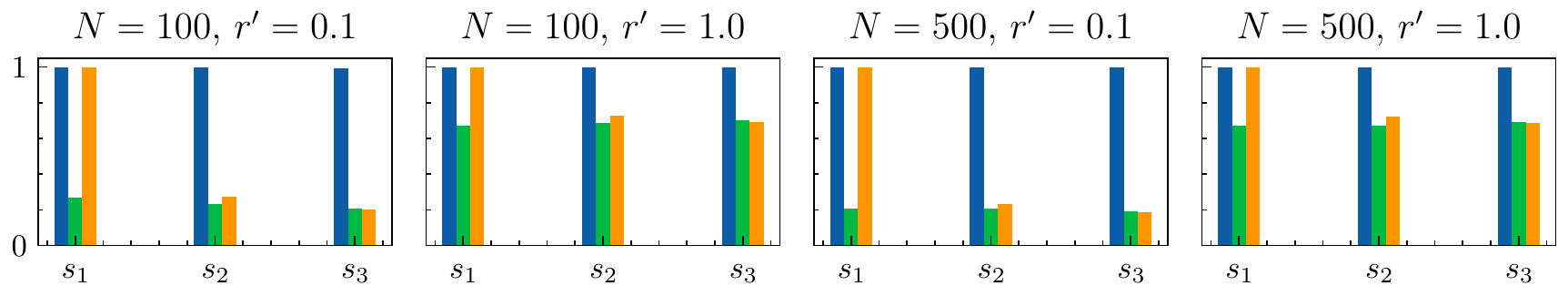}} \hfill
\subfloat[CIFAR-10, Blended]{\includegraphics[width=8.8cm]{\impath/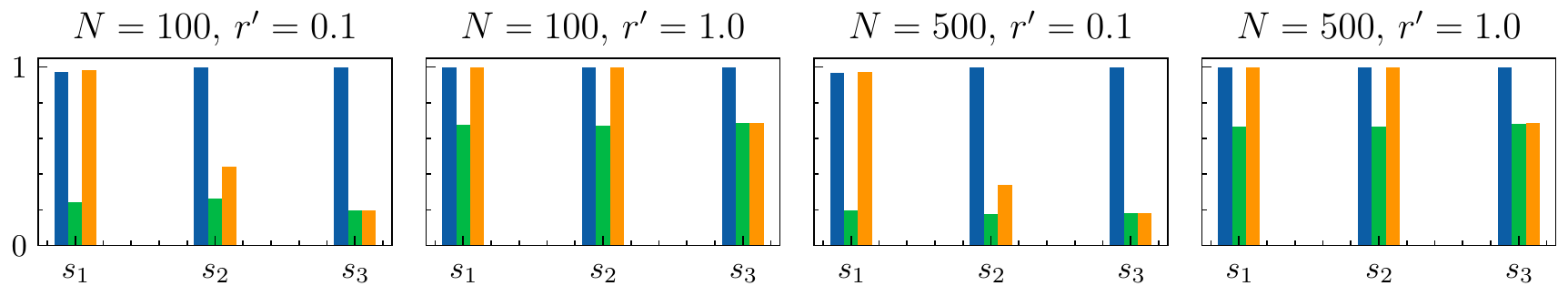}} \\
\subfloat[CIFAR-10, SIG]{\includegraphics[width=8.8cm]{\impath/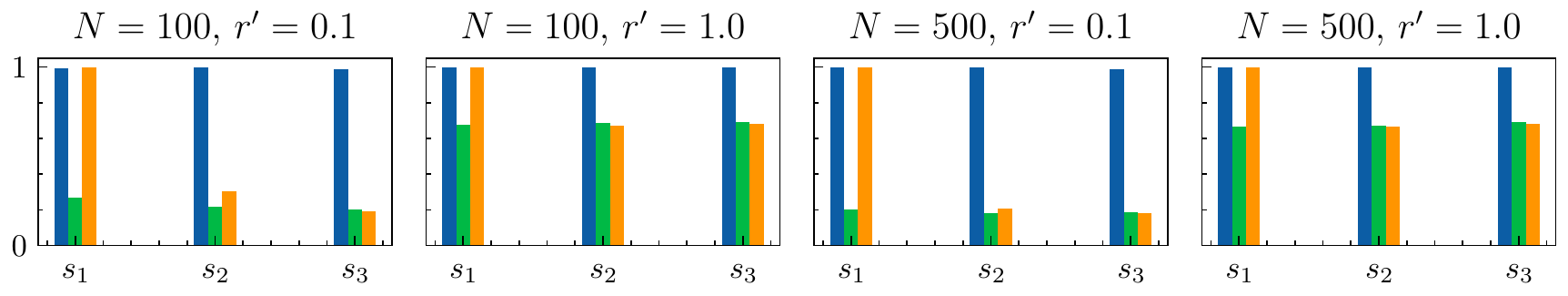}} \hfill 
\subfloat[CIFAR-10, Warped]{\includegraphics[width=8.8cm]{\impath/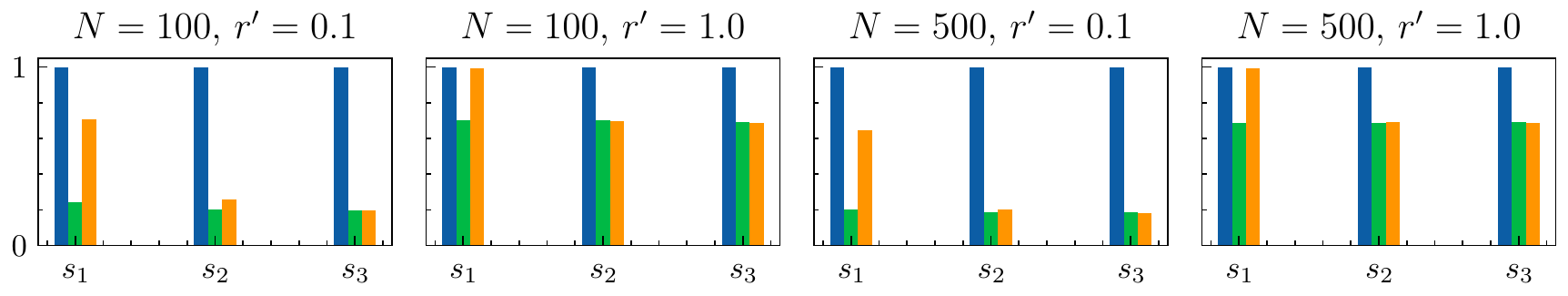}} \\
\subfloat[CelebA, Patched]{\includegraphics[width=8.8cm]{\impath/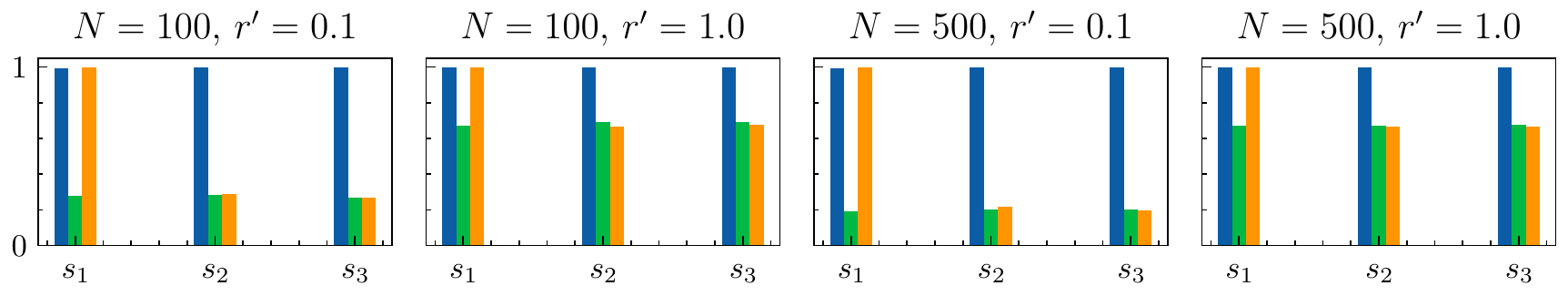}} \hfill
\subfloat[CelebA, Blended]{\includegraphics[width=8.8cm]{\impath/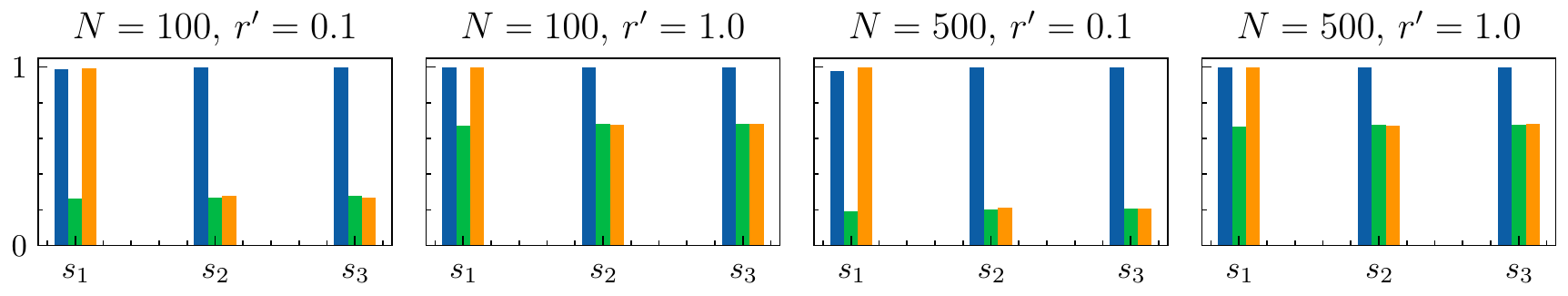}} \\
\subfloat[CelebA, SIG]{\includegraphics[width=8.8cm]{\impath/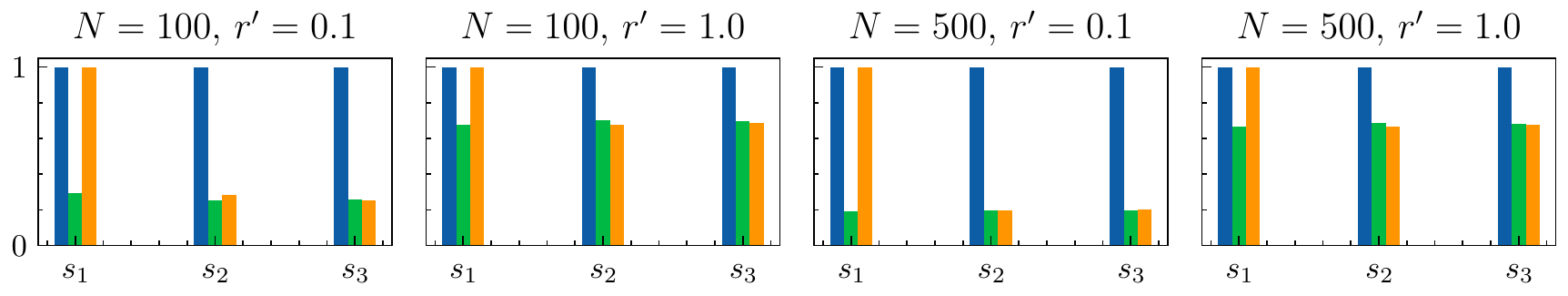}} \hfill 
\subfloat[CelebA, Warped]{\includegraphics[width=8.8cm]{\impath/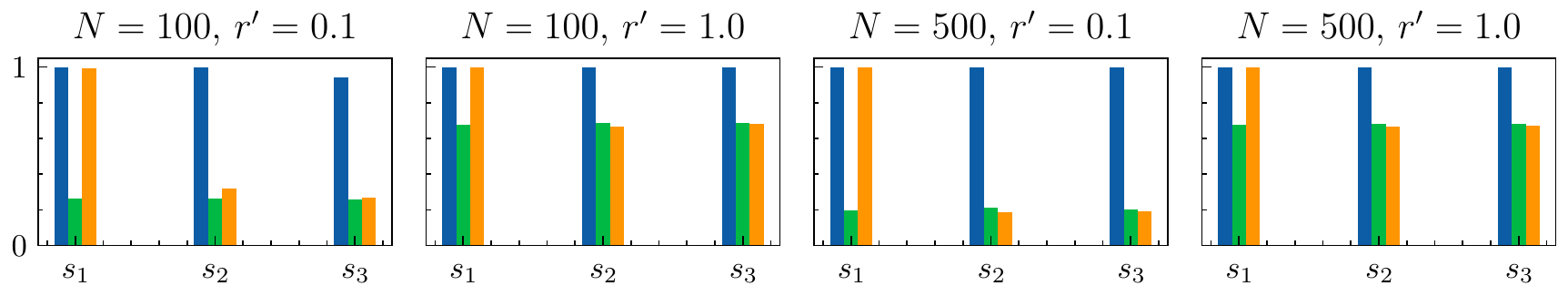}} \\
\subfloat{\includegraphics[height=0.5cm]{\impath/legend_defense.png}}
\caption{F1 scores of SS on the \{V-16\} models. X-axis: the level of features. Y-axis: the value of $F1$.}
\label{fig:ss_v16} 
\end{figure*}

\begin{figure*}[!ht]
\centering
\subfloat[CIFAR-10, Patched]{\includegraphics[width=8.8cm]{\impath/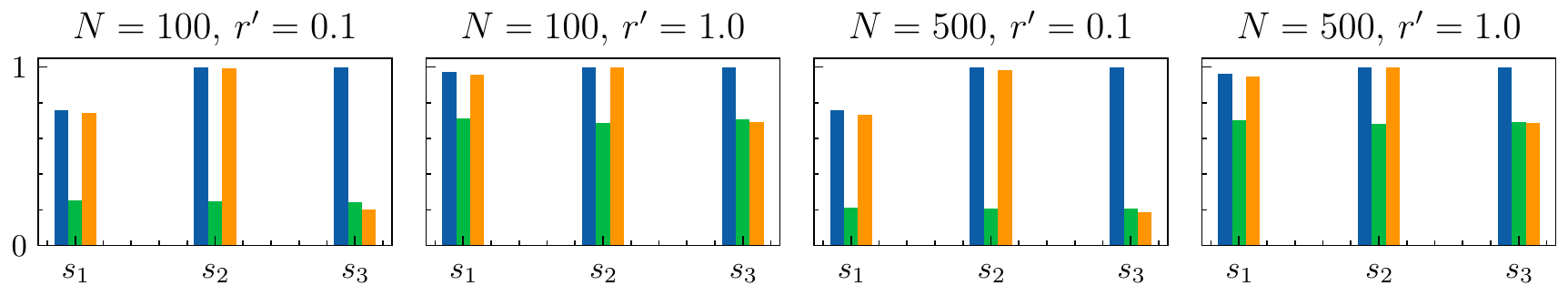}} \hfill
\subfloat[CIFAR-10, Blended]{\includegraphics[width=8.8cm]{\impath/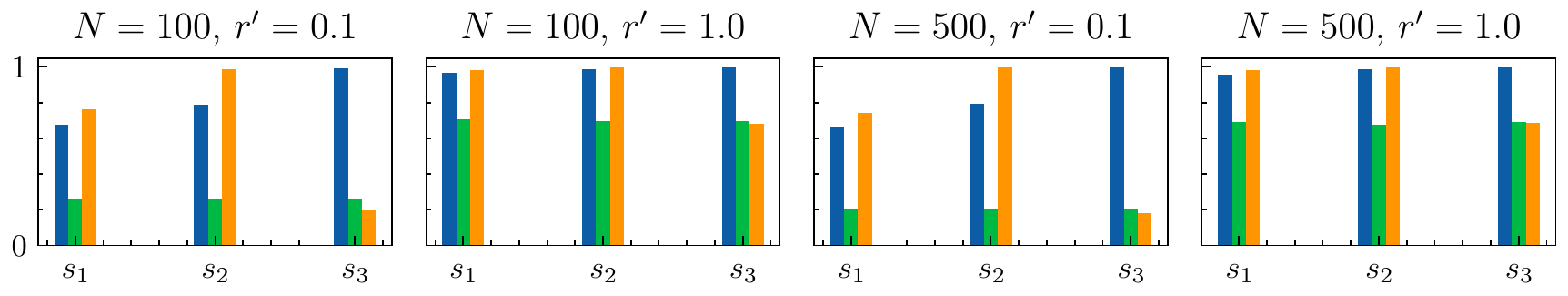}} \\
\subfloat[CIFAR-10, SIG]{\includegraphics[width=8.8cm]{\impath/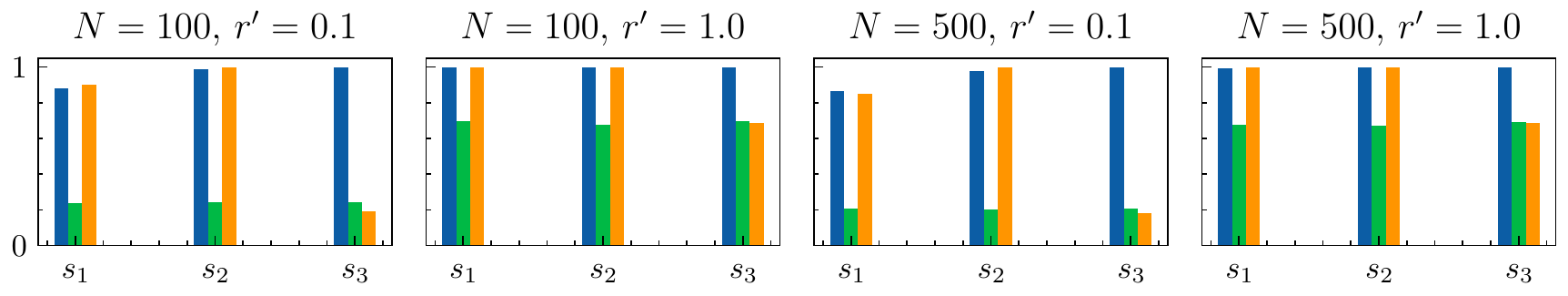}} \hfill 
\subfloat[CIFAR-10, Warped]{\includegraphics[width=8.8cm]{\impath/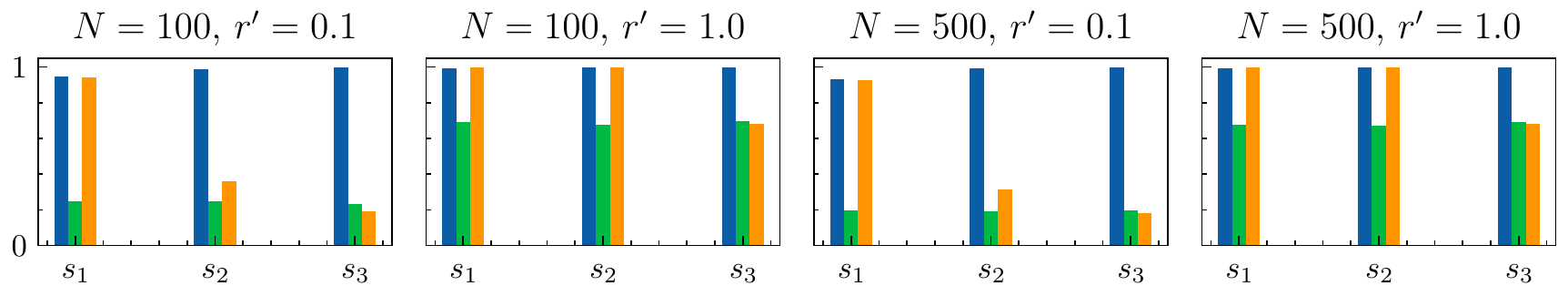}} \\
\subfloat[CelebA, Patched]{\includegraphics[width=8.8cm]{\impath/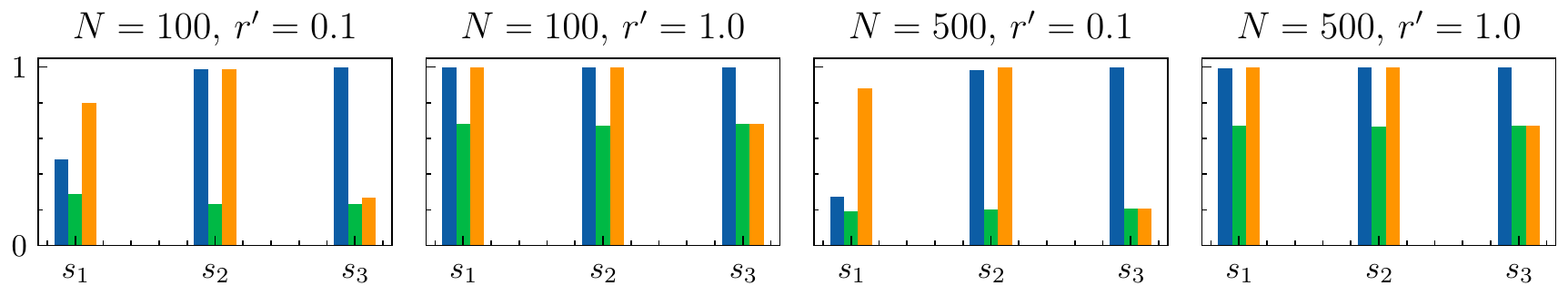}} \hfill
\subfloat[CelebA, Blended]{\includegraphics[width=8.8cm]{\impath/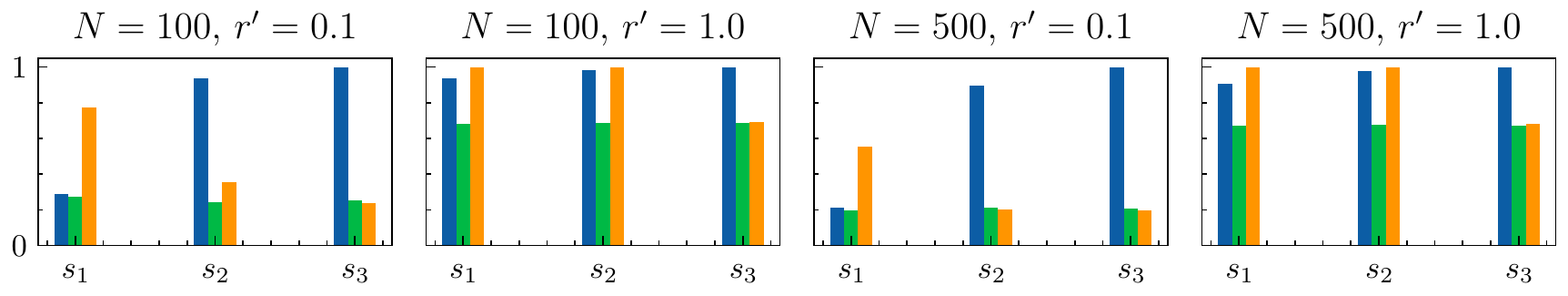}} \\
\subfloat[CelebA, SIG]{\includegraphics[width=8.8cm]{\impath/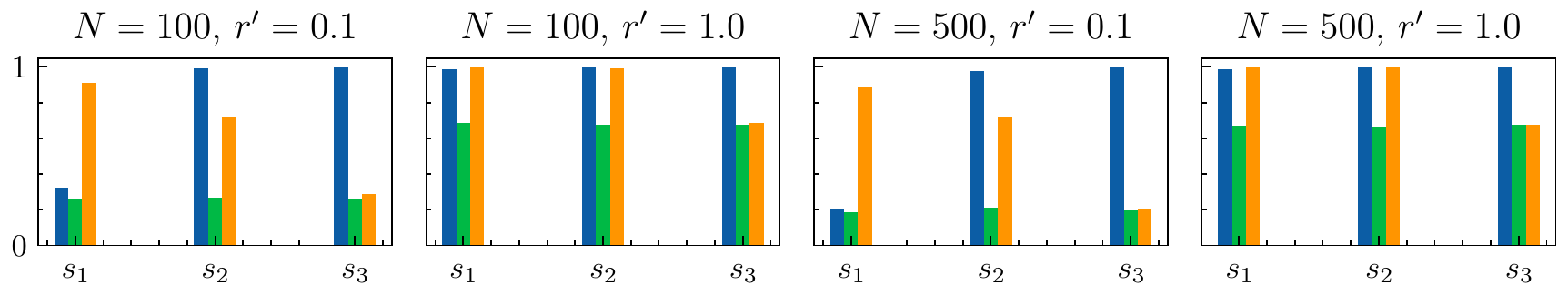}} \hfill 
\subfloat[CelebA, Warped]{\includegraphics[width=8.8cm]{\impath/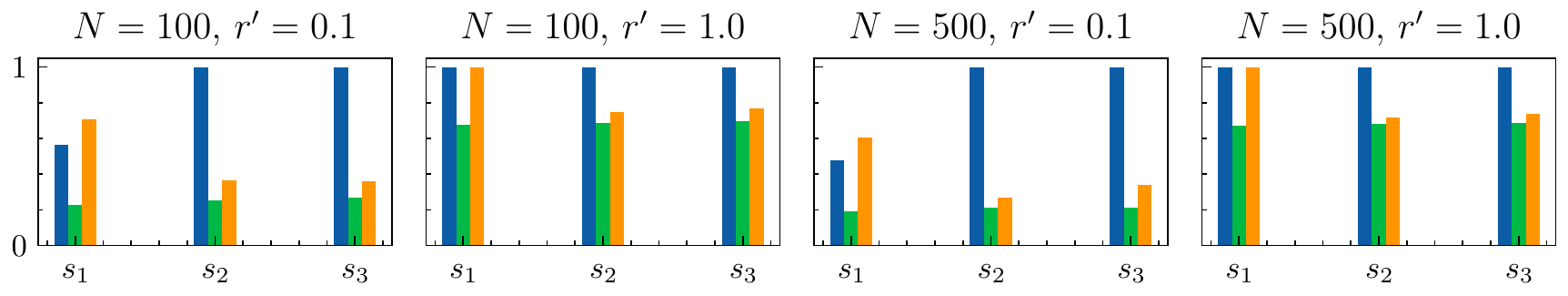}} \\
\subfloat{\includegraphics[height=0.5cm]{\impath/legend_defense.png}}
\caption{F1 scores of SS on the \{R-18\} models. X-axis: the level of features. Y-axis: the value of $F1$.}
\label{fig:ss_r18} 
\end{figure*}

\begin{figure*}[!ht]
\centering
\subfloat[CIFAR-10, Patched]{\includegraphics[width=8.8cm]{\impath/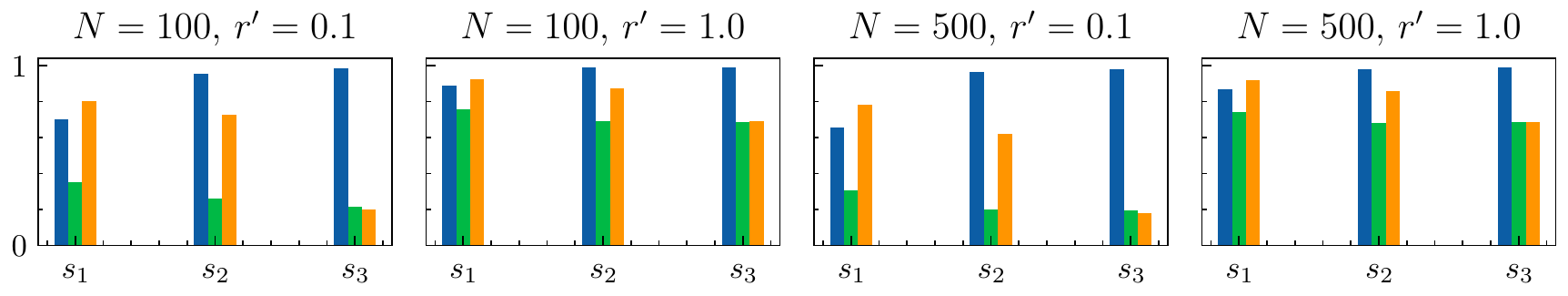}} \hfill
\subfloat[CIFAR-10, Blended]{\includegraphics[width=8.8cm]{\impath/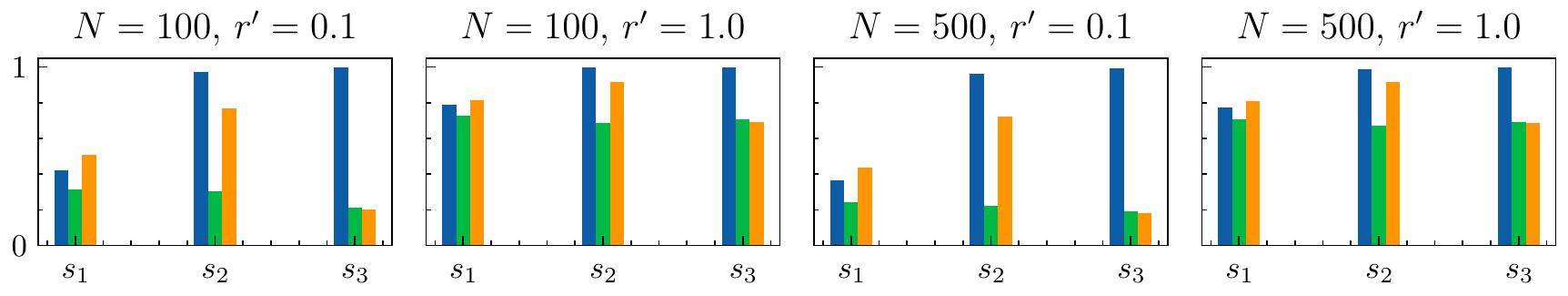}} \\
\subfloat[CIFAR-10, SIG]{\includegraphics[width=8.8cm]{\impath/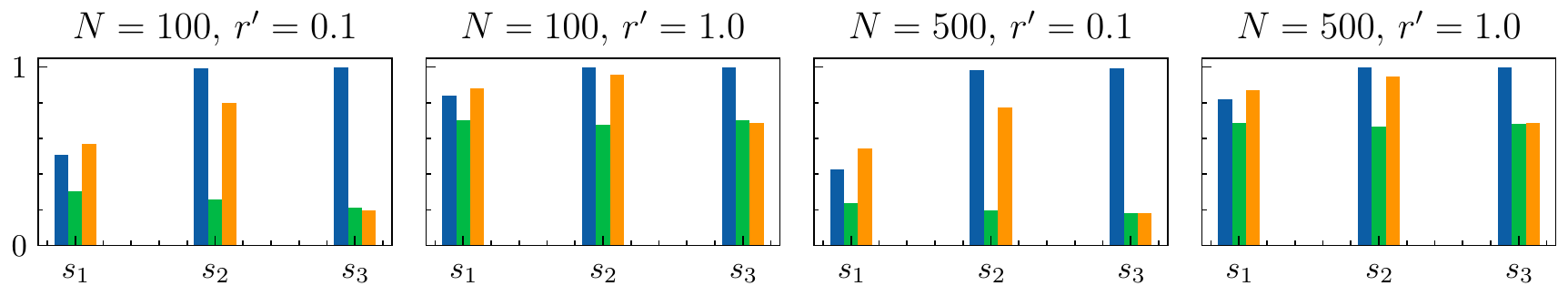}} \hfill 
\subfloat[CIFAR-10, Warped]{\includegraphics[width=8.8cm]{\impath/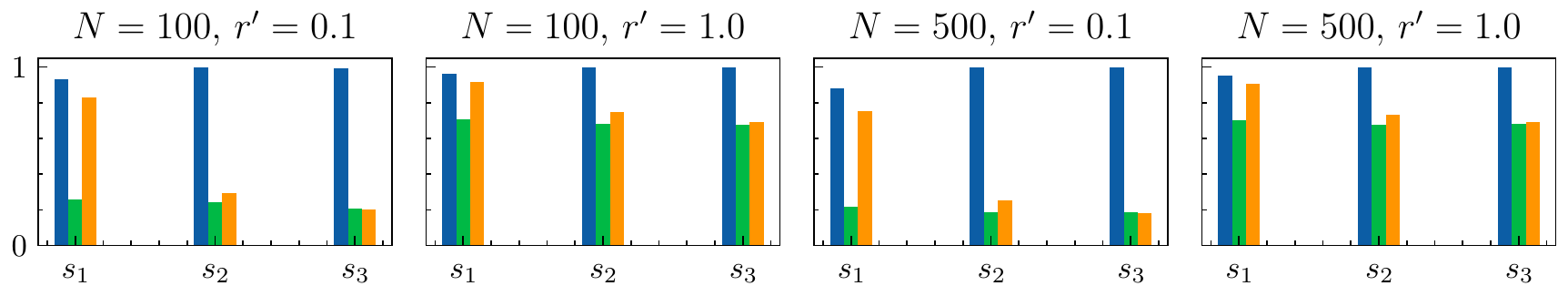}} \\
\subfloat[CelebA, Patched]{\includegraphics[width=8.8cm]{\impath/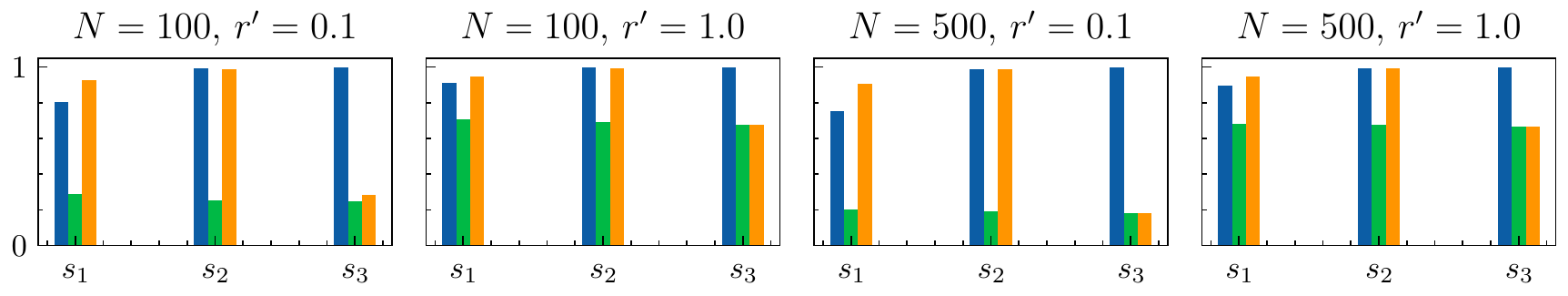}} \hfill
\subfloat[CelebA, Blended]{\includegraphics[width=8.8cm]{\impath/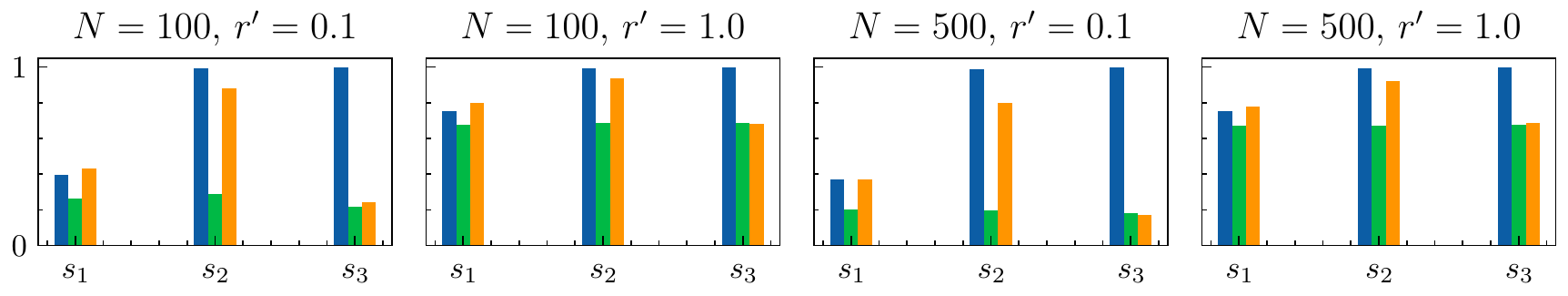}} \\
\subfloat[CelebA, SIG]{\includegraphics[width=8.8cm]{\impath/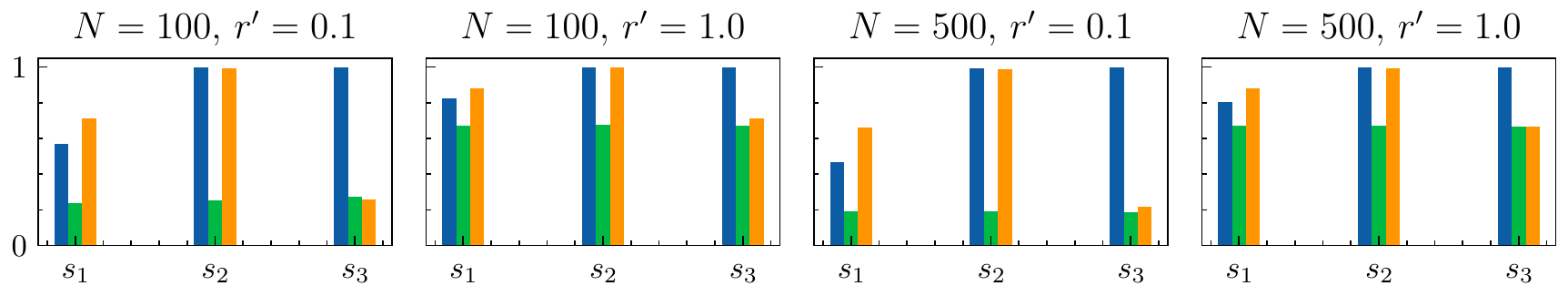}} \hfill 
\subfloat[CelebA, Warped]{\includegraphics[width=8.8cm]{\impath/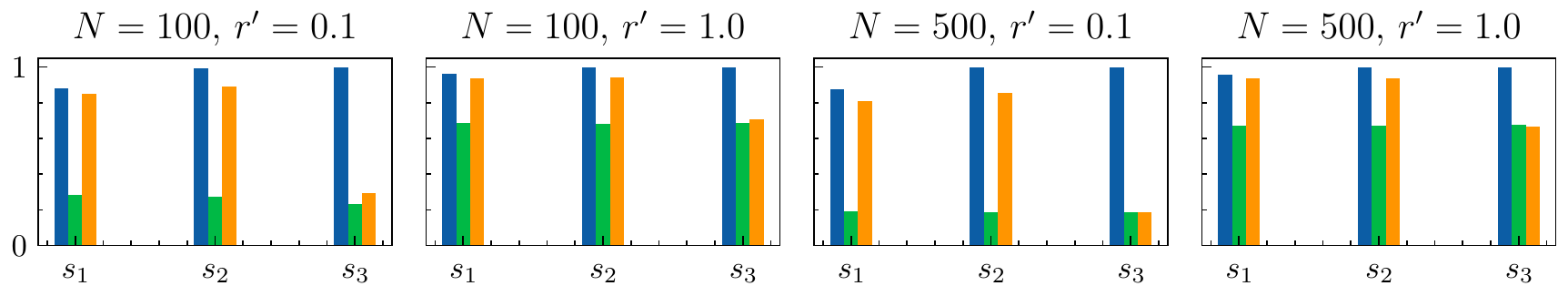}} \\
\subfloat{\includegraphics[height=0.5cm]{\impath/legend_defense.png}}
\caption{F1 scores of SR on the \{V-11\} models. X-axis: the level of features. Y-axis: the value of $F1$.}
\label{fig:sr_v11} 
\end{figure*}

\begin{figure*}[!ht]
\centering
\subfloat[CIFAR-10, Patched]{\includegraphics[width=8.8cm]{\impath/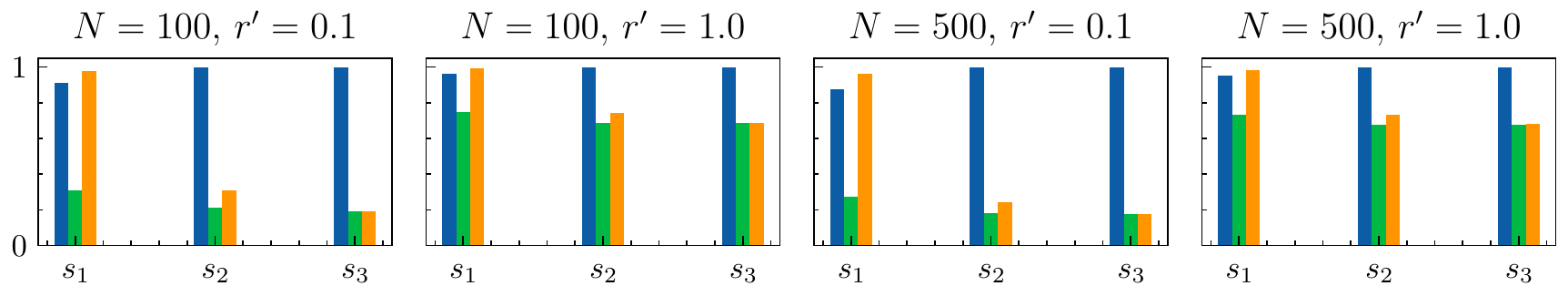}} \hfill
\subfloat[CIFAR-10, Blended]{\includegraphics[width=8.8cm]{\impath/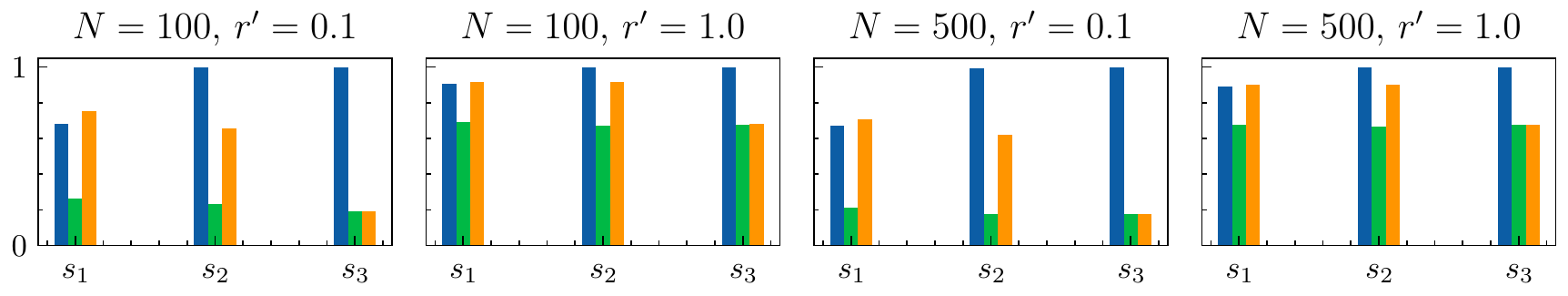}} \\
\subfloat[CIFAR-10, SIG]{\includegraphics[width=8.8cm]{\impath/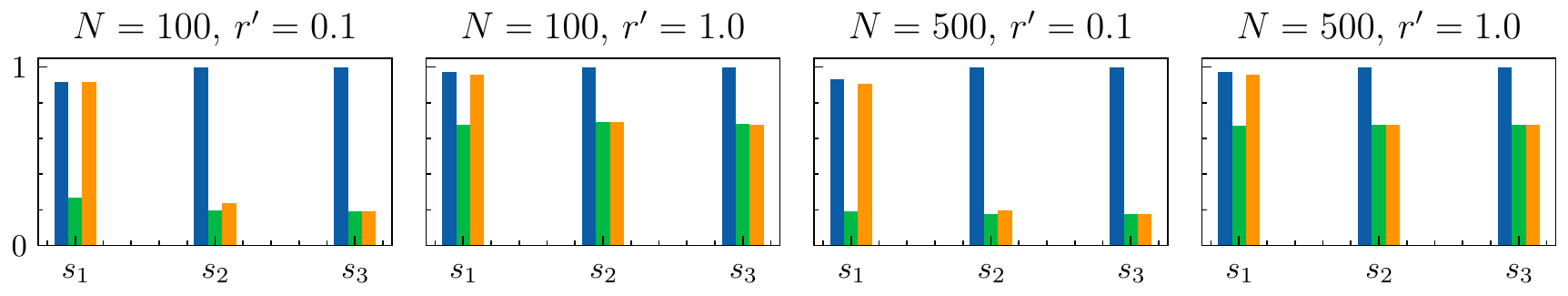}} \hfill 
\subfloat[CIFAR-10, Warped]{\includegraphics[width=8.8cm]{\impath/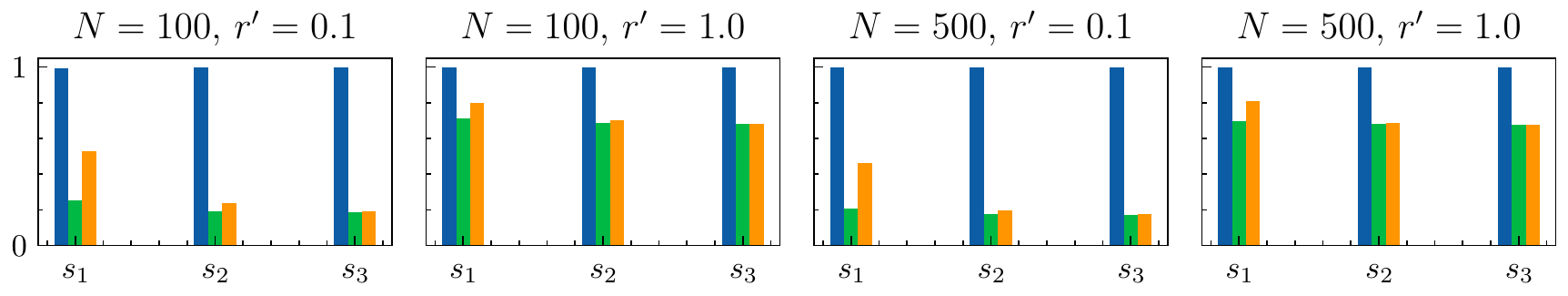}} \\
\subfloat[CelebA, Patched]{\includegraphics[width=8.8cm]{\impath/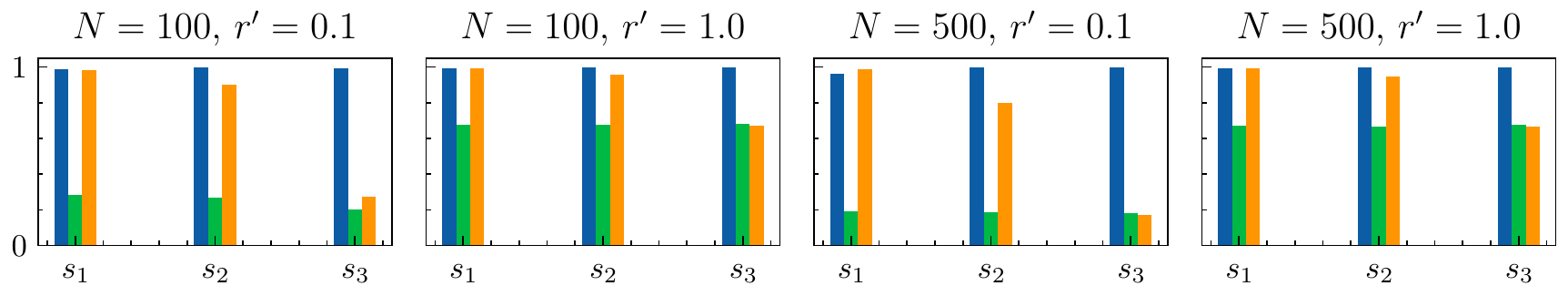}} \hfill
\subfloat[CelebA, Blended]{\includegraphics[width=8.8cm]{\impath/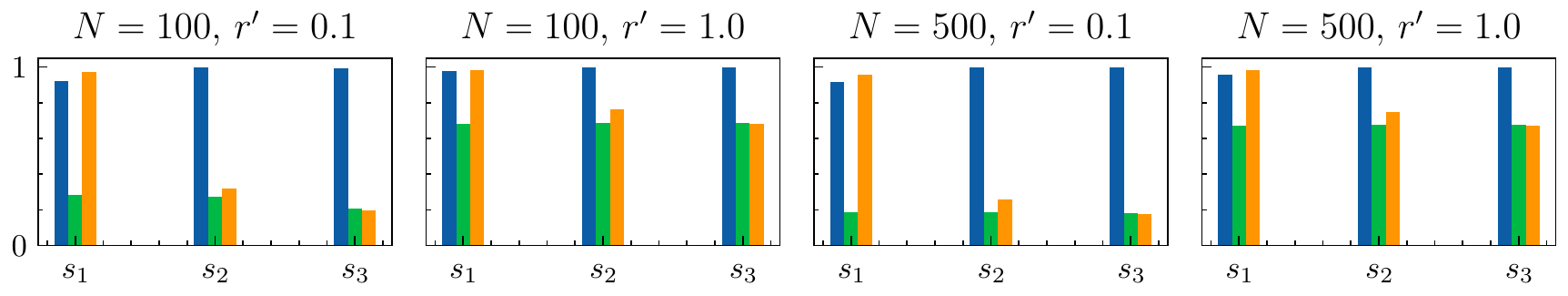}} \\
\subfloat[CelebA, SIG]{\includegraphics[width=8.8cm]{\impath/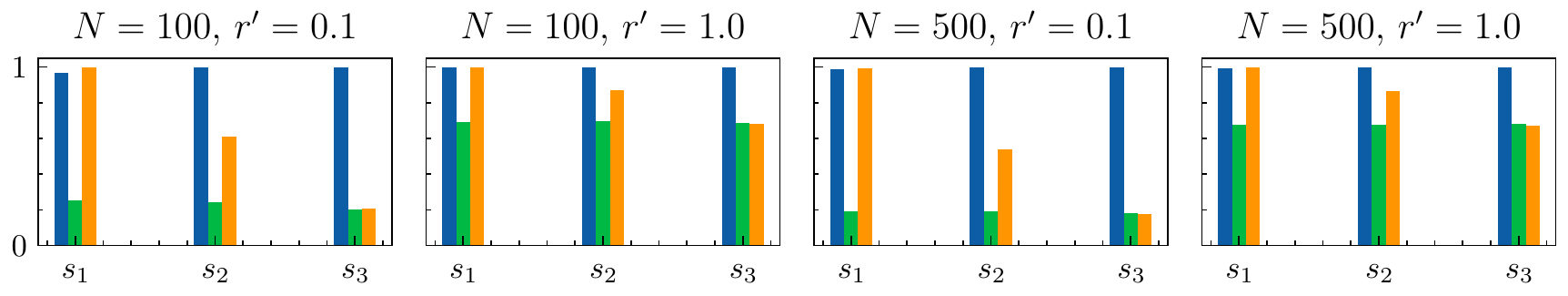}} \hfill 
\subfloat[CelebA, Warped]{\includegraphics[width=8.8cm]{\impath/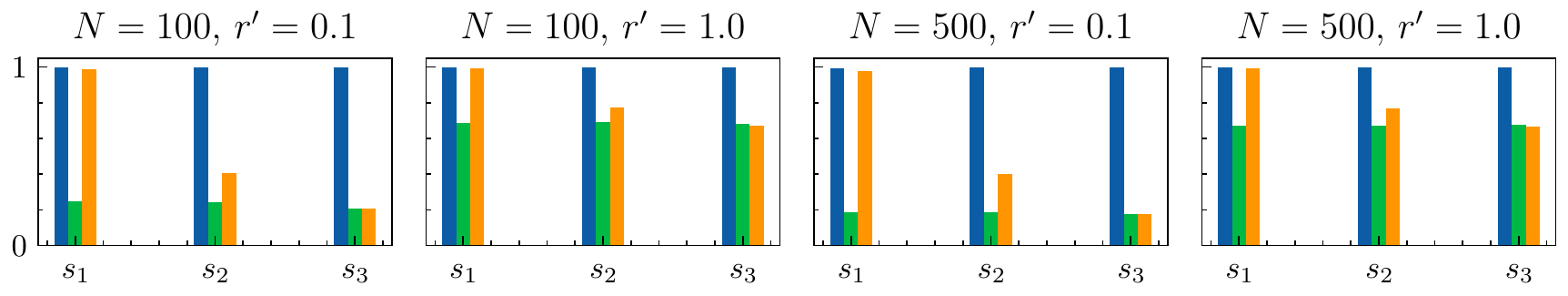}} \\
\subfloat{\includegraphics[height=0.5cm]{\impath/legend_defense.png}}
\caption{F1 scores of SR on the \{V-16\} models. X-axis: the level of features. Y-axis: the value of $F1$.}
\label{fig:sr_v16} 
\end{figure*}

\begin{figure*}[!ht]
\centering
\subfloat[CIFAR-10, Patched]{\includegraphics[width=8.8cm]{\impath/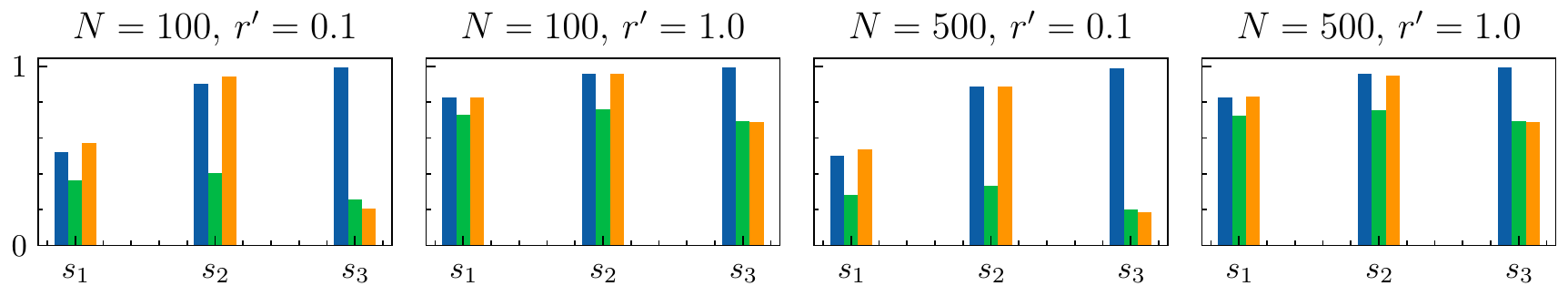}} \hfill
\subfloat[CIFAR-10, Blended]{\includegraphics[width=8.8cm]{\impath/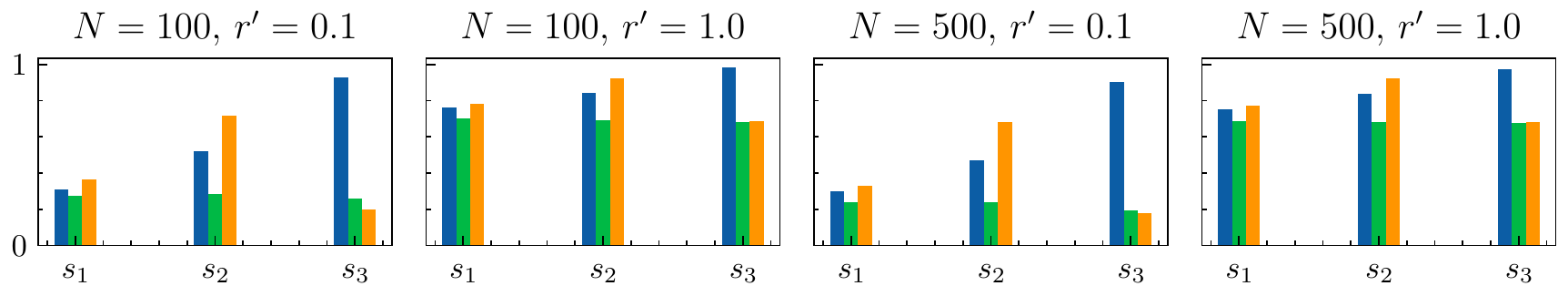}} \\
\subfloat[CIFAR-10, SIG]{\includegraphics[width=8.8cm]{\impath/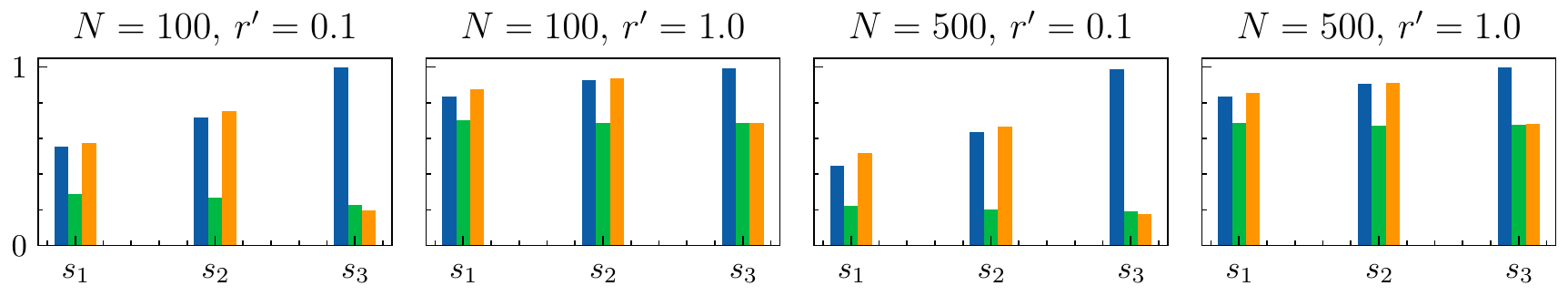}} \hfill 
\subfloat[CIFAR-10, Warped]{\includegraphics[width=8.8cm]{\impath/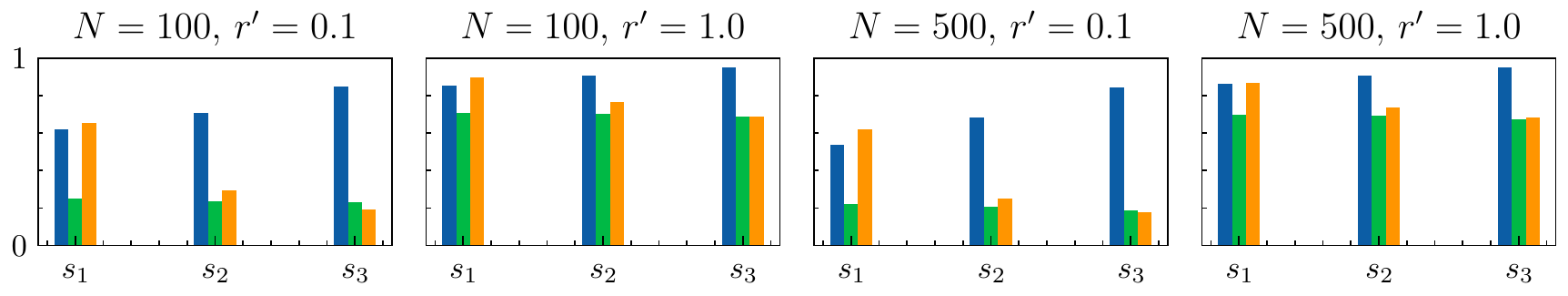}} \\
\subfloat[CelebA, Patched]{\includegraphics[width=8.8cm]{\impath/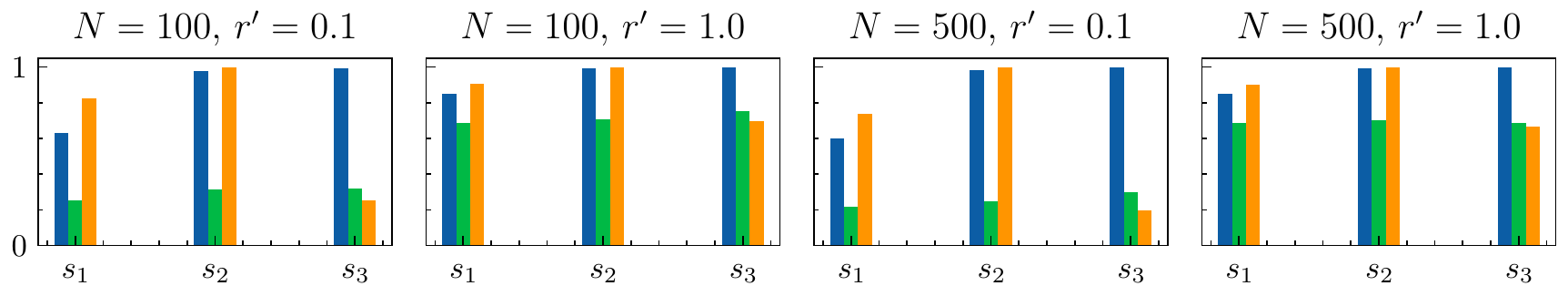}} \hfill
\subfloat[CelebA, Blended]{\includegraphics[width=8.8cm]{\impath/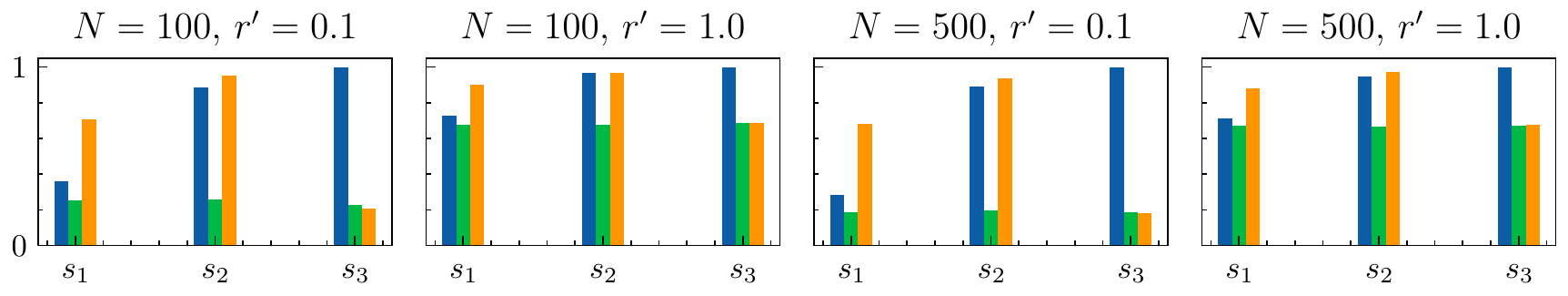}} \\
\subfloat[CelebA, SIG]{\includegraphics[width=8.8cm]{\impath/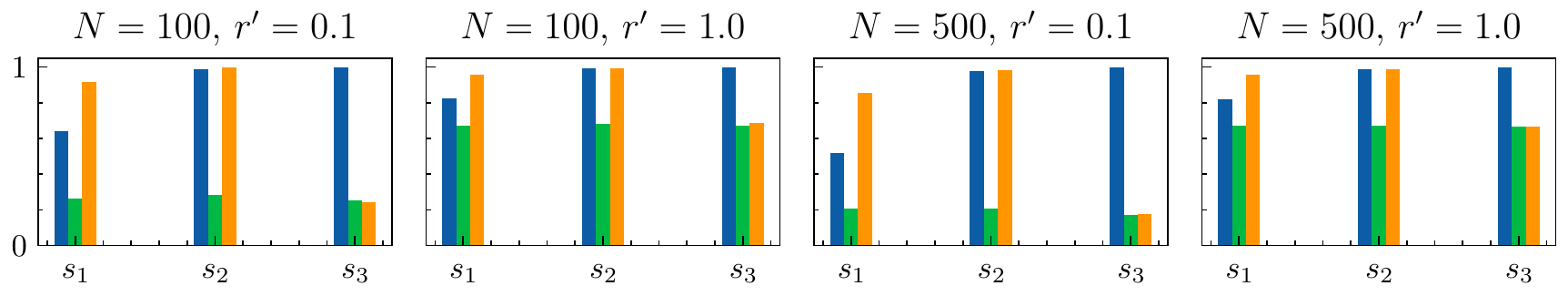}} \hfill 
\subfloat[CelebA, Warped]{\includegraphics[width=8.8cm]{\impath/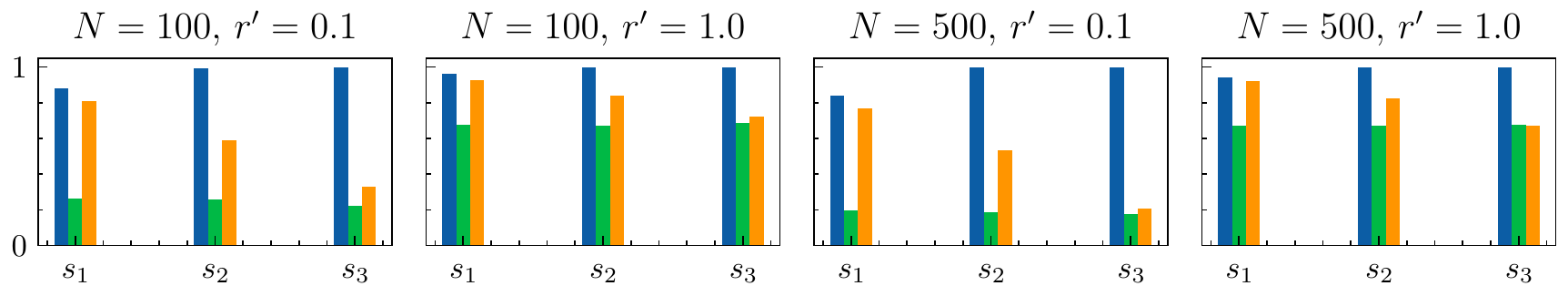}} \\
\subfloat{\includegraphics[height=0.5cm]{\impath/legend_defense.png}}
\caption{F1 scores of SR on the \{R-18\} models. X-axis: the level of features. Y-axis: the value of $F1$.}
\label{fig:sr_r18} 
\end{figure*}

The F1 scores of AC on the \{V-11\}, \{V-16\}, and \{R-18\} models are shown in Fig. \ref{fig:ac_v11}, Fig. \ref{fig:ac_v16}, and Fig. \ref{fig:ac_r18}, respectively. The F1 scores of SS on the \{V-11\}, \{V-16\}, and \{R-18\} models are shown in Fig. \ref{fig:ss_v11}, Fig. \ref{fig:ss_v16}, and Fig. \ref{fig:ss_r18}, respectively. The F1 scores of SR on the \{V-11\}, \{V-16\}, and \{R-18\} models are shown in Fig. \ref{fig:sr_v11}, Fig. \ref{fig:sr_v16}, and Fig. \ref{fig:sr_r18}, respectively. 

\end{document}